\documentclass[sn-basic]{sn-jnl}% Math and Physical Sciences Numbered Reference Style 
\usepackage{graphicx}%
\usepackage{multirow}%
\usepackage{amsmath,amssymb,amsfonts}%
\usepackage{amsthm}%
\usepackage{mathrsfs}%
\usepackage[title]{appendix}%
\usepackage[dvipsnames]{xcolor}%
\usepackage{textcomp}%
\usepackage{manyfoot}%
\usepackage{booktabs}%
\usepackage{algorithm}%
\usepackage{algorithmicx}%
\usepackage{algpseudocode}%
\usepackage{listings}%
\usepackage [latin1]{inputenc}
\usepackage{booktabs}       % professional-quality tables

\usepackage{niravstyle}
\usepackage{siunitx}
\usepackage{enumitem}
\usepackage[inline]{enumitem}

\NewDocumentCommand{\codeword}{v}{%
\texttt{\textcolor{black}{#1}}%
}

\usepackage{colortbl}
\usepackage{subfigure}

\usepackage{url}
\usepackage{niravstyle}
\usepackage{graphicx}
\usepackage{import}
\usepackage{multirow}
\usepackage{multicol}
\usepackage{rotating}
\usepackage{makecell}
\usepackage{rotating}

\usepackage{subcaption}

\usepackage{caption}

\renewcommand{\fbf}{\bfseries}

\newcommand{\method}{\fsc{ToP}\xspace}

\newcommand{\aod}[1]{\textcolor{blue!50!black}{AOD:~~#1}}
\newcommand{\tsf}[1]{\textcolor{brown!50!black}{TSF:~~#1}}

\newcommand{\revision}[1]{\textcolor{magenta}{#1}}
\newcommand{\todo}[1]{\textcolor{red!75!black}{TODO:~~#1}}

\newcommand{\done}[1]{\textcolor{green!75!black}{DONE:~~#1}}
\newcommand{\discuss}[1]{\textcolor{cyan!25!black}{DOING:~~#1}}

\renewcommand{\nsa}[1]{}
\renewcommand{\add}[1]{}
\renewcommand{\del}[1]{}
{}
\renewcommand{\tsf}[1]{}
\renewcommand{\aod}[1]{}
\renewcommand{\revision}[1]{}
\renewcommand{\todo}[1]{}
\renewcommand{\done}[1]{}
\renewcommand{\discuss}[1]{}
\renewcommand{\revision}[1]{\textcolor{black}{#1}}
\theoremstyle{thmstyleone}%
\theoremstyle{thmstyletwo}%

\theoremstyle{thmstylethree}%

\makeatletter
\renewcommand{\cite}[1]{%
  \begingroup
    \let\@citea\@empty
    (%
    \@for\@citeb:=#1\do{%
      \@citea\def\@citea{; }%
      \citeauthor{\@citeb}~\citeyear{\@citeb}%
    }%
    )%
  \endgroup
}
\makeatother

\begin{document}

\title[Topology Only Pre-Training]{Topology Only Pre-Training: Towards Generalised Multi-Domain Graph Models}


\author*[1]{\fnm{Alex O.} \sur{Davies}}\email{alexander.davies@bristol.ac.uk}

\author[1]{\fnm{Riku} \sur{Green}}\email{riku.green@bristol.ac.uk}

\author[1]{\fnm{Nirav S.} \sur{Ajmeri}}\email{nirav.ajmeri@bristol.ac.uk}

\author[2]{\fnm{Telmo M.} \sur{Silva Filho}}\email{telmo.silvafilho@bristol.ac.uk}

\affil*[1]{\orgdiv{School of Computer Science}, \orgname{University of Bristol}} %\orgaddress{\street{Street}, \city{City}, \postcode{100190}, \state{State}, \country{Country}}}

\affil[2]{\orgdiv{School of Engineering Mathematics and Technology}, \orgname{University of Bristol}}%, \orgaddress{\street{Street}, \city{City}, \postcode{10587}, \state{State}, \country{Country}}}


\abstract{The principal benefit of unsupervised representation learning is that a pre-trained model can be fine-tuned where data or labels are scarce.
    Existing approaches for graph representation learning are domain specific, maintaining consistent node and edge features across the pre-training and target datasets.
    This has precluded transfer to multiple domains.
    We present \textbf{T}opology \textbf{O}nly \textbf{P}re-Training (\method), a graph pre-training method based on node and edge feature exclusion.
    % Separating graph learning into two stages, topology and features, we use contrastive learning to pre-train models over multiple domains.
    We show positive transfer on evaluation datasets from multiple domains, including domains not present in pre-training data, running directly contrary to assumptions made in contemporary works.
    On 75\% of experiments, \method models perform significantly ($P \leq 0.01$) better than a supervised baseline.
    Performance is significantly positive on 85.7\% of tasks when node and edge features are used in fine-tuning.
    We further show that out-of-domain topologies can produce more useful pre-training than in-domain. 
    Under \method we show better transfer from non-molecule pre-training, compared to molecule pre-training, on 79\% of molecular benchmarks.
    Against the limited set of other generalist graph models \method performs strongly, including against models with many orders of magnitude larger.
    These findings show that \method opens broad areas of research in both transfer learning on scarcely populated graph domains and in graph foundation models.
    \aod{Should be 150-250 pages. Need to provide 4 to 6 keywords.}
    }

\maketitle

\section{Introduction}
\label{sec:introduction}

% Much work has been conducted on forming useful representations of graphs which are more easily used by well-explored methods for tabular data. 
Representation learning allows models to convert the useful information from rich data, for example text and images, into flexible vector encodings.
As pre-training, then fine-tuning on downstream tasks, representation learning leads to improved performance \cite{Chen2020ARepresentations, Le-Khac2020ContrastiveReview, Devlin2018BERT:Understanding}.
This is an invaluable technique where downstream labels are scarce, allowing expressive models to be applied where otherwise performance would be unsatisfactory.
Graph representation learning extends these methods to structured data, allowing much work on unsupervised learning over large datasets of graphs from a given domain \cite{Hamilton2017RepresentationApplications, Chen2020Graphsurvey, Khoshraftar2022AMethods}.

In other domains, such as images and text, it is common to pre-train on a large amalgamated dataset of samples, aiming to maximise coverage of the space \cite{Russakovsky2015ImageNetChallenge, Touvron2023LlamaModels, Ray2023ChatGPT:Scope}.
Large models pre-trained on these massive datasets then show strong performance on a large range of downstream tasks.
Models with this combination of large size, large data and strong performance are commonly referred to as \textit{foundation models} \cite{Myers2024FoundationImpacts}.

\begin{figure}[ht]
    \centering
    \includegraphics[width=\linewidth]{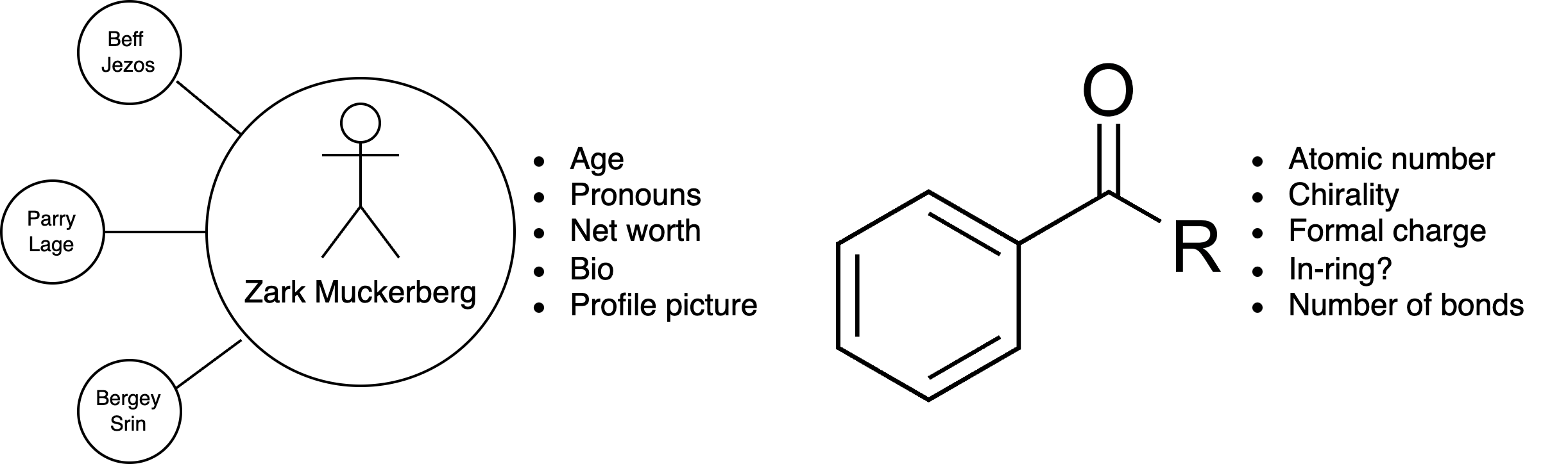}
    \caption{Graph data has rich, highly varied feature sets between domains. Here we provide model examples from social media (left) and chemistry (right). Social media feature sets might include a combination of textual, image and numeric types. Chemical feature sets are primarily numeric and categorical, representing properties of component atoms that might be difficult to infer for deep-learning models.}
    \label{fig:feature-sets}
\end{figure}

There are substantial obstacles to taking this approach for graph data, the largest being feature heterogeneity between domains.\nsa{What are those obstacles?} \aod{Hopefully clarified during paragraph}
Images or text from multiple domains are fundamentally the same data-type, uniform-size arrays of pixel values or sequences of tokens respectively, meaning that the multi-domain problem can be tackled head-on through data and model scale.
Graphs, on the other hand, have a large diversity of possible feature sets.
We present model examples of feature sets for social networks and molecules in Figure~\ref{fig:feature-sets}.
How a single model can handle highly heterogenous feature sets like these, for example both a text biography and the atomic number of an atom, is an open and challenging problem.

% Whilst there is a good deal of research into graph representation learning methods, until very recently there has been little work on methodologies to combine different domains of graph data effectively.
% Pre-training on a large amalgamated dataset of samples is ubiquitous in other fields, aiming to span the full space of the data or a realistic sample \cite{Russakovsky2015ImageNetChallenge}.
% The main obstacle to such an approach for graphs is consistency in node and edge features, as graph data presents a large diversity in possible feature-sets.
% A simple example of the differences between graph feature sets is presented in Figure~\ref{fig:feature-sets}.

This has lead current research to argue or assume, but not show, that multi-domain pre-training on graphs is not feasible \cite{Liu2023TowardsBeyond, Shirzad2022EvaluatingFeatures, Zhang2023GraphModels}. 
As a consequence there are few works attempting learning across multiple domains of graphs. 
Some very recent works have proposed approaches using LLMs as feature transformations \cite{Shang2024Path-LLM:Representation, Zhang2024CanMemorization,Xu2024LLMLearning, Liu2023OneTasks}, effectively moving a source domain into a unified text-featured domain.
% These LLM-based approaches require written schemas that transfer existing node and edge features into textual features, with the use of LLM models also drastically increasing computational expense.
This drastically increases computational expense, and  when features are not easily or commonly expressible as text, for example continuous features, LLM approaches presumably cannot be used.

% \tsf{I assume you're still reformulating the introduction, as the first sentence of the paragraph below seems misplaced.}
In downstream tasks feature information is often essential, so generalised approaches should allow them to be included in downstream applications, unlike in the very limited set of approaches for generalised graph pre-training \cite{Qiu2020GCC:Pre-Training}.
The lack of a generalised pre-training method, which allows domain features to be included downstream, excludes areas of data sparsity or scarcity from the benefits of pre-training.
Even on areas where data is plentiful enough for pre-training, it requires a different pre-trained model to be used each time, increasing the time, complexity, and cost of usage.
Multi-domain pre-training would avoid these issues, both opening transfer learning as an option for domains with limited access to data, and providing a baseline model on which applications can be built.

\subsection{Pre-Training without Features}

Graph learning can rely on information from both structure and features, with some of that information coming only from structure.
Therefore, beneficial learning during pre-training can occur with features excluded. 
Motivated by this view we introduce \textbf{\underline{T}opology \underline{O}nly \underline{P}re-Training (\method)}, a generalised pre-training method for graph models.
Under \method pre-training is conducted with node and edge features excluded, but during downstream application they can be re-included by swapping components of a given model.

% We create a multi-domain dataset, and using \method through adversarial contrastive learning, produce pre-trained models with consistently positive transfer across multiple domains and tasks.
% Transfer is also positive when node and edge features are included in downstream datasets during fine-tuning.

We investigate \method as a pre-training method for graph data that is not limited to a given domain or task.
The result of such pre-training is a single model that can be transferred with positive performance increases on arbitrary graph-level tasks and domains.
Experimentally we introduce and test two hypotheses:

% % \vspace{-0.5em}
\begin{description}[leftmargin=0em,itemsep=0em]
    \item[Hypothesis~1: Excluding Features] \label{item:h1} Pre-training on multiple domains with node and edge features excluded can lead to consistent positive transfer compared to fully-supervised, task-specific models.
    \item[Hypothesis~2: Sample Diversity] \label{item:h2} With the exclusion of node and edge features out-of-domain pre-training can have greater benefits than in-domain pre-training across downstream tasks.
\end{description}

\paragraph{Findings}

We evaluate \method using graph constrastive learning with both random and adversarial augmentations.
This evaluation includes the construction of an amalgamated, multi-domain dataset, spanning from road networks to molecules.
% We construct this dataset using both existing datasets of many graphs and varied exploration sampling over large single graphs.
We pre-train multiple models on different subsets of this amalgamated dataset and with varied augmentation schemes.
We perform multiple fine-tuning runs for each model in order to show statistical significance for our experiments.
Through this evaluation we contribute the following empirical findings:

\begin{enumerate}[itemsep=0em]
    \item \textbf{Effective pre-training.}
    Pre-training on varied domains, under the exclusion of node and edge features, leads to positive transfer. 
    We find that under fine-tuning, results are at worst on-par, and on 76\% of tasks significantly better, than a non-pretrained baseline.
    % At 95\% confidence we observe a reduction in error of 8 to 40\%.
    
    \item \textbf{Increased performance from domain diversity.}
    Pre-training without features from varied non-target domains, instead of the target domain, can bring greater downstream performance benefits.
    In the absence of features in pre-training a non-molecule learner consistently produces positive transfer compared to a molecule learner.
    On 56\% of tasks the non-molecule model performs significantly better, and never significantly worse, than the molecule-only model.
\end{enumerate}

\paragraph{Contributions}

We present two main contributions, although the work contains several more.
\aod{Including datasets, etc.}

\aod{Explicitly say novel}
Our first contribution is \method, a multi-domain pre-training method for graph models, with original domain features included during downstream tasks.
Therefore, and under recent definitions, models pre-trained using \method are close to being foundation graph models \cite{Bommasani2021OnModels, Liu2023OneTasks, Liu2023TowardsBeyond, Zhang2023GraphModels}.
% The above findings go beyond the current consensus of graph pre-training, opening broad avenues of research and application.
\nsa{This is novelty} \aod{Is a rephrase possible? ToP is definitely the main contribution} \method is the first method that, to our knowledge, has the following desirable properties:
\begin{enumerate}
    \item Positive transfer across multiple, highly diverse domains
    \item Original node and edge features during downstream transfer
    \item No very large auxiliary models
    \item No downstream feature transformations
\end{enumerate}

\revision{
Across molecular benchmarks our implementation of \method out-performs benchmarks from the limited set of generalist GNN approaches that exist.
Despite not aiming to out-perform domain specific specialist models, \method performs better-than or on-par with generalist LLMs  many orders of magnitude larger in both parameters and training dataset size.
This provides great utility when pre-training data is not abundant.
}

Our second contribution is a new research direction: with domain features excluded, diverse non-domain pre-training can be more valuable than in-domain.
This runs contrary to the single-domain training paradigm, and directly opposite to the assumptions made in other works \cite{Mao2024Position:Here, Shirzad2022EvaluatingFeatures}.
\revision{
In \citet{Mao2024Position:Here}'s review of LLMs for for multi-domain graph models, enumerating why they show promise, it is assumed that \textit{``\ldots training on [graphs from multiple domains] simultaneously shows no positive transfer benefit while increasing the risk of the negative transfer''}.
Our work shows that the opposite is true when pre-training is conducted only on structure.
This implies that greater performance through \method can be achieved simply by increasing the scale and diversity of the pre-training dataset.
% One of this work's core contributions is a demonstration of how to produce positive transfer with pre-training while maintaining flexibility in domain applications and  maintaining low computational costs.
}

\paragraph{Organisation} In Section~\ref{sec:background}, we describe the notation and concepts necessary to understand our method and provide an overview of related works.
In Section~\ref{sec:method}, we describe our generalised pre-training method \method, and our hypotheses about it as a method.
In Section~\ref{sec:experiments}, we describe how we evaluate the hypotheses detailed above, and present the findings of those experiments.
In Section~\ref{sec:discussion}, we discuss the implications of the results from these experiments, as well as list the assumptions we make and threats to validity.
In Section~\ref{sec:conclusion}, we conclude with avenues for future research.

\section{Background and Related Work} \label{sec:background}

Here we describe the concepts necessary to understand the rest of our method, as well as closely related works.

\paragraph{Representation Learning}
In most contexts, representation learning is the process of forming useful numerical encodings of non-tabular input data through an `encoder' \cite{Bengio2013RepresentationPerspectives}.
Let $X \in \mathcal{X}$ represent a random variable $X$ in the data space $\mathcal{X}$, and $Y \in \mathcal{Y}$ the same for the output space.
Representation learning aims, given their joint distribution $P(X, Y)$, to learn a mapping $X \rightarrow \hat{X}$ that maximises their mutual information $I(\hat{X};Y)$.
In basic terms, representation learners aim to transfer some piece of rich data, for example images, into a vector that describes that piece of data.

\paragraph{Contrastive Learning (CL)}

CL is a semi-supervised school of representation learning based on altering input samples \cite{Le-Khac2020ContrastiveReview}.
Let $\mathcal{T}$ be a set of small transformations called `augmentations', where $t \in \mathcal{T} : X \rightarrow X'$.
These augmentations are generally designed to manipulate the input sample in a way non-destructive to the `useful' information it represents.
\revision{
Augmentations can be understood as a method to increase the semantic coverage in the input space.
% So for any augmentation $t\in \mathcal{T}$, $t(x) \sim x \rightarrow f(t(x)) \sim f(x)$}.
}
Common choices for images, for example, are gray-scaling and rotations. 
% \revision{The choice of augmentations used in contrastive learning can be seen as an inductive bias \cite{chen2020simple}. 
\revision{Intuitively, the augmentation defines an invariant transformation for the encoder to learn.}
Given an encoder's mapping $F(X_i) \rightarrow \hat{X_i}$, multiple augmentations of the same input sample should be as similar as possible.
Again in simple terms, contrastive learners aim to represent slight variations on the same sample in similar ways, for example through cosine similarity.

\paragraph{Graph Representation Learning}
\citet{Hamilton2020GraphLearning} provides a comprehensive summary of methods for graph representation learning.
Traditional methods to encode graphs as vectors include graph statistics such as centrality measures and clustering coefficients, but also kernel methods such as the Weisfeiler-Lehman kernel \cite{SchweitzerPASCAL2011Weisfeiler-LehmanBorgwardt}. 
Expressive Graph Neural Networks (GNNs) models, generally through parameterised message-passing, can produce powerful representations of graphs.
A number of GNN architectures have been proposed with the Graph Isomorphism Network (GIN) being the most expressive  in \revision{message-passing} representation learning contexts \cite{Zhou2020GraphApplications}.

\paragraph{Graph Contrastive Learning}
Multiple studies exist for graph contrastive learning  \cite{Chu2021CuCo:Learning., Hafidi2022NegativeRepresentations, Hassani2020ContrastiveGraphs, Hu2020StrategiesNetworks, Li2021DisentangledGraphs, You2020GraphAugmentations, Zhu2021AnLearning}. 
The typical graph contrastive learning formulation, Graph Contrastive Learning (GraphCL), is put forward by \citet{You2020GraphAugmentations}. 
In GraphCL and many subsequent works, such augmentations are performed randomly using fixed parameters, i.e., \fsl{`drop 20\% of nodes'}.
\revision{These augmentation strategies do not vary based on the graph-instance being augmented.}

\paragraph{Adversarial Graph Contrastive Learning}
The quality of augmentations is critical for graph contrastive learning \cite{ You2020GraphAugmentations, Zhu2021AnLearning}.
To maximise expressivity in these augmentations adversarial strategies have been proposed \cite{Hassani2020ContrastiveGraphs, Zhao2023UnsupervisedLearning}.
\citet{Suresh2021AdversarialLearning} propose Adversarial-Graph Contrastive Learning (AD-GCL), with their method training both an encoder and a `view-learner' $\mathcal{T}_{\theta}$. 
AD-GCL performs the standard GCL protocol with weight updates to a parameterised ($\theta$) set of augmenting functions $\mathcal{T_{\theta}}$ through the contrastive loss.
Here $\mathcal{T_{\theta}}$ learns augmentations that minimise mutual information between the source and augmented graphs through edge dropping.
The encoder, adversarially to this, learns to maximise shared information between the respective augmentations.
The updates for $\mathcal{T_{\theta}}$ are regularised such that augmentations are not entirely destructive.

% \subsection{Related Work} \aod{Roll into background -> "background and related work". GCC section goes into graph contrastive learning section. LLM approaches is last para. Expand LLM section, elaborate on different approaches, describe downsides.}

\revision{
\paragraph{Large Language Models on Graphs}
}
\revision{
Large Language Models (LLMs) are models pre-trained, typically generatively, across massive corpuses of natural language data.
Relying on scaling through both dataset size and model size, such models are highly performant on natural language processing.
The first foundation models, though trained in the same manner, were smaller and typically required fine-tuning for strong performance on a specific task.
Bi-Directional Encoder Representations from Transformers (BERT) is the archetypal example here \cite{Devlin2018BERT:Understanding}.
The most recent LLMs vary in size from a few billion parameters \cite{Dubey2024TheModels} to many hundreds of billions \cite{Ray2023ChatGPT:Scope}.    
}

\revision{
LLMs have in very recent research been applied to building multi-domain graph models, often termed Graph Foundation Models (GFMs) in their respective works.
Some have argued that LLMs themselves, by default, may be GFMs \cite{Mao2024Position:Here}.
LLMs carry the significant benefit of allowing one-shot or few-shot application on downstream tasks in some circumstances.
However, given their size, LLMs can drastically increase the cost and complexity of using graph models.
}

\revision{
The simplest approach for applying LLMs on graph data is simply to encode a graph structure in natural language, along with features.
We model our description of this approach on \citet{Fatemi2023TalkModels}.
Here a graph structure is encoded via some schema, for example the edge $(Callum, Zack)$ might be encoded \textit{`Callum and Zack are connected'} in a social network.
\citet{Fatemi2023TalkModels} find that the choice of encoding schema, for example \textit{`is friends with'} vs \textit{`is connected to'}, has a significant impact on performance, and that in-general this approach is limited.
}

% An approach to multi-domain graph models, developed using Large Language Models (LLMs), is to
% move - or maintain - existing features as textual 
\revision{
Alternatively the performant capacity of LLMs as encoders has been used to unify the graph feature space across domains.
By describing the features of a node or edge in natural language, the LLMs encodings of that new textual feature can be shared between domain feature sets \cite{Shang2024Path-LLM:Representation, Zhang2024CanMemorization,Xu2024LLMLearning, Liu2023OneTasks}. 
This assumes and requires that:
\begin{enumerate}
    \item Original features are easily encoded in text
    \item Natural language versions of features are meaningful
    \item LLM encodings are in fact a coherent embedding space
\end{enumerate}
% Where features are not textual regular expression schemas are used to create textual features.
For natural-language features, the approach is trivial, and numeric features are converted to text via a schema.
An example for molecular graphs would be \textit{`An atom with X protons, Y covalent bonds, which (is/is not) part of a ring structure'}.
These textual features are then passed to an LLM to form numeric encodings as features.
This has the significant advantage of allowing one-shot or few-shot downstream use, assuming that downstream text features are close to in-domain.
The approach  has yet to be applied to graphs with rich feature sets.
We assume this  because it is not clear how previous approaches would encode non-intuitive graph features textually, such as continuous ones.
% Indeed 
Many of these initial works use multiple domains where features are already text, and do not attempt to move other feature sets into textual features.
}

\revision{
Another approach has emerged in LLMs for chemistry, where there are expressive sequential descriptions of molecular graphs commonly used in natural language \cite{Liu2023MolXPT:Pre-training, Zhao2024ChemDFM:Chemistry, Cao2023InstructMol:Discovery}.
Here LLM approaches use sequential SMILES encodings of molecules, for example O=C=O from CO$_2$, to represent molecules at some point of their pipeline.
SMILES strings, as a natural language expression of a graph structure, are a useful tool for LLM approaches.
Most other graph structures, for example social networks, are much harder to express in natural language.
A core assumption of these models, with strong evidence for validity through experimentation, is that forming datasets of text surrounding these SMILES encodings using chemical literature allows learning about chemical dynamics to occur.
Where these models do not include explicit usage of graph structures, for example InstructMol \cite{Cao2023InstructMol:Discovery} and ChemDFM \cite{Zhao2024ChemDFM:Chemistry}, they require far larger models and datasets for the same performance.
}

\revision{
\method carries few of the downsides associated with LLMs for graph data, yet achieves many of the same goals.
We show that \method models have strong downstream performance across domains on structure-only tasks, and in comparison, LLMs perform far more weakly.
Despite the suitability of molecular tasks for LLMs, as described above, \method performs at a level comparable to the best-performing LLMs on molecular tasks.
This comes at a massively smaller time and computational cost compared to LLMs.
Our reasonably large GNN encoders complete pre-training in less than half a day, on consumer hardware, with fine-tuning and inference both feasible and fast on consumer laptops.
}

\revision{
\paragraph{Graph Constrastive Coding}
A similar work to \method is Graph Contrastive Coding (GCC) \cite{Qiu2020GCC:Pre-Training}.
% Focusing on local explorations of large graphs, GCC also omits node and edge features, and designs its own contrastive learning scheme to produce graph encoders.
GCC begins by generating subgraphs from various large graphs through ego sampling.
As augmentations it used random walks with restarts, with embeddings of the seed node used for its contrastive loss, resulting in a local-focus model.
This is in contrast to this work, where we focus on individual graph level representations, instead of pre-training as a route to tasks on large graphs.
}

\revision{
As in this work, GCC's primary aim is to sidestep the issue of varied feature sets between graph domains.
\method simply removes feature sets during pre-training.
GCC goes a step further, replacing existing node and edge features with positional information, principally the graph Laplacian.
While this has obvious merit in encoding structural information, models can learn from specific local-level features, allowing overfitting and other feature-space biases to occur.
Our assumption, validated through experiment, is that such an approach also makes re-introduction of domain-specific features much less effective.
Where we do not extend GCC by including domain features during transfer, using structural encodings as in the original work, performance on out-of-domain data is weak.
On out-of-domain structure-only tasks, \method models perform more strongly or on-par with GCC models, despite lacking the advantage of structural features included with GCC.
}

\section{\method: Topology Only Pre-Training} \label{sec:method}

% Generalised pre-training with domain features included is an highly challenging task.
% One naive approach would be to pass features as-is, with padded vectors where feature sets are of a lower dimensionality.
% Here components would be shared across domain features, representing an extremely difficult space to learn.
% Alternatively, one could reserve components of a long feature vector for each feature set, with padding tokens in components for other domains and feature sets.
% This would require learning a massive span of features and preclude movement to new feature sets.
% Feature-management approaches, which move domain features to the same feature set \cite{Liu2023OneTasks}, are in their infancy.

Assume a dataset with features $X \in \mathbf{R}^{|V| \times D}$ and similar for edges, topology as an edgelist $E:\{ (v_1, v_2), \ldots \}$, and graph labels $y$, with graphs $G:\{X, E\}$.
$y$ is some target label.
When a given model aims to learn some mapping $f(G) \rightarrow y$, in reality this is $f(X, E) \rightarrow y$.
% We view graph learning as a task occurring on both topologies $G$ and features on that topology $X$.
% We can assume that structure and features, together or independently, contain useful information for a given downstream task.
The usual graph learning assumption is that topology $E$ and features $X$ must be considered together for information to be useful.
In other words their combination has mutual information with a target $y$, $I(y ; X, E)$.

We can assume that in the absence of features, structural information is still present, i.e. $I(y;E) \geq 0$.
From this view, pre-training without features is well-motivated, with models learning generalised information about graph structures before features are introduced.
This information from $E$ is useful for downstream tasks, even while missing task-specific information from features.
This is an abstraction away from feature-specific correlations, with abstraction a core quality of effective representation learners \cite{Le-Khac2020ContrastiveReview}.
Removing features $X$ necessarily removes their own information, but also the interdependence of features and structure:
\begin{equation}
    \label{eqn:information_terms}
    I(y;E,X) \geq I(y;X) + I(y;E)
\end{equation}
In situations where data is abundant, therefore, pre-training on both $X$ and $E$ will likely still to higher performance.

% Pre-training without features would then provide a smoother loss landscape for downstream tasks \cite{Liu2019TowardsRepresentations}.

Graph without features are sampled from the same high dimensional space, which is highly constrained in real world graphs.
As such they can be similar across highly varied domains, much as in the image space, even if they are entirely heterogenous with features included.
On the local scale at which message passing GNNs operate, the same patterns may occur across any number of domains.
In this case a wider range of topologies could span a wider range of the basic graph space, and so contain more useful information, than a narrow range closer to a downstream domain.

This pre-training on only topologies would give an initial bias towards structural features during downstream transfer.
Such a structural bias is encouraged in other works, in particular those works on Topological GNNs \cite{Horn2021TopologicalNetworks,Chen2021TopologicalGraphs,Wen2024Tensor-viewNetwork}.
Here layers or models themselves are designed to encourage learning from patterns in structure over patterns in features.

\textbf{\underline{T}opology \underline{O}nly \underline{P}re-Training (\method)} is a method for generalised pre-training of graph models.
Under \method pre-training is conducted only on topologies, with node and edge features excluded through replacement with identical values.
This use of only topologies $G$, instead of graphs with specific features $\underline{G}$, allows pre-training on a multi-domain dataset.
In the works prior to \method, the use of such a graph dataset was not tractable \cite{Liu2023TowardsBeyond, Shirzad2022EvaluatingFeatures, Zhang2023GraphModels}.

\begin{figure}[ht]
    \centering
    \includegraphics[width=\linewidth]{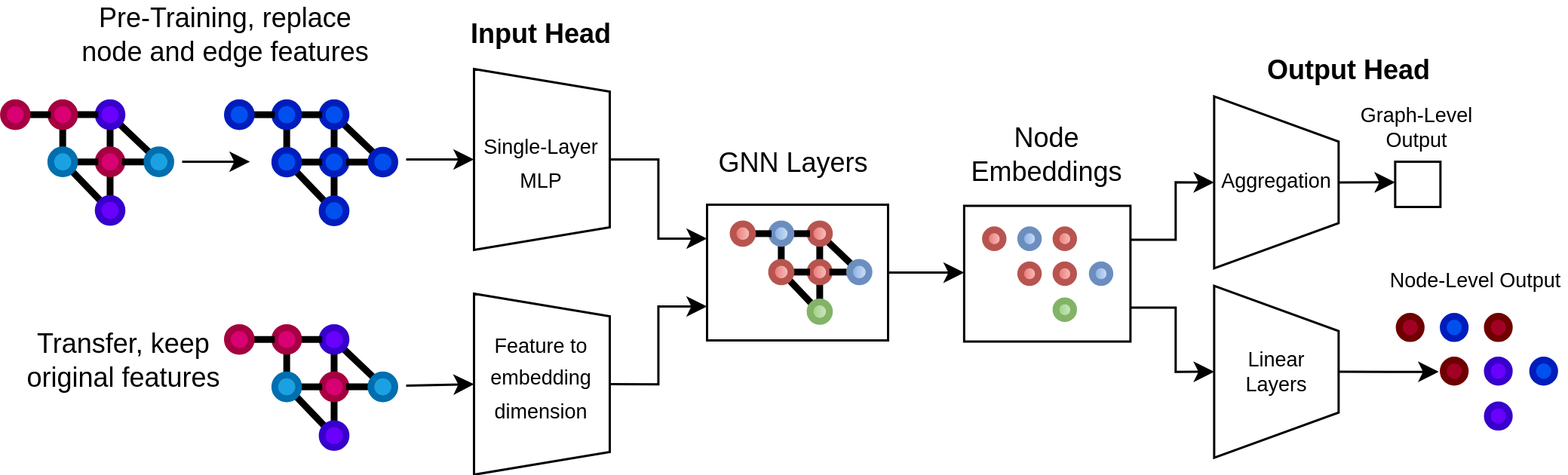}
    \caption{A schematic for \method models. During pre-training, features are replaced with a single integer label for each node or edge (all identical), which are then passed to a single-layer MLP. This results in an identical input vector for each node and edge. During transfer, a different input head is used, moving the original dimensions to the hidden dimensionality of the GNN encoder block. This allows arbitrary node and edge features to be included in transfer. Omitting the output head of the model allows transfer onto node and edge-level tasks.}
    \label{fig:input-output-head}
\end{figure}

Figure~\ref{fig:input-output-head} shows a schematic for \method models. 
\method's implementation is elegant:
Features are removed from nodes and edges and replaced with some arbitrary identical value.
In our case all nodes and edges carry the integer value 1.
With the input head some simple MLP, the choice of value has no significant impact, as the GNN layers are presented the same vector for all nodes and edges.
An encoder is then pre-trained using some unsupervised representation learning method.
In our implementation we use contrastive learning for pre-training, given its strong performance in representation learning on graphs.

By swapping the input head of the pre-trained model,
arbitrary node and edge features can be included during transfer, allowing integration of feature information alongside structural during downstream transfer.
In the same manner the output head of a model can be replaced or excluded entirely, allowing transfer onto node and edge-level tasks.

% We have two hypotheses concerning \method.

\subsection{Hypothesis~1: Excluding Features} 
\label{sec:c1}

\revision{
\textbf{Pre-training on multiple domains with node and edge features excluded can lead to consistent and positive transfer compared to fully-supervised, task-specific models.}
The motivating concept behind \method is that as graph learning requires learning on both structural and feature levels, pre-training on only structure can lead to useful knowledge transfer.
As such, a model that hasn't undergone such pre-training should perform worse on downstream tasks when applied under the same supervised training conditions.
}

Supporting Hypothesis~1 means producing a model that has learnt useful representations of graphs \fsl{in general}.
Our first step to achieving this is to construct a dataset that covers a variety of graph domains.
Such a dataset should contain graphs that present a large diversity of multi-node patterns.
Our approach is simple: include graphs from several domains and assume that there is adequate coverage of the multi-node patterns that occur in real-world graphs for a given set of downstream tasks.
The same approach has been taken in the domains of text \cite{Brown2020LanguageLearners} and images \cite{Rombach2022High-ResolutionModels}, where enormous numbers of data samples are scraped from the web for generative pre-training.
This relies on two assumptions \textbf{1)} this dataset adequately covers the relevant space for downstream transfer, and \textbf{2)} a single model can usefully represent such a space.
These assumptions, as in other fields, are justified through performance metrics on downstream tasks on seen and unseen domains.

We formulate our test for Hypothesis~1, that pre-training on graphs without node or edge features is beneficial, as beating a simple supervised baseline.
Said baseline is a model with the same architecture as the pre-trained model but without pre-training.
Both models are then trained, or fine-tuned for the pre-trained model, on the same downstream datasets under the same conditions.
On each downstream task, should the pre-trained model match or out-perform the supervised model, this represents positive transfer.
Positive transfer on a task indicates that at least some of the learning during pre-training was beneficial to subsequent learning during fine-tuning.
A positive transfer rate across datasets of more than 50\% indicates that in-general transfer is positive, thus meeting the minimum criteria for Hypothesis~1.
A high positive transfer rate strongly supports Hypothesis~1.

\subsection{Hypothesis~2: Sample Diversity} \label{sec:c2}

\revision{
\textbf{With the exclusion of node and edge features, out-of-domain pre-training can have greater performance benefits than in-domain pre-training across downstream tasks.}
If structure-only learning is useful, as proposed by Hypothesis~1, it follows that out-of-domain of pre-training data with greater diversity  may be more useful than in-domain.
This relies on the assumption that a given encoder is learning to represent local node patterns, and as diverse data presents a wider range of such patterns, the model must learn more expressively how to represent such patterns.
}

From Hypothesis~2 we expect that for a given task models trained on a more diverse pre-training domain can bring better transfer than a model trained on the target domain.
We do not argue that this is always true, only that in some settings out-of-domain samples can be more useful for pre-training than in-domain.
This has clear parallels to the performance-by-scale approach taken in other fields \cite{Russakovsky2015ImageNetChallenge, Devlin2018BERT:Understanding, Myers2024FoundationImpacts}.
To this end we compare two models, the first trained on a dataset from a single domain, as is the current practice.
This single domain matches that of the downstream target dataset(s).
The second model is instead trained on the rest of the data \fsl{while excluding the target domain}.

The total size of the pre-training datasets for each model should match for fair comparison.
By pre-training both models under the same controlled conditions (hyperparameters etc.), then evaluating their performance under fine-tuning, we directly compare pre-training on a single domain to pre-training on multiple domains.
General positive transfer from the multi-domain model, in comparison to transfer from the in-domain model, would constitute a clear supporting evidence for Hypothesis~2.

\subsection{Models for Evaluation}

In order to address these hypotheses, we apply \method over five principal different combinations of pre-training data and augmentation scheme.
\begin{description} %[label=(\textbf{\arabic*})]
    \item[\textbf{GIN}] An untrained GIN model of identical architecture to the other methods. This serves to benchmark what constitutes positive transfer.
    
    \item[\textbf{\method-Chem}] An AD-GCL model fit only on molecules. This constitutes an in-domain model, albeit in the absence of domain features.
    
    \item[\textbf{\method-Social}] An AD-GCL model fit only on non-molecules. This serves as a contrast to \method-Chem, allowing us to assess the benefits of varied out-of-domain pre-training in comparison to in-domain pre-training.
    
    \item[\textbf{\method-All}] An AD-GCL model fit on all the training data. This is our primary measure of the effectiveness of \method.
    
    \item[\textbf{\method-All (CL)}] A model trained across the whole amalgamated dataset with the loss function from \citet{You2020GraphAugmentations} using random edge dropping ($\lambda=0.2$). This allows us to assess the impact of adversarial augmentations in \method pre-training.
\end{description}

Models during pre-training, unless specified, are six GIN \cite{Xu2019HowNetworks} layers with single linear layers as input and output.
The GIN model acts as our supervised baseline (Hypothesis~1 in Section \ref{sec:experiments}).
The \method-Chem/\method-Social models indicate the impact of training data diversity on downstream transfer performance, particularly on our downstream molecular datasets (Hypothesis~2).
The \method-All (CL) model shows the impact of adversarial augmentations.

\begin{figure}[hbtp]

\centering
\includegraphics[width=\linewidth]{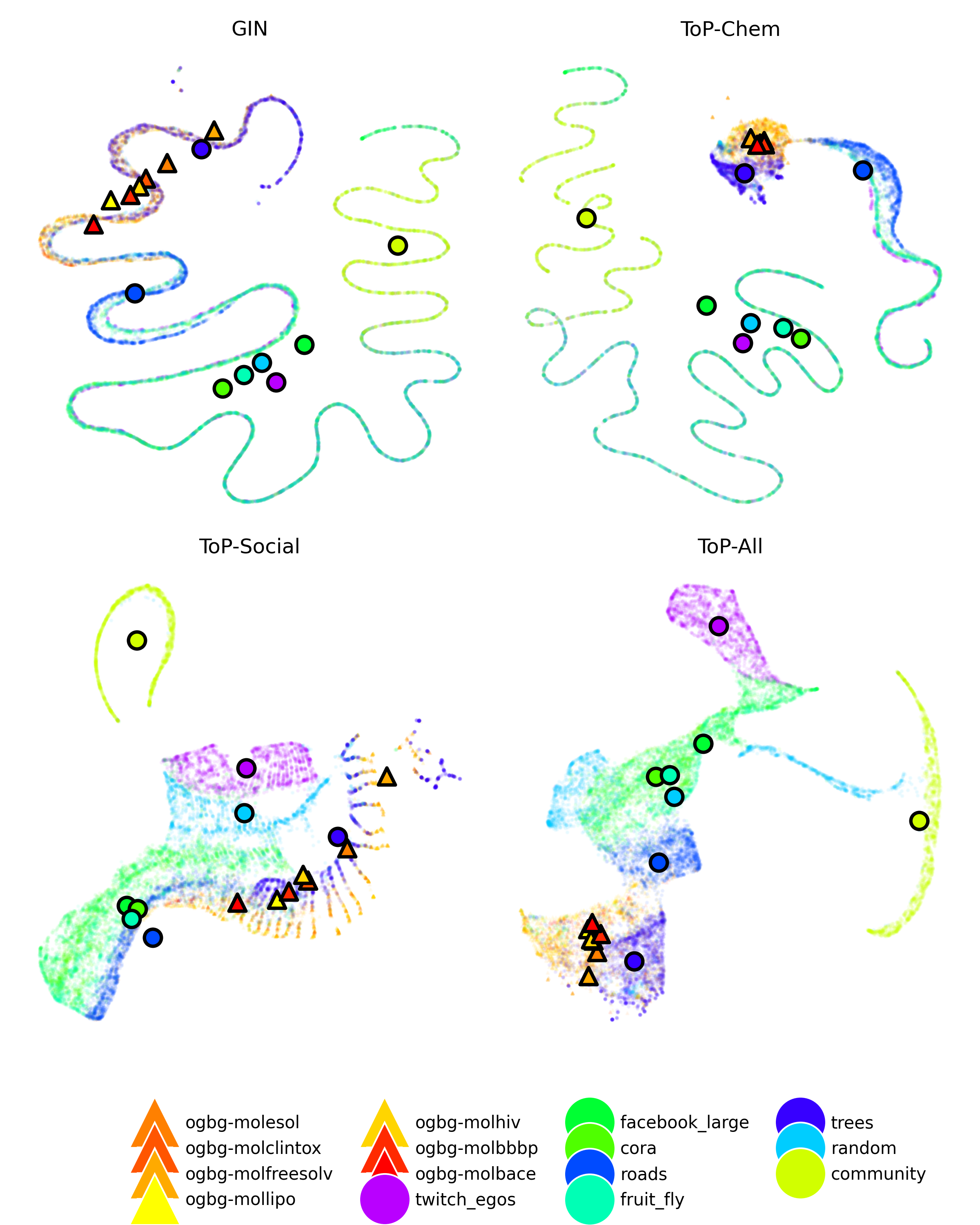}

\caption{UMAP embeddings of encodings from each model, as well as an untrained GIN model. Triangular markers show molecular graphs, and circles non-molecules. We plot the centroid of each dataset for added clarity. Qualitatively the untrained and \method-Chem models are noticeably more fragmented than the \method-Social and \method-All models. In turn the \method-Social model is more fragmented than the \method-All model. 
}

\label{fig:all-embeddings}

\end{figure}

\section{Experiments}
\label{sec:experiments}

\aod{Mini-contents paragraph about experiments. Emphasise that noise-noise is novel?}

\subsection{Datasets for Evaluation}

Evaluating \method requires collecting datasets of graphs from multiple domains.
A portion of these datasets should have downstream tasks, the useful information for which relies on both structure and features, with varied weighting to one or the other.
To this end we curate the following datasets, with Table~\ref{tab:dataset-statistics} in the Appendix providing dataset statistics:

\paragraph{Pre-Training Only}

\begin{description}[leftmargin=0em,itemsep=0em]
\item[molpcba \cite{Hu2020StrategiesNetworks,Wu2017MoleculeNet:Learning}] A curated dataset of \np{437929} small molecules, from which we use \np{250000}. Used only in pre-training.

\item[Cora \cite{McCallum2000AutomatingLearning}] A much used citation graph. Used only in pre-training.

\item[Fly Brain \cite{Winding2023TheBrain}] The full neural connectome of a fruit fly larvae. 
As the original multi-graph is dense at \np{548000} edges, we include a single edge between neurons if there are more than two synapses between them. Used only in pre-training.

\end{description}

\paragraph{Pre-Training and Evaluation}

\begin{description}[leftmargin=0em,itemsep=0em]

\item[Facebook \cite{Rozemberczki2021Multi-ScaleEmbedding}] A single graph of page-page connections on Facebook. Used both in pre-training and downstream transfer. On a graph level the downstream task is predicting average clustering, and we also evaluate transfer to node classification. This task does not use feature information.

\item[Twitch \cite{Rozemberczki2020KarateGraphs}] Ego networks from the streaming platform Twitch, used both in pre-training and downstream transfer, with a downstream binary classification task.  This task does not use feature information.

\item[Roads \cite{Leskovec2009CommunityClusters}] The road network of Pennsylvania. Used both in pre-training and downstream transfer, with the task set to be predicting graph diameter normalised by graph size. This task does not use feature information.

\end{description}

\paragraph{Evaluation Only}

\begin{description}[leftmargin=0em,itemsep=0em]

\item[Benchmark Molecules \cite{Hu2020StrategiesNetworks}] molesol, molfreesolv, mollipo, molclintox, molbbbp, molbace, molhiv.  Datasets sampled from MoleculeNet, with single-target classification or regression downstream tasks. These are used only in downstream transfer. These benchmarks have pre-prepared node and edge features, allowing us to measure the extent to which \method models can integrate feature information alongside structural. We source train/validation/test splits from the Open Graph Benchmark \cite{Hu2020OpenGraphs}.

\item[Trees] Trees, of a randomly sampled maximum depth, and a randomly fixed probability of branching at each level, with a downstream task of predicting depth. Used only in downstream evaluation. This task does not use feature information.

\item[Random] Erdos-Renyi (ER) graphs \cite{Erdos1960ONGRAPHS}, with a number of nodes and edge probability randomly sampled, and predicting edge probabilities as a downstream task.  Used only in downstream evaluation. This task does not use feature information.

\item[Community] Small ER subgraphs, with a set intra-subgraph and random inter-subgraph connection probability, with inter-subgraph connectivity as target.  Used only in downstream evaluation. This task does not use feature information.
\end{description}

\paragraph{Datasets from Single Graphs}

As the Facebook, Cora, Roads and Fly Brain datasets consist of a single graph, we employ exploration sampling to construct datasets of many smaller graphs.
We use a randomly selected exploration sampler for each sample \cite{Rozemberczki2020LittleSampling}, which in turn samples a random number of nodes in the range $24 \leq |V| \leq 96$ from the source graph.
The use of a selection of exploration samplers should further ensure that a large variety of graphs and structural patterns are included in the datasets.

\paragraph{Hyperparameters}
We take the downstream task on each validation set where available and add their relevant scores (MSE and $1-$ AUROC) with a linear model as a crude monitor for the models' representational abilities during pre-training.
\np{5000} samples are used for non-molecular datasets.
We conduct a limited Bayesian hyperparameter sweep.
The heuristic used is addition of evaluation scores on each dataset weighted by the number of samples in that dataset.
Following the results of the hyperparameter sweep, we train each model for 100 epochs with a batch size of 512, learning rates of \np{0.001}, embedding and projection dimensions of 300 and a regularisation constant of \np{0.2}.
This is very close to the default AD-GCL parameters.
Figure~\ref{fig:all-embeddings} shows UMAP embeddings of encodings of the validation datasets for our models.
Larger versions for the four \method models can be found in the Appendix.

\paragraph{Compute}
Pre-training was conducted on a workstation with an NVIDIA A6000, using around 16GB of VRAM capacity. 
Pre-training the \method-All model took a little more than eight hours.
Transfer results were produced on consumer laptops, primarily with an Apple M1 processor, and on GPU use around 3GB of video memory with batch sizes of 128.
\aod{Inference takes X seconds on GPU and Y seconds on CPU.}

\subsection{Transfer}

Here we apply our pre-trained models to various downstream tasks through transfer learning.
This follows stages of abstraction away from \method pre-training: first we fine tune on datasets without features, then on datasets with features included, and then compare \method models to reasonable benchmarks. 
Finally we report performance on node and edge-level tasks.

\begin{table}[ht]
\caption{Scores for fine-tuning on each validation dataset.
Molecular benchmarks here do not include node or edge features.
For the classification datasets we report AUROC and for regression datasets RMSE. \textbf{Bold} text indicates a result beats the supervised (GIN) baseline.
\underline{Underlined} text indicates a superior result, down to three significant figures.
Checkmarks \checkmark indicate that the \method-All model significantly out-performs the GIN baseline at $p < 0.01$.
`Best' indicates the proportion of datasets over which a model has superior performance, and `Positive' the proportion of datasets over which performance is superior to the fully-supervised baseline.
}

\centering
% \rotatebox{90}{
\begin{tabular}{l r r r r r p{0.02in}}
\toprule
& \multicolumn{1}{r}{GIN} & \multicolumn{1}{r}{\method-All (CL)} & \multicolumn{1}{r}{\method-Chem} & \multicolumn{1}{r}{\method-Social} & \multicolumn{1}{r}{\method-All} & \\ \midrule

Facebook & 0.31 $\pm$ 0.07 & \fbf 0.24 $\pm$ 0.06 & 1.17 $\pm$ 0.17 &  \fbf \underline{0.13 $\pm$ 0.00} &  \fbf \underline{0.13 $\pm$ 0.00} & \checkmark\\
Twitch Egos & 0.72 $\pm$ 0.01 & \fbf 0.74 $\pm$ 0.01 & 0.69 $\pm$ 0.02 & \fbf 0.74 $\pm$ 0.01 & \fbf  \underline{0.79 $\pm$ 0.01} & \checkmark\\
Roads & 0.46 $\pm$ 0.07 & \fbf 0.25 $\pm$ 0.03 & 0.57 $\pm$ 0.07 & \fbf  \underline{0.10 $\pm$ 0.00} & \fbf  \underline{0.10 $\pm$ 0.00} & \checkmark\\ \midrule
Trees & 0.20 $\pm$ 0.01 &\fbf  0.19 $\pm$ 0.02 & 0.24 $\pm$ 0.03 & \fbf  \underline{0.12 $\pm$ 0.00} & \fbf \underline{0.12 $\pm$ 0.00} & \checkmark\\
Community & 0.74 $\pm$ 0.11 & \fbf 0.31 $\pm$ 0.07 & \fbf 0.72 $\pm$ 0.13 & \fbf  0.02 $\pm$ 0.00 & \fbf  0.02 $\pm$ 0.00 & \checkmark\\
Random & 0.53 $\pm$ 0.23 & \fbf 0.29 $\pm$ 0.10 & 0.59 $\pm$ 0.09 & \fbf  \underline{0.04 $\pm$ 0.00} & \fbf  \underline{0.04 $\pm$ 0.00} & \checkmark\\

\midrule

molfreesolv & 4.98 $\pm$ 0.48 & \fbf 4.20 $\pm$ 0.10 &  \fbf 3.89 $\pm$ 0.21 &  \fbf \underline{3.83 $\pm$ 0.11} & \fbf 4.26 $\pm$ 0.03 & \checkmark\\
molesol     & 1.78 $\pm$ 0.09 & \fbf 1.57 $\pm$ 0.06 & \fbf 1.44 $\pm$ 0.09 & \fbf 1.47 $\pm$ 0.04 & \fbf  \underline{1.31 $\pm$ 0.03} & \checkmark\\
mollipo     & 1.13 $\pm$ 0.04 & \fbf 1.07$\pm$ 0.03 & \fbf 1.04 $\pm$ 0.02 & \fbf 0.99 $\pm$ 0.01 &  \fbf \underline{0.97 $\pm$ 0.01} & \checkmark\\
\midrule

% Classification
molclintox & 0.53 $\pm$ 0.04 & 0.46 $\pm$ 0.03 & 0.51 $\pm$ 0.03 &  0.50 $\pm$ 0.05 & \fbf  \underline{0.55 $\pm$ 0.11} & \\
molbbbp    & 0.57 $\pm$ 0.04 & \fbf  \underline{0.60 $\pm$ 0.03} & 0.54 $\pm$ 0.04 & \fbf  0.60 $\pm$ 0.03 &\fbf  0.59 $\pm$ 0.02 & \\
molbace    & 0.68 $\pm$ 0.05 & \fbf  \underline{0.76 $\pm$ 0.03} & 0.63 $\pm$ 0.13 & \fbf  0.75 $\pm$ 0.03 & \fbf 0.71 $\pm$ 0.02 & \\
molhiv     & 0.36 $\pm$ 0.06 & \fbf 0.50 $\pm$ 0.14 & \fbf 0.54 $\pm$ 0.13 &  \fbf 0.64 $\pm$ 0.05 & \fbf  \underline{0.66 $\pm$ 0.04} & \checkmark\\
\midrule

Best & 0\% & 15.4\% & 0.0\% & 38.5\% &  \underline{69.2\%} & \\
Positive & \shyphen & 92.3\% & 38.5\% & 92.3\% &  \underline{100\%} & \\

\bottomrule
\end{tabular}
\label{tab:full-transfer}
\end{table}

\subsubsection{Features Excluded}
We replace the output head with a simple two-layer MLP.
We then perform 10 fine-tuning runs of 50 epochs on each validation dataset, with and without features, for each pre-trained model.
Table~\ref{tab:full-transfer} presents results of transfer and fine-tuning without features.

The \method-All model consistently out-performs the other models under fine-tuning on the majority of datasets.
At $p \leq 0.01$ the \method-All model significantly out-performs the GIN baseline on 77\% of datasets.
When performance is not significantly better, it is never significantly worse.
The standard deviation of the results on each dataset are lower than those for the non-pretrained model and the domain-specific pre-trained models.
In Figure \ref{fig:training-examples} in the Appendix we show validation loss on the several datasets during fine-tuning for the \method-All model and an un-pre-trained GIN with the same architecture.
Here the advantages of pre-training are clear: the pre-trained model has consistently lower validation loss, seems to plateau later and at a lower loss, and has a lower deviation in that loss than the baseline GIN.

On our evaluation-only datasets (Trees, Community, Random), on which no models are pre-trained, we see markedly better performance from models that are pre-trained on a range of domains.
In particular the \method-All and \method-Social models drastically out-perform our supervised baseline and \method-Chem.
On the Trees dataset, which intuitively and on aggregate measures is the closest of the evaluation-only datasets to molecular data, \method-Chem also has negative transfer.

Pre-training only on molecules (`Chem' data subsets) results at-best on-par transfer compared to the supervised GIN baseline, with negative transfer on the majority of datasets.
Pre-training only on non-molecules, in comparison, results in positive transfer on the large majority of datasets and pre-training methods.
At $p \leq 0.01$ the \method-Social model significantly out-performs the \method-Chem model on the majority of datasets.
As before, where performance is not significantly better, it is never significantly worse.
This study corroborates the results presented in Table~\ref{tab:full-transfer}, with the Chem model consistently out-performed by non-molecular models.
We find that under pre-training with random edge dropping performance actually drops when including molecular data (`Social' vs `All').
This suggests that AD-GCL is able to differentiate between molecules and non-molecules and applies different augmentations to these domains. 
Nonetheless, we show that using a variety of non-molecular graphs in-general provides a more solid pre-training foundation than including molecules for both methods of augmentation.

\subsubsection{Features Included}
We replace the input head of our model with a three layer Multi-Layer Perceptron (MLP) for both node and edge features, then perform fine-tuning runs on our molecular datasets.
Table~\ref{tab:full-transfer-features-gin} presents the results of this transfer and fine-tuning.
These tasks are assumed to require node and edge features to obtain good performance, and here positive transfer rates are similar to when no features are included. %, despite \method's exclusion of features during pre-training.
This shows that the focus on structure during \method pre-training does not inhibit the use of feature-based information in downstream transfer.
The \method-All and \method-Social models again demonstrate consistent positive transfer, and on most datasets their performance improves compared to when features were excluded.
The same results are presented for different GNN backbones in Table~\ref{tab:chem-backbones-features} in the Appendix, but we perform no optimisation for these models, using hyperparameters from the GIN models here for the GAT and GCN models presented.
As such these results are included only for completeness.

\begin{table}[ht]
% \small
\caption{Scores for fine-tuning pre-trained GIN models (and a non-pre-trained baseline) on the molecular evaluation datasets with node and edge features included via projection heads. Results are reported in the same manner as in Table \ref{tab:full-transfer}.
}

\centering

\begin{tabular}{l r r r r r p{0.02in}}
\toprule

& \multicolumn{1}{r}{GIN} & \multicolumn{1}{r}{\method-All (CL)} & \multicolumn{1}{r}{\method-Chem} & \multicolumn{1}{r}{\method-Social} & \multicolumn{1}{r}{\method-All} & \\ \midrule

% Regression
molfreesolv & 4.00 $\pm$ 0.43 & 4.00 $\pm$ 0.24  & 4.16 $\pm$ 0.29 & \fbf 3.61 $\pm$ 0.43 & \fbf \underline{3.01 $\pm$ 0.59} & \checkmark\\
molesol     & 1.72 $\pm$ 0.25 & \fbf 1.18 $\pm$ 0.09 & \fbf 1.62 $\pm$ 0.31 & \fbf 1.03 $\pm$ 0.17 & \fbf \underline{0.90 $\pm$ 0.04} & \checkmark\\
mollipo     & 1.09 $\pm$ 0.05 & \fbf 0.99 $\pm$ 0.05 & 1.09 $\pm$ 0.06 & \fbf 0.80 $\pm$ 0.02 & \fbf \underline{0.79 $\pm$ 0.02} & \checkmark\\
\midrule

% Classification
molclintox & 0.65 $\pm$ 0.12 & 0.56 $\pm$ 0.16  & 0.52 $\pm$ 0.03 & \fbf \underline{0.86 $\pm$ 0.02} & \fbf 0.84 $\pm$ 0.05 & \checkmark \\ 
molbbbp    & 0.60 $\pm$ 0.03 & \fbf 0.63 $\pm$ 0.04  & \fbf 0.61 $\pm$ 0.03 & \fbf 0.62 $\pm$ 0.03 & \fbf \underline{0.67 $\pm$ 0.02} & \checkmark\\
molbace    & \underline{0.75 $\pm$ 0.03} &  0.72 $\pm$ 0.05 & 0.70 $\pm$ 0.07 & 0.70 $\pm$ 0.05 & 0.72 $\pm$ 0.05 & \\
molhiv     & 0.32 $\pm$ 0.03 & \fbf 0.46 $\pm$ 0.14 & \fbf 0.37 $\pm$ 0.07 & \fbf \underline{0.62 $\pm$ 0.11} & \fbf 0.57 $\pm$ 0.13 & \checkmark\\

\midrule
Best & 14.3\% & 0.0\% & 0.0\% & 28.6\% &  \underline{57.1\%} & \\
Positive & \shyphen & 57.1\% & 42.9\% & \underline{85.7\%} &  \underline{85.7\%} & \\

\bottomrule
\end{tabular}
\label{tab:full-transfer-features-gin}
\end{table}

\revision{
\subsection{Comparison to Other Methods}
}

\revision{
The selection of valid benchmarks from other works is limited.
We start by splitting existing works into two broad categories, generalists that function on any graph domain, and specialists that are designed for a specific domain of graph data.
The selection of specialist graph models in chemistry is broad, but not necessarily a fair comparison, as these works do not share the same fundamental aims as \method.
They are something of a best-case for pre-training: abundant pre-training data, without missing features.
In the following sections we include specialist GNNs, all pre-trained with constrastive learning, as well as specialist LLMs designed for molecular tasks.
We don't expect \method models to out-perform these specialists.
Instead, we include these results to evaluate how our multi-domain pre-training framework compares to the current state of the art within a domain.
}

\revision{
The selection of generalist models is far more limited.
GCC is, to the best of the author's knowledge, the only other generalist GNN work that does not rely on some level of feature transformation using LLMs.
This carries the significant caveat that the original work completely eliminates features, and the information they carry, from downstream transfer.
We extend the original GCC work somewhat, swapping the input and output heads as with our own models, in order to provide fair comparison against \method.
}

\revision{
The use of LLMs to unify the feature input space is we argue, a level of specialism in-and-of itself, as these datasets cannot be immediately applied to new domains, instead requiring hand-design of a textual schema.
LLMs themselves have been argued to be generalists \cite{Mao2024Position:Here}, in that any graph structure can function as input given some encoding schema.
}

% It bears repeating, however, that \method is the 
\revision{
\subsubsection{Molecular Tasks}
In Table~\ref{tab:comparison-results} we compare \method to other pre-training schemes across molecular tasks.
Results are sourced from the original works.
These methods are divided into specialist (domain-specific) and generalist (domain-agnostic) approaches.
For domain-specific approaches we include InfoGraph \cite{Xu2021Infogcl:Learning}, GraphCL \cite{Hassani2020ContrastiveGraphs} and AD-GCL \cite{Suresh2021AdversarialLearning}, contrastive learning approaches.
These models are pre-trained with all features included, and hyper-parameters optimised, over Zinc, a dataset of around 2M molecules.
We also include MolXPT \cite{Liu2023MolXPT:Pre-training}, InstructMol \cite{Cao2023InstructMol:Discovery} and ChemDFM \cite{Zhao2024ChemDFM:Chemistry}, LLM-based approaches.
}

\begin{table*}[ht]
% \small
\caption{Our results for the \method-All and \method-Social models presented alongside the same results from other pre-training works. 
Alongside dataset names we include the total size of the dataset.
Results are ROC-AUC, and other than for \method and GCC, taken verbatim from the respective work. 
Where standard deviations or errors are not present, they were also not present in the respective work.
The table is organised by specialism and type (Specialist GNN, Specialist with LLM, Generalist LLM, Generalist GNN).
($n$B) indicates $x$ billion parameters.
Non-LLMs are GINs with 6 layers and in the region of 7 million parameters.
\textbf{Bold} text indicates a superior result within a type of model.
}

\centering
\small
\begin{tabular}{l r r r r}
\toprule

& clintox (1329) & bbbp (1835) & bace (1361) & hiv (37014) \\ \midrule

InfoGraph (2.4M)     & \multirow{2}{*}{0.70 $\pm$ 0.03}      & \multirow{2}{*}{0.69 $\pm$ 0.01}       & \multirow{2}{*}{0.76 $\pm$ 0.02}       & \multirow{2}{*}{0.76 $\pm$ 0.02} \\ 
\footnotesize \cite{Xu2021Infogcl:Learning} &  \\
GraphCL (2.4M)  & \multirow{2}{*}{0.76 $\pm$ 0.03}      &  \multirow{2}{*}{\fbf 0.70 $\pm$ 0.01} & \multirow{2}{*}{0.75 $\pm$ 0.01}       &  \multirow{2}{*}{\fbf 0.78 $\pm$ 0.01} \\
\tiny \cite{Hassani2020ContrastiveGraphs} & \\
AD-GCL (2.4M)  & \multirow{2}{*}{\fbf 0.80 $\pm$ 0.04} &  \multirow{2}{*}{\fbf 0.70 $\pm$ 0.01} &  \multirow{2}{*}{\fbf 0.79 $\pm$ 0.01} &  \multirow{2}{*}{\fbf 0.78 $\pm$ 0.01} \\
\footnotesize \cite{Suresh2021AdversarialLearning} & \\

\midrule

% & clintox & bbbp & bace & hiv \\ \midrule
MolXPT  (350M)      &   \multirow{2}{*}{\fbf 0.95 $\pm$ 0.00}    &   \multirow{2}{*}{\fbf 0.80 $\pm$ 0.01}    &   \multirow{2}{*}{\fbf 0.88 $\pm$ 0.01}    &   \multirow{2}{*}{\fbf 0.78 $\pm$ 0.00}    \\
\footnotesize \cite{Liu2023MolXPT:Pre-training}  & \\
InstructMol  (7B) &                      &  \multirow{2}{*}{0.70}                &  \multirow{2}{*}{0.82}                &  \multirow{2}{*}{0.69}     \\
\footnotesize \cite{Cao2023InstructMol:Discovery}  & \\
ChemDFM (13B)        &  \multirow{2}{*}{ 0.90}               &   \multirow{2}{*}{0.67}               &   \multirow{2}{*}{0.78}               &   \multirow{2}{*}{0.74}    \\
\footnotesize \cite{Zhao2024ChemDFM:Chemistry}  & \\
% UniMol $\dagger$ \cite{Zhou2023Uni-Mol:Framework}             &  0.92 $\pm$ 0.02   &  0.73 $\pm$ 0.01   &  0.86 $\pm$ 0.00  & 0.81  $\pm$ 0.00   \\

\midrule
% & clintox & bbbp & bace & hiv \\ \midrule
GPT-4  ($>$200B\footnote{Best-guess speculation, as OpenAI is closed-source})                                       &   \multirow{2}{*}{0.52}         &   \multirow{2}{*}{\fbf 0.62}    &   \multirow{2}{*}{0.63}    &  \multirow{2}{*}{0.66}    \\
\footnotesize \cite{OpenAI2023GPT-4Report} & \\
LLaMa-2 (13B)                                  &   \multirow{2}{*}{0.46}         &   \multirow{2}{*}{0.60}         &   \multirow{2}{*}{0.26}    &   \multirow{2}{*}{0.29}    \\
\footnotesize \cite{Touvron2023LlamaModels} & \\
Galactica (30B)  $\dagger$     &   \multirow{2}{*}{\fbf 0.82}    &   \multirow{2}{*}{0.60}         &   \multirow{2}{*}{\fbf 0.73}    &  \multirow{2}{*}{\fbf 0.76}    \\
\footnotesize \cite{Taylor2022Galactica:Science} & \\

\midrule
GCC (untrained) (200k)      &  0.50 $\pm$ 0.03      & 0.59 $\pm$ 0.01      & 0.44 $\pm$ 0.04      & 0.46 $\pm$ 0.01 \\
GCC  (200k)      &  0.40 $\pm$ 0.07      & 0.52 $\pm$ 0.02      & 0.50 $\pm$ 0.04      & 0.52 $\pm$ 0.01 \\
GCC+Feat (untrained)     &  0.79 $\pm$ 0.04      & 0.64 $\pm$ 0.02      & 0.70 $\pm$ 0.05      & \fbf  0.70 $\pm$ 0.03 \\
GCC+Feat &  0.50 $\pm$ 0.04      & 0.61 $\pm$ 0.03      & 0.70 $\pm$ 0.1      & 0.68 $\pm$ 0.05 \\
\method-\textbf{Social} (2.4M)                   &  \fbf 0.86 $\pm$ 0.02 & 0.62 $\pm$ 0.03      & 0.70 $\pm$ 0.05      &0.62 $\pm$ 0.11 \\
\method-\textbf{All}                      &  0.84 $\pm$ 0.05      & \fbf 0.67 $\pm$ 0.02 & \fbf 0.72 $\pm$ 0.05 & 0.57 $\pm$ 0.13 \\

\bottomrule
\end{tabular}
\label{tab:comparison-results}
\end{table*}

\revision{
We also include results from other domain-agnostic graph models. 
For LLM-based approaches these chemical tasks these results are sourced the ChemDFM work \cite{Zhao2024ChemDFM:Chemistry}.
Performing molecular tasks with these models relies on the same assumptions as the chemistry-specific models, but without fine-tuning specifically on chemical literature\footnote{Galactica is trained on scientific literature, including but not limited to chemistry}.
}

\revision{
Finally we include GCC, to our knowledge the only other framework or approach for producing generalised graph models without relying on LLMs, albeit at the cost of feature inclusion for downstream tasks.
We fine-tune the two pre-trained encoders packaged with GCC\footnote{Our GCC version, updated to current packages, is available at \url{https://github.com/alexodavies/GCC-Updated}}.
Given that GCC does not provide a method for re-introducing domain features, we take the same approach as with our \method models, swapping the projection heads for nodes and edges (GCC+Feat).
We also include GCC with positional encodings as features, as in the original work (GCC).
Alongside the pre-trained GCC models we also include results for an equivalent but un-pre-trained model.
Where applicable we use the same hyper-parameters as for our fine-tuning of \method models.
In practise this is a batch size of \np{512} and a learning rate of \np{0.001}.
}

\revision{
We find that AD-GCL performs most strongly among the specialist GNN models, MolXPT strongest among the specialist LLMs, Galactica among the generalist LLMs, and \method-All among the generalist GNNs.
Among specialists MolXPT performs most strongly.
Among generalists Galactica and our \method models perform comparably, though both strongly out-perform other models in both classes.
}

\revision{
GCC models show consistent negative transfer, both when using their original positional encoding features and when we apply the \method process for swapping input heads.
On the much larger molhiv dataset the smaller (200k parameters vs 7M) GCC models reach greater performance within our 50 epoch fine-tuning runs.
Performance from GCC models is consistently better when we apply techniques from this work to include specialist features in place of positional encodings using the \method approach.
As the GCC encoders are much smaller than out \method encoders, we suspect that smaller models trained with \method would reach comparable performance.
}

\revision{
\subsubsection{Non-Molecular Tasks}
}

\revision{
We apply two recent LLMs, ChatGPT-4o-mini \cite{OpenAI2023GPT-4Report}  and Llama-3.2-Instruct (3B) \cite{Dubey2024TheModels}, across our validation-only datasets (random, community, trees), as well as the Twitch Egos dataset.
We use these smaller versions from their respecitve LLM families to reduce computational cost, which is still drastically higher than our GNN-only approach.
From here we denote Llama-3.2-Instruct (3B) as Llama-3.
% We also include the specialist chemical model ChemDFM \cite{Zhao2024ChemDFM:Chemistry} as an exploration into whether specialist graph learning in LLMs transfers well to new domains.
These tasks are chosen as they do not rely on node or edge features, meaning that learning and performance directly follows the use of structural information.
We also include GCC with positional encodings as features, as in the original work.
}

\revision{
\paragraph{LLM Prompting}
Using LLMs for arbitrary graph structures requires defining a schema for natural language representations of graphs.
We opt for the `incident' encoding style from \citet{Fatemi2023TalkModels}.
In this format a graph is encoded in the style \textit{nodes: $(1, 2, 3, ...)$ edges: $(1\leftrightarrow2, 2\leftrightarrow3, ...)$}.
We define prompts for each dataset, aimed to both provide contextual information and make tasks clear, which can be found in Listing~\ref{lst:prompts} in the Appendix.
We include further prompting text in an attempt to ensure uniformity across responses:
\textit{`Return a number, with no other text or filler. Do not explain your working. Answer in the format THE ANSWER IS X.'}
The prompt structure is then \{task description + extra prompt information + graph information\}.
We show an example in Listing~\ref{lst:prompt}, and an example response from Llama-3 in the Appendix in Listing~\ref{lst:response}.
}

\begin{lstlisting}[frame=shadowbox, rulesepcolor=\color{MidnightBlue}, caption={An example prompt for LLMs on the Trees dataset.}, label={lst:prompt}] % alt: single

    question: This is a Random graph. What is the connection probability between two nodes?  
    Return a number, with no other text or filler. Do not explain your working. Answer in the format THE ANSWER IS X. 
    graph: nodes: 0, 1, 2, 3, ...
    edges: 0 <-> 23, 0 <-> 17, 1 <-> 5, 1 <-> 16, ...
\end{lstlisting}

\begin{table*}[ht]
% \small
\caption{Our results for the \method-All and \method-Social models presented alongside LLMs and GCC applied to structure-only datasets.
Random, Community and Trees are unseen data for all models.
Twitch Egos is AUC-ROC, other datasets are RMSE. 
Due to cost constraints we do not produce error margins for LLM models.
}

\centering
\small
\begin{tabular}{l r r r r}
\toprule

& Twitch Egos & Random & Community & Trees \\ \midrule

% & clintox & bbbp & bace & hiv \\ \midrule
GPT-4o-mini (8B\footnote{Best-guess speculation, as OpenAI is closed-source})                &   0.50    &   0.30    &   0.37    &   0.28    \\
LLaMa-3.2 (3B)                      &   0.46                        &   0.12                 &   0.10    &   0.42    \\

\midrule
GCC (untrained)                     &  0.66 $\pm$ 0.01              & \fbf 0.03 $\pm$ 0.00   & \fbf 0.02 $\pm$ 0.01 & \fbf 0.12 $\pm$ 0.01 \\
GCC (200k)                          &  0.66 $\pm$ 0.01              & \fbf 0.03 $\pm$ 0.00   & \fbf 0.02 $\pm$ 0.00 & \fbf 0.12 $\pm$ 0.00 \\

\midrule
\method-\textbf{Social} (7M)        &  0.74 $\pm$ 0.01              & 0.04 $\pm$ 0.00        & \fbf 0.02 $\pm$ 0.00 & \fbf 0.12 $\pm$ 0.00 \\
\method-\textbf{All}                &  \fbf 0.79 $\pm$ 0.01         & 0.04 $\pm$ 0.00        & \fbf 0.02 $\pm$ 0.00 & \fbf 0.12 $\pm$ 0.00 \\

\bottomrule
\end{tabular}
\label{tab:llm-results}
\end{table*}

\revision{
As can be seen in Listing~\ref{lst:response}, models frequently ignored instructions to return only a number, and instead generated large amounts of text.
To extract numbers from generated text, we design a simple process, shown in Appendix in Listing~\ref{lst:extraction}.
}

\paragraph{Findings}

\revision{
LLMs, across all tasks, perform worse than \method or GCC models.
This is despite three of our datasets consisting of basic counting tasks, meaning that they are possible to perform in natural language.
Larger LLMs might perform better, but given that these models are already many orders of magnitude larger than \method or GCC models, we view the LLMs used here as fair benchmarks.
On the three synthetic datasets (Random, Community, Trees), with their simple structure-based tasks, \method models and GCC reach essentially the same performance.
}

\revision{
That said, a non-pretrained version of the small GCC encoder produces almost exactly the same performance as the pretrained encoder.
This indicates that in this setting GCC pre-training has minimal positive transfer benefits.
Also worth noting is that the positional encodings GCC uses as node and edge features are explicitly useful for these tasks.
\method models reach the same performance with no such advantage.
}

\revision{
On the real-world Twitch Egos dataset \method models strongly out-perform the LLMs and GCC.
A caveatte is that \method-Social and \method-All both have graphs from this dataset in their pre-training data.
In Section~\ref{sec:varied-pretrain} we present results for \method models with the Twitch Ego networks excluded from pre-training data, which under fine-tuning had an ROC-AUC performance of 0.73 $\pm$ 0.03, still significantly better than GCC.
}

\subsubsection{Further Transfer Results} 

% \paragraph{Node \& Edge Transfer}
By excluding the output head (again see Figure~\ref{fig:input-output-head}) we are able to transfer \method pre-trained models onto node and edge level tasks.
Here we focus on edge prediction and node classification, and though \method pre-training does seem to offer some performance increase, the benefits are far smaller than on graph level tasks.
These results can be found in the Appendix in Table~\ref{tab:full-transfer-node-edge}.

Further results can be found in the Appendix in Section~\ref{subsec:further-results}: see Table~\ref{tab:linear-transfer} for transfer with a linear models, Tables~(\ref{tab:graphcl-node-transfer},\ref{tab:graphcl-edge-transfer}) for random GraphCL-like augmentations, Table~\ref{tab:chem-backbones-features} for \method with varied GNN backbones, and Table~\ref{tab:full-transfer-node-edge} for node and edge classification.

\subsection{Further Experiments}

Having demonstrated the efficacy of \method, we present two more experiments, designed to give some insight into why and how \method pre-training is effective.
The first ensures that our results supporting Hypothesis~2, that out-of-domain data can be more useful than in-domain, are not the result of artefacting.
The second investigates the extent to which \method, during downstream transfer, relies on feature information in comparison to structural information.

\subsubsection{Varied Pre-Training Data}
\label{sec:varied-pretrain}
\revision{
While the results we present are statistically significant, there remains the possibility of artefacting.
In particular it is feasible that a particular domain in the \method-Social pre-training data lead to positive transfer, not the overall diversity of data.
In order to study and eliminate that possibility we perform a limited ablation over the component dataset within the Social pre-training data.
We exclude one of the five pre-training datasets at a time, pre-train an encoder with \method, then fine-tune ten times across our validation datasets, under the same transfer settings as our main results.
These results are presented in Table~\ref{tab:ablation-transfer}.
}

\begin{table}[h]
\small
\caption{Scores for fine-tuning models pre-trained on non-molecules, each with one non-molecular pre-training dataset excluded. For the classification datasets we report AUROC and for regression datasets RMSE. \textbf{Bold} text indicates a result performs equally (or better) than the model pre-trained on molecules (\textbf{Chem}).
Molecular tasks include domain features in fine-tuning.
}

\centering

\begin{tabular}{l  r r r r r}
\toprule
                 & \multicolumn{5}{c}{\textbf{Social} (Name indicates excluded dataset)} \\
                  & Twitch           & Neurons& Cora             & Facebook         & Roads    \\ \midrule
% molfreesolv       & 3.89 $\pm$ 0.21 & 4.14 $\pm$ 0.07 & 4.09 $\pm$ 0.02 & 4.17 $\pm$ 0.03 & 3.97 $\pm$ 0.11 & 4.24 $\pm$ 0.16   \\
% molesol           & 1.44 $\pm$ 0.09 & \fbf 1.37 $\pm$ 0.08 & 1.64 $\pm$ 0.06 & 1.67 $\pm$ 0.04 & 1.58 $\pm$ 0.13 & 1.59 $\pm$ 0.07   \\
% mollipo           & 1.04 $\pm$ 0.02 & \fbf 1.00 $\pm$ 0.01 & \fbf 0.99 $\pm$ 0.01 & \fbf 1.01 $\pm$ 0.01 & \fbf 1.00 $\pm$ 0.01 & \fbf 1.01 $\pm$ 0.01   \\ \midrule

% molclintox        & 0.51 $\pm$ 0.03 & 0.48 $\pm$ 0.12 & 0.44 $\pm$ 0.03 & 0.43 $\pm$ 0.00 & 0.43 $\pm$ 0.01 & \fbf 0.54 $\pm$ 0.10   \\
% molbbbp           & 0.54 $\pm$ 0.04 & \fbf 0.56 $\pm$ 0.04 & 0.40 $\pm$ 0.02 & \fbf 0.60 $\pm$ 0.02 & \fbf 0.62 $\pm$ 0.01 & 0.51 $\pm$ 0.02   \\
% molbace           & 0.63 $\pm$ 0.13 & \fbf 0.73 $\pm$ 0.04 & 0.62 $\pm$ 0.03 & \fbf 0.71 $\pm$ 0.05 & \fbf 0.70 $\pm$ 0.04 & \fbf 0.69 $\pm$ 0.02   \\
% molhiv            & 0.54 $\pm$ 0.13 & 0.42 $\pm$ 0.06 & \fbf 0.65 $\pm$ 0.07 & \fbf 0.54 $\pm$ 0.15 & \fbf 0.66 $\pm$ 0.04 & \fbf 0.61 $\pm$ 0.06   \\ \midrule

Facebook           & \fbf 0.29 $\pm$ 0.02 & \fbf 0.14 $\pm$ 0.00 & \fbf 0.84 $\pm$ 0.01 & \fbf 0.14 $\pm$ 0.01 & \fbf 0.22 $\pm$ 0.06   \\
Twitch Egos        & \fbf 0.73 $\pm$ 0.03 & \fbf 0.74 $\pm$ 0.01 & \fbf 0.73 $\pm$ 0.01 & \fbf 0.74 $\pm$ 0.01 & \fbf 0.77 $\pm$ 0.01   \\
Roads              & \fbf 0.19 $\pm$ 0.02 & \fbf 0.11 $\pm$ 0.00 & \fbf 0.12 $\pm$ 0.01 & \fbf 0.11 $\pm$ 0.00 & \fbf 0.18 $\pm$ 0.01   \\
Trees              & \fbf 0.15 $\pm$ 0.01 & \fbf 0.12 $\pm$ 0.00 & \fbf 0.13 $\pm$ 0.01 & \fbf 0.12 $\pm$ 0.00 & \fbf 0.15 $\pm$ 0.01   \\
Community          & \fbf 0.28 $\pm$ 0.03 & 0\fbf .02 $\pm$ 0.00 & \fbf 0.07 $\pm$ 0.01 & \fbf 0.02 $\pm$ 0.01 & \fbf 0.22 $\pm$ 0.03   \\
Random             & \fbf 0.15 $\pm$ 0.01 & \fbf 0.05 $\pm$ 0.00 & \fbf 0.09 $\pm$ 0.02 & \fbf 0.04 $\pm$ 0.00 & \fbf 0.13 $\pm$ 0.02   \\ \midrule

% & \multicolumn{1}{r}{\textbf{GIN}-Features} & \multicolumn{1}{r}{\textbf{\method-All (CL)}-Features} & \multicolumn{1}{r}{\textbf{Chem}-Features} & \multicolumn{1}{r}{\textbf{Social}-Features} & \multicolumn{1}{r}{\textbf{All}-Features} & \\ \midrule

                 % & \multicolumn{5}{c}{\textbf{Social} with Features (Name indicates excluded dataset)} \\
                 %       & Twitch           & Neurons& Cora             & Facebook         & Roads    \\ \midrule
% Regression
molfreesolv  & \fbf 3.30 $\pm$ 0.32   & \fbf 3.90 $\pm$ 0.24 & \fbf 3.82  $\pm$ 0.31 & \fbf 3.67  $\pm$ 0.37 & \fbf 3.83 $\pm$ 0.38 \\
molesol      & \fbf 1.03 $\pm$ 0.08   & \fbf 1.03 $\pm$ 0.07 & \fbf 1.26  $\pm$ 0.16 & \fbf 1.13  $\pm$ 0.16 & \fbf 1.17 $\pm$ 0.10 \\
mollipo      & \fbf 0.93 $\pm$ 0.03   & \fbf 0.83 $\pm$ 0.03 & \fbf 0.92  $\pm$ 0.05 & \fbf 0.81  $\pm$ 0.03 & \fbf 0.94 $\pm$ 0.02  \\
\midrule

% Classification
molclintox  & 0.49 $\pm$ 0.03 & \fbf 0.76 $\pm$ 0.13 & \fbf 0.72 $\pm$ 0.10 & \fbf 0.76 $\pm$ 0.18 & \fbf 0.74 $\pm$ 0.06 \\ 
molbbbp     & \fbf 0.64 $\pm$ 0.02 & \fbf 0.64 $\pm$ 0.02 & 0.61 $\pm$ 0.03 & \fbf 0.62 $\pm$ 0.02 & \fbf 0.64 $\pm$ 0.03 \\
molbace     & \fbf 0.70 $\pm$ 0.04 & 0.69 $\pm$ 0.07 & 0.68 $\pm$ 0.05 & 0.67 $\pm$ 0.06 & \fbf 0.72 $\pm$ 0.05 \\
molhiv      & \fbf 0.41 $\pm$ 0.06 & \fbf 0.55 $\pm$ 0.11 & \fbf 0.48 $\pm$ 0.13 & \fbf 0.63 $\pm$ 0.07 & \fbf 0.52 $\pm$ 0.11 \\

\bottomrule
\end{tabular}
\label{tab:ablation-transfer}
\end{table}

\revision{
We find that while performance does in-general drop compared to \method-Social, it is still consistently higher than \method-Chem.
Some datasets do appear to have a larger benefit for transfer than others - for example, when the Twitch Ego networks are excluded, performance on the molclintox dataset drops to near-random.
It is worth noting that we do not scale-up the Social pre-training data when a dataset is included, so these ablation models are all pre-trained on 20\% less data than \method-Chem and \method-Social.
}

\subsubsection{Features vs. Structure}

% We show that in-general $I(y;E) \neq 0$, and that through pre-training how to use this information can be transferred between domains.

% On a given graph dataset we can assume that $I(y;X, E) > 0$, but also that there is some imbalance between $I(y;X)$ and $I(y;E)$, with three possible results:

% \begin{description}
%     \item[$I(y;X) > I(y;E)$] Features carry more useful information
%     \item[$I(y;E) > I(y;X)$] Topology carries more useful information
%     \item[$I(y;X) = I(y;E)$] There is an equal level of useful information in both topology and features
% \end{description}

% In some tasks $X$ will contain all useful information, $f(X, E) \sim f(X) \rightarrow y$.
% In others topology contains the useful information, $f(X, E) \sim f(E)   \rightarrow y$.
Assume some imbalance between features and structure, $I(y;X) \neq I(y;E)$.
Here we propose a way of investigating this imbalance between feature information and structure information.
By degrading the useful information in either (or both of) $X$ and $E$, the degree to which performance relies on one or the other should be apparent.
A caveat here is that degrading either will presumably also degrade the information from their interdependence.

Consider some destructive noising process $N_t(\cdot)$, that degrades the useful information in either $X$ or $E$.
$N_0(x) = x$, with no useful information after $N_T(\cdot)$.

From here edge noising is denoted $N_E = N_t(E)$, and node feature noising $N_X = N_t(X)$
Assume some imperfect model produces predictions $f(X, E) \rightarrow \hat{y}$.
Performance is some $h(y, \hat{y})$ between real and predicted values.
By sampling $h(y, f(N_X, N_E) \rightarrow \hat{y})$, the useful information in both $X$ and $E$, and whether performance requires both to be present, should be observable.
We apply this experiment over our molecular benchmarks.
Here $X$ includes both node and edge features.

% \paragraph{Feature noise - Bernoulli}
% \aod{Try shuffling (column-wise) noise instead, across the whole dataset}
% We use additive Bernoulli noise, essentially bit flipping, for the bit-features in our molecular datasets.
% We weight this bit-flipping according to the original data.
% For a given bit of state $x_0$, originally positive with probability $P(x = 1) = p_{d}$, and the current noise rate $p_t$, we have the probability for that bit being flipped in Equation~\ref{eqn:feature-bit-flipping}:

% \begin{equation}
%     \label{eqn:feature-bit-flipping}
%     P_t(flip) =
%     \begin{cases}
%         x_0 = 1 & p_t \cdot (1 - p_d) \\
%         x_0 = 0 & p_t \cdot p_d\\
%     \end{cases}
%     % P_t(flip | x_0 = 1) = p_t \cdot (1 - p_d), P(flip | x_0 = 0) = p_t \cdot p_d
% \end{equation}

\paragraph{Structure noise}
We use random edge removal/addition for structure noise.
At $t$, we remove $|E_r| = p_t \cdot |V|$ edges at random from the graph, $E' = E \setminus E_r$.
We then add an equal number of randomly chosen edges $|E_a| = p_t \cdot |V|$, for a final noised edgelist $E'' = E' \cup E_a$.

% Both are weighted appropriately, bit-flipping by the probability of each component being positive, and edge swapping by maintaining the density of the original graph.
This is very close to the procedure taken by graph diffusion models as forward noising processes \cite{Vignac2023DiGress:Generation}.
At $N_T(E)$, the structure is a random graph with the same density as the original.

\paragraph{Feature Noise - Permutation}

We employ additive Bernoulli noise across node features.
At $t$, we have a probability of permuting the entries across a feature $p_t$.
Let node (or edge features) across a whole dataset be $X \in \{0,1\}^{N \times D}$ with $N$ points and $D$ features.
At $t$ we randomly select $N_f = p_t \cdot D$ features to permute.
Each selected feature is then shuffled, across the whole dataset, removing the useful information from that feature that pertains to each sample, while maintaining the marginal distribution of the given feature.

\begin{sidewaysfigure}[hbtp]
    \centering
    % First row of images
    \begin{minipage}[t]{0.48\textwidth}
        \centering
        \subfigure[molesol]{\includegraphics[width=\linewidth]{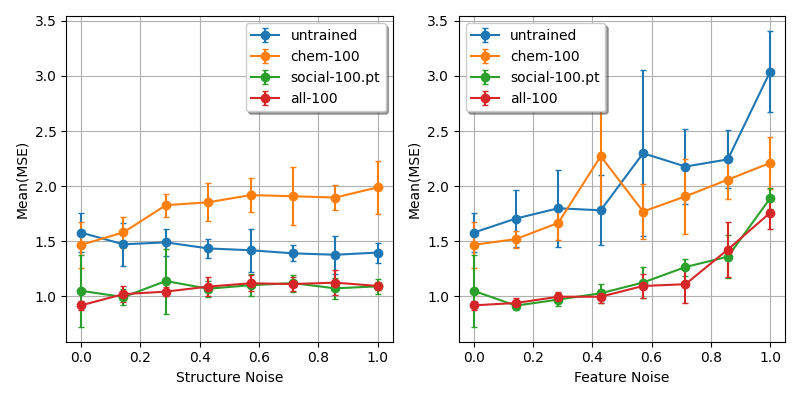} \label{fig:noise-noise-molesol}}
    \end{minipage}%
    \hfill
    \begin{minipage}[t]{0.48\textwidth}
        \centering
        \subfigure[molfreesolv]{\includegraphics[width=\linewidth]{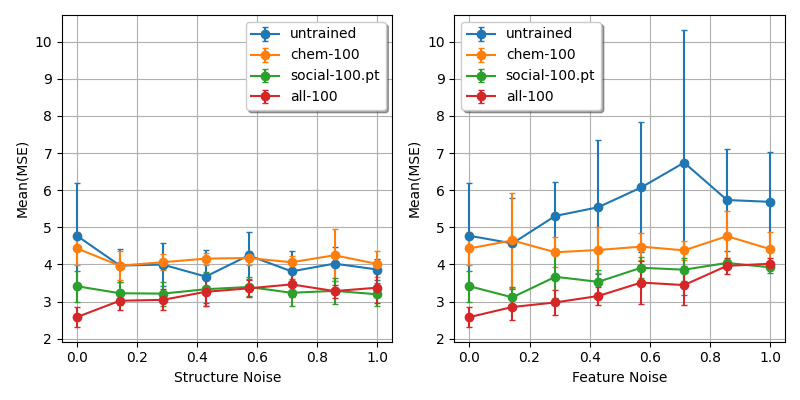} \label{fig:noise-noise-molfreesolv}}
    \end{minipage}%
    \vspace{0.3em}
    \begin{minipage}[t]{0.48\textwidth}
        \centering
        \subfigure[mollipo]{\includegraphics[width=\linewidth]{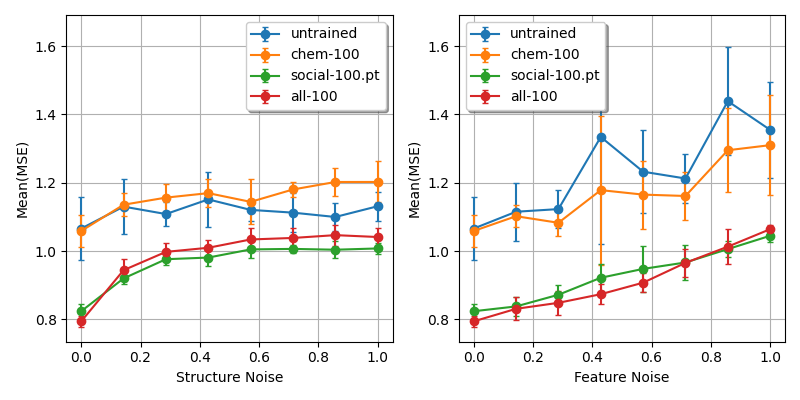} \label{fig:noise-noise-mollipo}}
    \end{minipage}%
    \hfill
    % Caption as the 8th entry
    \begin{minipage}[t]{0.48\textwidth}
        \centering
        \caption{Performance variation for each molecular regression benchmark dataset with increasing noise on structure and features. For each dataset, we benchmark performance for an untrained GIN and our three models pre-trained with \method{} (Chem-100, Social-100, All-100).}
        \label{fig:noise-noise-regression}
    \end{minipage}
\end{sidewaysfigure}

\begin{sidewaysfigure}[hbtp]
    \centering
    \begin{minipage}[t]{0.48\textwidth}
        \centering
        \subfigure[molbbbp]{\includegraphics[width=\linewidth]{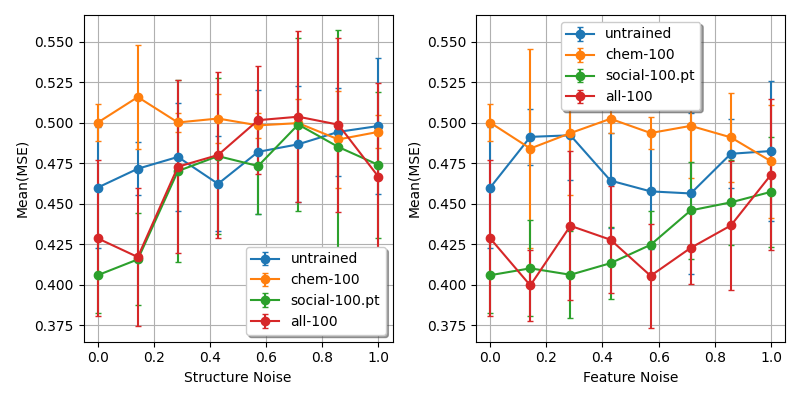} \label{fig:noise-noise-molbbbp}}
    \end{minipage}
    \hfill
    % Second row of images
    \begin{minipage}[t]{0.48\textwidth}
        \centering
        \subfigure[molclintox]{\includegraphics[width=\linewidth]{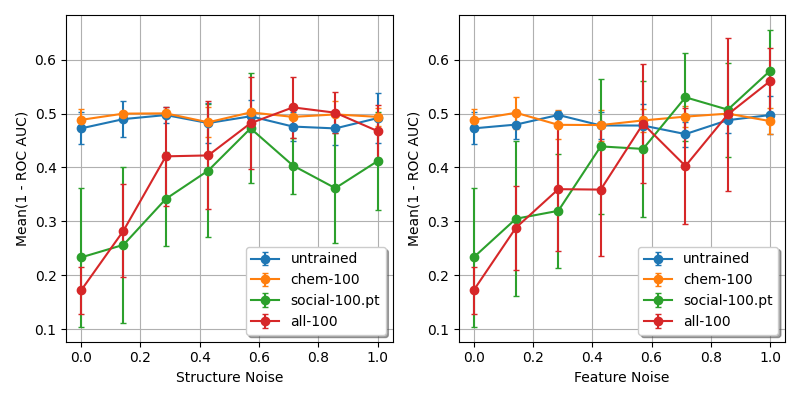} \label{fig:noise-noise-molclintox}}
    \end{minipage}%
    
    \vspace{0.3cm} % Space between rows
    
    \begin{minipage}[t]{0.48\textwidth}
        \centering
        \subfigure[molhiv]{\includegraphics[width=\linewidth]{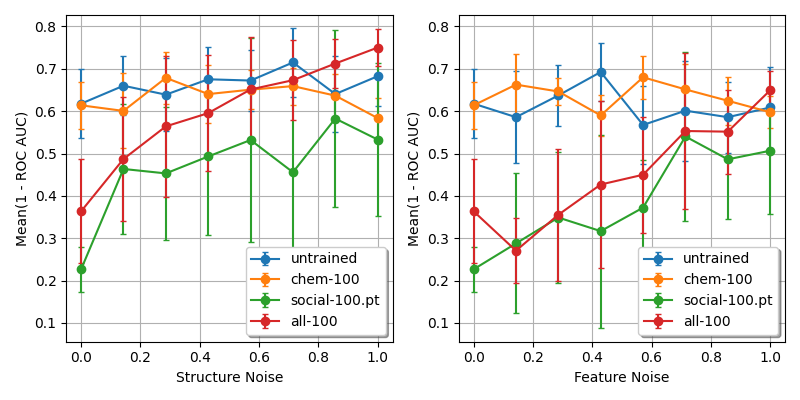} \label{fig:noise-noise-molhiv}}
    \end{minipage}%
    \hfill
    \begin{minipage}[t]{0.48\textwidth}
        \centering
        \subfigure[molbace]{\includegraphics[width=\linewidth]{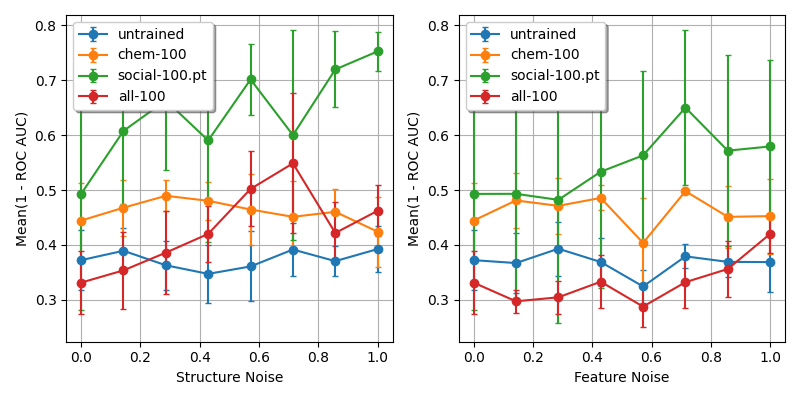} \label{fig:noise-noise-molbace}}
    \end{minipage}%
    \vspace{0.3cm}
    % Caption as the 8th entry
    \begin{minipage}[t]{0.96\textwidth}
        \centering
        \caption{Performance variation for each molecular classification benchmark dataset with increasing noise on structure and features. For each dataset, we benchmark performance for an untrained GIN and our three models pre-trained with \method{} (Chem-100, Social-100, All-100).}
        \label{fig:noise-noise-classification}
    \end{minipage}
\end{sidewaysfigure}

\paragraph{Results}
We fix the test set of each benchmark as un-noised, then perform 8 supervised training runs across 8 increasing noise levels on features and structure.
In contrast to our transfer results we perform these supervised training runs only for 25 epochs.
When noise is applied to structure, no noise is applied to features, and vice-versa.
We present the results in Figure~\ref{fig:noise-noise-regression} for regression datasets and Figure~\ref{fig:noise-noise-classification} for classification datasets.

Performance suffers by varying amounts across different datasets.
On most datasets the untrained GIN's response to structural noise is minimal, with little performance change even at complete noise in the graph structure.
The untrained GIN's reponse to feature noise is far larger, in particular on regression datasets (molesol, molfreesolv, mollipo).
The Chem model, pre-trained only on molecules, follows the same patterns as the untrained GIN model.

On most datasets the two non-molecular models, Social and All, show large performance drops as both structural and feature noise increases.
This is most clear on the datasets where \method pre-training was most effective, for example molclintox.
Interestingly this suggests that \method pre-training allows models not only to use structural patterns to greater effect, but also integrate information from features more usefully.
An additional possibility is that \method pre-training protects against overfitting to patterns in graph features, hence those features being used to greater effect.

% \begin{figure}[hbtp]
%     \centering
%     \subfigure[molesol]{\includegraphics[width=\linewidth]{images/noise-noise/performance_comparison-ogbg-molesol.png} \label{fig:noise-noise-molesol}}
%     \subfigure[molfreesolv]{\includegraphics[width=\linewidth]{images/noise-noise/performance_comparison-ogbg-molfreesolv.png} \label{fig:noise-noise-molfreesolv}}
%     \subfigure[mollipo]{\includegraphics[width=\linewidth]{images/noise-noise/performance_comparison-ogbg-mollipo.png} \label{fig:noise-noise-mollipo}}

%     \caption{} 
%     % We include the average scores and the best scores over the ten training runs on each dataset.}
%     \label{fig:noise-noise-regression}
% \end{figure}

\section{Discussion} \label{sec:discussion}

Here we discuss our findings in terms of our hypotheses.
This includes discussion on the performance of, and reasonable use cases for, models trained with \method.
Finally we detail explicitly assumptions we make and potential risks to the validity of this work.

\subsection{Findings}

Pre-training on multiple domains of topologies offers significant performance improvements on downstream tasks (Hypothesis~1).
With features excluded we find entirely positive transfer across a range of domains and graph-level tasks using adversarially pre-trained models with \method.
With features included, on popular molecular benchmarks, there are significant performance increases on graph-level tasks compared to fully supervised models.
This means, broadly, that \method can be used without risk of negative transfer.

Further, exposure to a larger variety out-of-domain samples can benefit performance.
In Table~\ref{tab:full-transfer}, assuming that the \method-Chem model has the least varied training data, then the \method-Social model, then the \method-All model, performance increases along the same order (Hypothesis~2).
Performance between the \method-Social and \method-All models is highly comparable, despite the \method-All model having double the training samples.
We expect, though we leave this as an area for future research, that a yet greater breadth of domains in the pre-training data would bring greater performance benefits.

\subsection{Intuitions and Comparisons}

The most intuitive assumption would be that exposing a molecular representation learner to non-molecular graphs would hinder its performance.
Instead the \method-Social and \method-All models out-perform the \method-Chem model on almost all molecular datasets under fine-tuning.
This holds true for varied pre-training regimes and model backbones (see Tables~\ref{tab:graphcl-node-transfer},~\ref{tab:graphcl-edge-transfer} and \ref{tab:chem-backbones-features} in the Appendix).
This shows potential for future work using \method.
Pre-training on graphs could be possible through data scale and diversity in-general, instead of within a specific domain.
This has clear parallels to the current trends in generalised pre-training in other fields \cite{Myers2024FoundationImpacts,Wang2023InternImage:Convolutions}.

Similarly one might expect that including node and edge features during supervised training of a non-pretrained model would likely lead to these domain-specific models out-performing our \method multi-domain models.
Our results instead show that performance is both superior and more consistent from our all-domain model on featured molecular graphs, with performance increasing compared to the same model on non-featured graphs.

In comparison to other generalist graph approaches, \method models perform strongly, including against science-aimed LLM models many orders of magnitude larger in both training data and cost of use.
This allows models, after using \method, to be transferred to arbitrary tasks on graph structures, without the inherent costs of LLM-based approaches.
% Against generalist GNN approaches, of which there are few, and none without significant caveattes on downstream applications, \method models prove more flexible and powerful.

We extend, using techniques from this work, GCC models to include domain features.
Here pre-trained GCC encoders show consistently negative transfer.
This indicates that GCC's use of positional encodings as node and edge features during pre-training actively inhibits their ability to use the information from domain features, which in most applications of graph models are essential for strong performance.
\method models, in contrast, show large performance increases when domain features are included during downstream fine-tuning compared to when domain features are not included.
Positive transfer is at least as common with these features included as when they are not.

Given these findings the reasonable use-cases for models pre-trained with \method are broad.
The potential benefit is two-fold.
Firstly, pre-training and fine-tuning requires less data for the downstream task to be viable.
This is particularly valuable in areas of data scarcity or expense, as well as offering presumed decreases in training time for the final model.
Secondly, as shown by our results in this work, our pre-trained models offer actual performance increase over simple supervised training.
Our assumption is that \method pre-training introduces a strong bias towards structural information, which in many applications is a positive characteristic, in-particular where it inhibits overfitting to patterns in the space of node or edge features.
% Our assumption is that the broad range of generic topological features learnt by the \method model provide a more robust foundation for training than simple random initialisation, so pre-trained models avoid overfitting to non-useful features better than randomly initialised models.
Guarantees on this point are left for future research.

\paragraph{Graph Foundation Models}
\citet{Liu2023TowardsBeyond} recently proposed definitions for, and expected properties of, graph foundation models.
The models produced with \method satisfy their broad definition, \fsl{`\ldots benefit from the pre-training of broad graph data, and can be adapted to a wide range of downstream graph tasks'}.
The two further properties proposed by \citet{Liu2023TowardsBeyond}, emergence and homogenization, are less easy to demonstrate.
Liu \etal themselves note that how these properties manifest is an open question, and therefore we similarly leave whether \method models exhibit these properties as an area for future research.

\revision{
A useful lens is to compare \method to the much more mature field of foundation models for natural language, ie LLMs.
While the current batch of LLMs are deeply expressive, and capable of performing a massive range of natural language tasks at a high level, earlier iterations of the same fundamental technology were more rudimentary.
The best parallel for use-cases of \method might be BERT \cite{Devlin2018BERT:Understanding}.
BERT is, by current standards, quite a small LLM, and normally requires fine-tuning to reach good performance on downstream tasks.
We're confident that future graph foundation models will be both more powerful and more flexible in terms of tasks, in particular zero-shot and few-shot learning, and leave whether \method models are an early version of those GFMs as an open question.
}

\subsection{Limitations}

In the full scope of pre-training and downstream tasks for graph data, this work has some limitations.
Firstly we use only static, non-hyper graphs, as in many other graph learning works.
We do not experiment with novel graph augmentations (or ``views''), instead employing only node and edge dropping, both random and adversarial.
We show that \method offers performance benefits with random views, and that adversarial views bolster these performance benefits.
While adversarial augmentations do improve pre-training, novel augmentations could be developed that include some techniques present in \citet{Hassani2020ContrastiveGraphs}, for example, edge addition alongside edge dropping.
Recent works  find that such augmentations, including combined additive and subtractive edge-based augmentations, can lead to more robust representation learners \cite{Guerranti2023OnMethods}.

Our method primarily focuses on graph-level tasks, in the same manner as the original AD-GCL work \cite{Suresh2021AdversarialLearning}.
As such, and again as in that original work, we find that performance benefits on node or edge level tasks are much smaller than on graph-level tasks.
For node and edge related tasks, future works can optimise \method through targeted augmentation schemes.
Similarly, we employ a limited set of domain compositions in evaluating \method.
A variety of augmentations, paired with a more detailed study of how different compositions of domain data influence representations, could bolster the expressivity of these topology-pre-trained models.

% Note that we have not compared \method to other graph pre-training methods where features are included.
% Such methods are, with few exceptions \cite{Liu2023OneTasks}, domain specific.
% As such comparisons would be between our generalised models and models pre-trained on a given domain with all features included.
% Such pre-training requires a large dataset of labelled samples, significantly limiting viable domains, and would not be fair benchmarks for \method.
% The Appendix includes a comparison of \method models with AD-GCL \cite{Suresh2021AdversarialLearning}, GraphCL \cite{Hassani2020ContrastiveGraphs} and InfoGraph \cite{Xu2021Infogcl:Learning}.

\section{Conclusion} \label{sec:conclusion}

We present \method, a method for generalised multi-domain pre-training of graph models.
\method relies on excluding node and edge features during pre-training.
Features can then be re-included during fine-tuning on downstream tasks.

Through contrastive learning, we use \method to present graph models capable of positive transfer across domains and tasks.
These are an important step towards graph foundation models with use-cases in the manner of models such as BERT \cite{Devlin2018BERT:Understanding}.
Like BERT, an early foundation language model, \method models can be fine-tuned with positive transfer on multiple domains and tasks.
On 75\% of downstream tasks from multiple domains we show significant ($p \leq 0.01$) positive transfer compared to a non-pre-trained model with the same architecture, as expected from our first hypothesis.
This is also true when node and edge features are re-introduced during fine-tuning.
Compared to a supervised baseline, where results are not significantly better, they are never significantly worse.
Additionally we showed that \method pre-training on a single target domain leads to at-best on-par, and on most tasks significantly worse ($p \leq 0.01$), performance compared to pre-training on multiple non-target domains.
This clearly demonstrates that the variety provided by out-of-domain pre-training can in this setting hold more value than in-domain pre-training, as in our second hypothesis. 

We anticipate that future work will focus on several key areas:
\begin{enumerate*}[label=\textbf{\arabic*})]
    \item Improved or tailored graph augmentation strategies.
    \item Investigation of domain weighting in dataset compositions.
    \item A larger pre-training dataset spanning a wider range of domains.
    \item Improved performance on node-level and edge-level tasks.
    \item Two-stage pre-training with an in-domain, features-included method.
    \item Integration and adaption for graph transformers
    % \item Integration with LLM-hybrid approaches like MolXPT
\end{enumerate*}

\backmatter

\bmhead{Supplementary information}

\revision{
All code for \method, including data processing, result-reproduction and pre-trained models is available in our  GitHub repository \url{https://github.com/alexodavies/general-gcl}.
Our updated code for GCC, including the inclusion of features in downstream tasks, is available in its own GitHub repository \url{https://github.com/alexodavies/GCC}.
}

% If your article has accompanying supplementary file/s please state so here. 

% Authors reporting data from electrophoretic gels and blots should supply the full unprocessed scans for key as part of their Supplementary information. This may be requested by the editorial team/s if it is missing.

% Please refer to Journal-level guidance for any specific requirements.

\bmhead{Acknowledgements}

AOD and RG acknowledge support from the UK Research and Innovation (UKRI) Centre for Doctoral Training in Interactive Artificial Intelligence Award (EP/S022937/1).

% \bibliographystyle{sn-apacite}

\bigskip
\begin{flushleft}%
Editorial Policies for:

\bigskip\noindent
Springer journals and proceedings: \url{https://www.springer.com/gp/editorial-policies}

\bigskip\noindent
Nature Portfolio journals: \url{https://www.nature.com/nature-research/editorial-policies}

\bigskip\noindent
\textit{Scientific Reports}: \url{https://www.nature.com/srep/journal-policies/editorial-policies}

\bigskip\noindent
BMC journals: \url{https://www.biomedcentral.com/getpublished/editorial-policies}
\end{flushleft}

% \clearpage
% \newpage

\begin{appendices}

\appendix
% \section{Appendix}
% \nsa{Was missing section}
% \nsa{Move refs before appendix}

% \newpage

\section{Dataset Statistics}

\begin{table*}[ht]
    \centering
    \caption{Statistics for each of our datasets. If used for training, we denote these statistics averaged across the training data. Otherwise we report statistics for their validation (downstream fine-tuning) datasets. All tasks are single-target.}
    \small
    \rotatebox{90}{
    \begin{tabular}{@{}lrrrrr@{~~~}l@{}}
        \toprule
        & \textbf{Num. Graphs} & \textbf{Number of Nodes} & \textbf{Number of Edges} & \textbf{Diameter} & \textbf{Avg. Clustering} & \textbf{Downstream Task}\\
        \midrule

        \multicolumn{7}{l}{\textbf{Training Only}}\\\midrule
        molpcba  & 250000 & 25.7 $\pm$ 6.28 & 27.7 $\pm$ 7& 13.5 $\pm$ 3.29 & 0.00128 $\pm$ 0.0115 & -- \\

        Cora  & 50000 & 59.3 $\pm$ 20.7 & 119$\pm$ 56.6 & 10.1 $\pm$ 4.68 & 0.322 $\pm$ 0.0845 &  --\\

        Fly Brain  & 50000 & 59.5 $\pm$ 20.8 & 146$\pm$ 99.6 & 8.8 $\pm$ 3.49 & 0.237 $\pm$ 0.0875 &  --\\\midrule
        %  & 5000 & 27.1 $\pm$ 8.77 & 29.6 $\pm$ 9.4 & 14 $\pm$ 3.75 & 0.00255 $\pm$ 0.0158 \\
        %  & 1000 & 27 $\pm$ 7.64 & 29.7 $\pm$ 8.43 & 13.8 $\pm$ 3.58 & 0.004 $\pm$ 0.0212 \\  

        \multicolumn{7}{l}{\textbf{Training and Evaluation}}\\\midrule
        Twitch  & 50000 & 29.6 $\pm$ 11.1 & 86.6 $\pm$ 70.7 & 2$\pm$ 0& 0.549 $\pm$ 0.149 &  Class. Games Played\\
        %  & 5000 & 30$\pm$ 11.2 & 88.4 $\pm$ 71.6 & 2$\pm$ 0& 0.55 $\pm$ 0.15 \\
        %  & 2000 & 29.8 $\pm$ 11& 87.7 $\pm$ 69.6 & 2$\pm$ 0& 0.552 $\pm$ 0.148 \\  

        Facebook  & 50000 & 59.5 $\pm$ 20.7 & 206 $\pm$ 170 & 10.2 $\pm$ 6.39 & 0.429 $\pm$ 0.133 &  Avg. Clustering\\
        %  & 2000 & 60$\pm$ 20.5 & 203$\pm$ 154& 10.3 $\pm$ 6.42 & 0.426 $\pm$ 0.132 \\
        %  & 5000 & 59.8 $\pm$ 20.8 & 206$\pm$ 167& 10.2 $\pm$ 6.35 & 0.429 $\pm$ 0.132 \\  

        %  & 5000 & 58.9 $\pm$ 20.6 & 117$\pm$ 55.4 & 10.2 $\pm$ 4.78 & 0.324 $\pm$ 0.0844 \\
        %  & 2000 & 58.3 $\pm$ 20.6 & 116$\pm$ 55.8 & 10.1 $\pm$ 4.75 & 0.324 $\pm$ 0.0857 \\  

        Roads  & 50000 & 59.5 $\pm$ 20.8 & 73.9 $\pm$ 28.2 & 16.4 $\pm$ 5.92 & 0.0559 $\pm$ 0.0426 &  Diameter / Num. Nodes\\
        %  & 5000 & 59.5 $\pm$ 20.6 & 73.7 $\pm$ 27.7 & 16.4 $\pm$ 6.12 & 0.0564 $\pm$ 0.0423 \\
        %  & 2000 & 59.6 $\pm$ 20.7 & 73.5 $\pm$ 27.6 & 16.5 $\pm$ 6.31 & 0.0568 $\pm$ 0.0421 \\  

        \midrule
        %  & 5000 & 59.7 $\pm$ 20.8 & 146$\pm$ 99.8 & 8.82 $\pm$ 3.5 & 0.236 $\pm$ 0.0873 \\
        %  & 2000 & 60.2 $\pm$ 20.7 & 148$\pm$ 103& 8.8 $\pm$ 3.45 & 0.235 $\pm$ 0.0849 \\  
        
        \multicolumn{7}{l}{\textbf{Evaluation Only}}\\\midrule
        
        % molesol  & 113 & 17.7 $\pm$ 6.4 & 19.4 $\pm$ 7.24 & 8.24 $\pm$ 3.1 & 0.00136 $\pm$ 0.0101 \\
        molesol & 1015& 12.7 $\pm$ 6.64& 12.9 $\pm$ 7.62& 6.81 $\pm$ 3.30& 0.000353 $\pm$ 0.00417 &  Water Solubility\\  
        %  & 113 & 18.8 $\pm$ 6.52 & 20.5 $\pm$ 7.25 & 8.86 $\pm$ 3.22 & 0.0124 $\pm$ 0.0642 \\
        
        molfreesolv  & 577& 8.46 $\pm$ 3.97& 8.02 $\pm$ 4.50& 5.02 $\pm$ 2.09& 0.0 $\pm$ 0.0 &  Solvation\\  
        %  & 65 & 11$\pm$ 5.22 & 11.7 $\pm$ 5.87 & 5.23 $\pm$ 2.34 & 0.0293 $\pm$ 0.137 \\       
      
        mollipo  & 3780& 27.0 $\pm$ 7.44& 29.4 $\pm$ 8.22& 13.8 $\pm$ 4.03& 0.00366 $\pm$ 0.017 &  Lipophilicity\\ 
        %  & 420 & 27.6 $\pm$ 7.62 & 30.4 $\pm$ 8.38 & 14$\pm$ 4.1 & 0.0066 $\pm$ 0.0231 \\  

        molclintox  & 1329& 26.3 $\pm$ 15.8& 28.0 $\pm$ 17.1& 12.4 $\pm$ 6.09& 0.00259 $\pm$ 0.0191 &  Class. Toxicity\\ 
        %  & 148 & 24.6 $\pm$ 13.4 & 26.7 $\pm$ 14.3 & 12$\pm$ 5.13 & 0.00868 $\pm$ 0.0555 \\  

        molbbbp & 1835& 23.5 $\pm$ 9.89& 25.4 $\pm$ 11.0& 11.2 $\pm$ 4.03& 0.00284 $\pm$ 0.0278 &  Class. Penetration\\          

        molbace & 1361& 34.0 $\pm$ 7.89& 36.8 $\pm$ 8.13& 15.2 $\pm$ 3.14& 0.00672 $\pm$ 0.0205 &  Class. Bioactivity\\   

        molhiv & 37014& 25.3 $\pm$ 12.0& 27.3 $\pm$ 13.1& 12.0 $\pm$ 5.15& 0.00158 $\pm$ 0.0156 &  Class. Inhibition\\   

        \midrule

        \multicolumn{7}{l}{\textbf{Evaluation Only, Unseen Domain}}\\\midrule
        
        Trees  & 5000 & 19.6 $\pm$ 6.96 & 18.6 $\pm$ 6.96 & 10.1 $\pm$ 3.11 & 0$\pm$ 0 &  Depth\\
        %  & 2000 & 19.5 $\pm$ 7.05 & 18.5 $\pm$ 7.05 & 9.97 $\pm$ 3.13 & 0$\pm$ 0\\     
        
        Random  & 5000 & 29.3 $\pm$ 10.7 & 111$\pm$ 82.1 & 3.88 $\pm$ 1.16 & 0.227 $\pm$ 0.0998 &  Density\\
        %  & 2000 & 29.2 $\pm$ 10.5 & 110$\pm$ 82& 3.91 $\pm$ 1.2 & 0.224 $\pm$ 0.0988 \\       
      
        Community  & 5000 & 48$\pm$ 0& 323$\pm$ 26.2 & 3$\pm$ 0.0346 & 0.408 $\pm$ 0.0256 &  Inter-Community Density\\
        %  & 2000 & 48$\pm$ 0& 323$\pm$ 26& 3$\pm$ 0.0316 & 0.408 $\pm$ 0.0256 \\
        \bottomrule
    \end{tabular}
    }
    \label{tab:dataset-statistics}
\end{table*}

\clearpage
\newpage

\section{Embedding Space}

% \begin{table}[ht]

% \begingroup

%     \includegraphics[width=\linewidth]{images/everyone.png}
%     \captionof{figure}{A UMAP embedding of encodings from each model, as well as an untrained model. The untrained and Chem models are noticeably more fragmented than the Social and All models. In turn the Social model is more fragmented than the All model. In the All model embedding, molecules (primarily dark blue) are all in the same region, furthest from the Twitch Ego networks (lilac), with the Cora (brown) and Facebook Page-Page data (pink) ``filling the gap''.}\label{fig:all-embeddings-large}

% \endgroup

% \end{table}

\begin{figure}[ht]

\centering
\includegraphics[width=\linewidth]{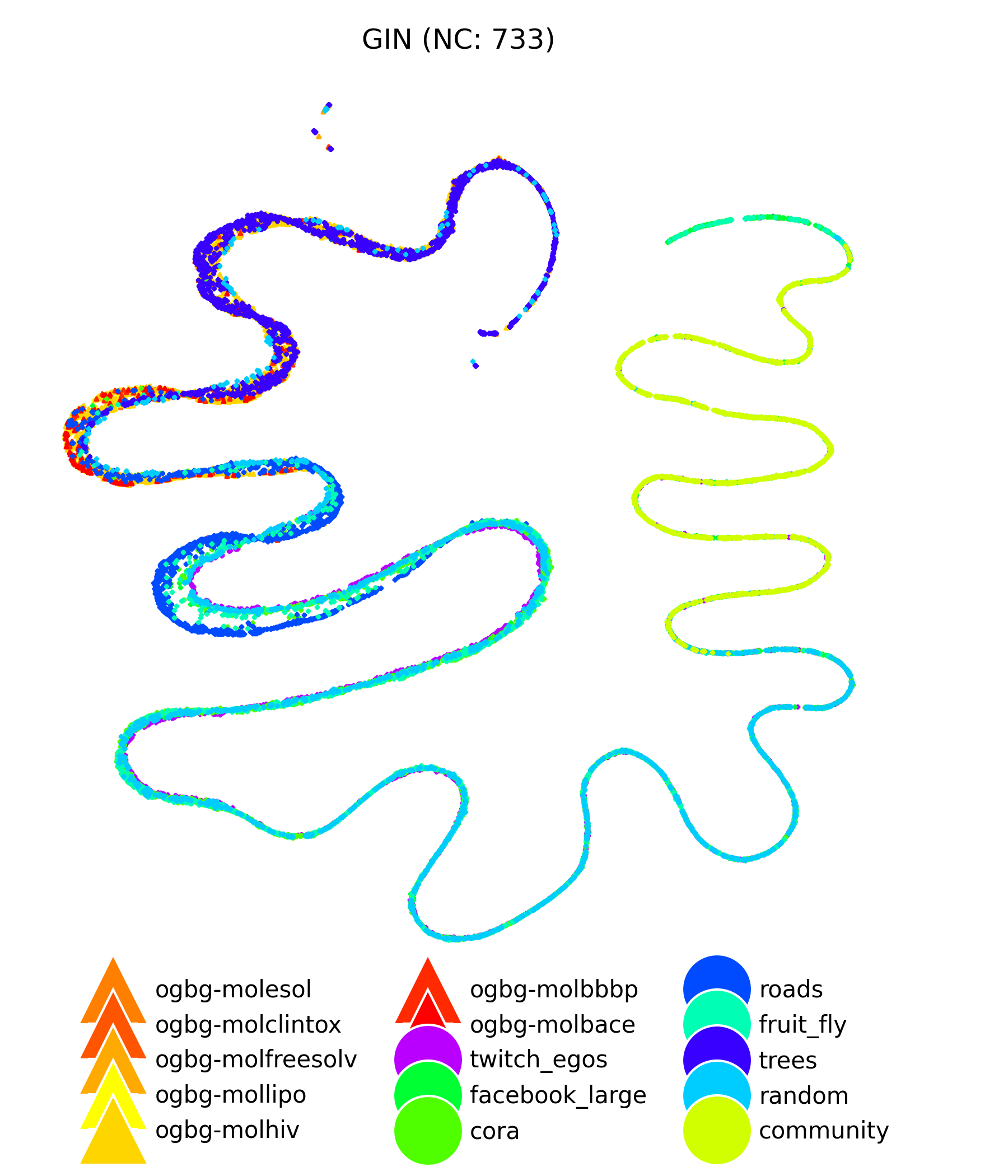}

\caption{A UMAP embedding of encodings from an untrained encoder.}

\label{fig:gin-embeddings}

\end{figure}

\begin{figure}[ht]

\centering
\includegraphics[width=\linewidth]{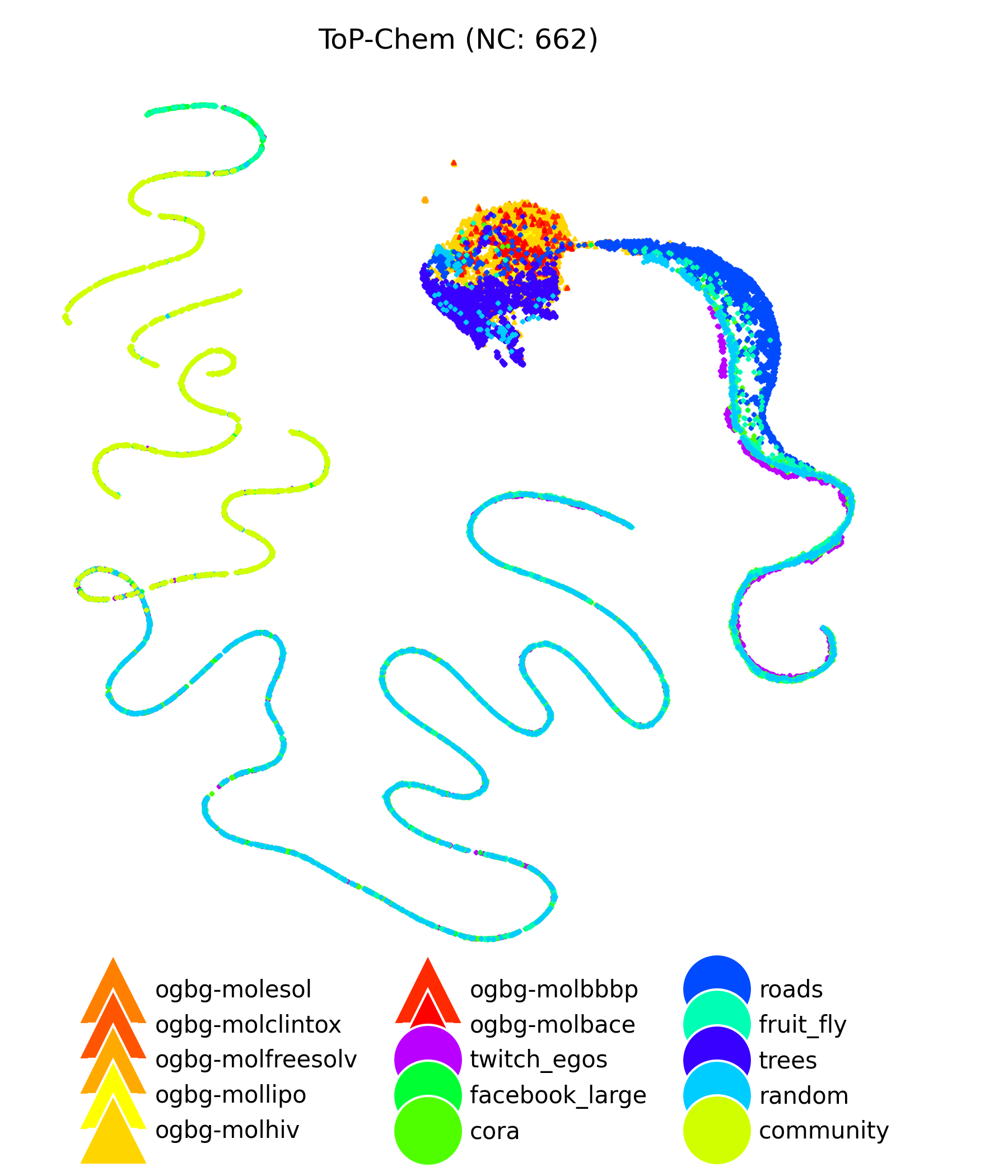}

\caption{A UMAP embedding of encodings from the \method-Chem model, pre-trained on only molecules.}

\label{fig:chem-embeddings}

\end{figure}

\begin{figure}[ht]

\centering
\includegraphics[width=\linewidth]{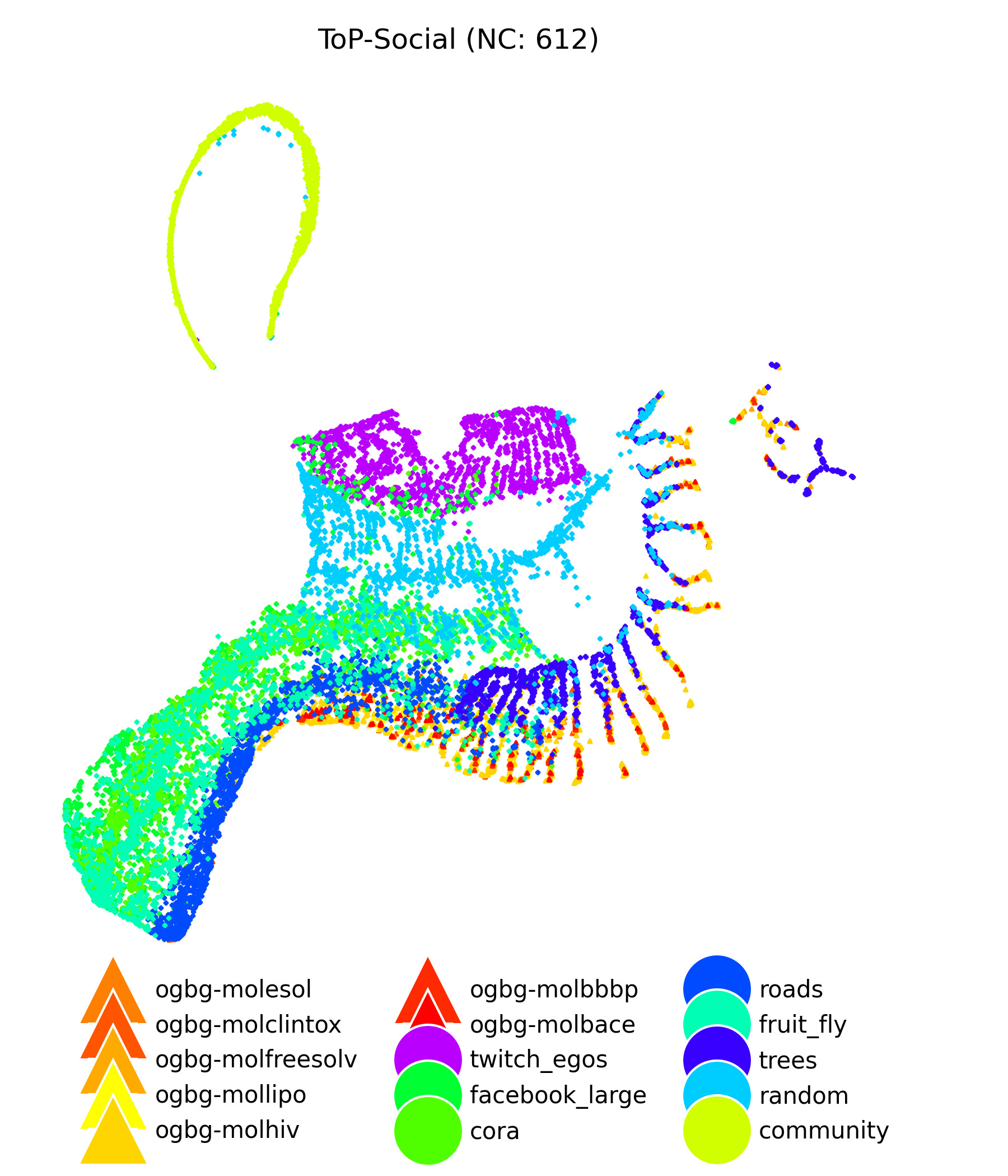}

\caption{A UMAP embedding of encodings from the \method-Social model, pre-trained on only non-molecules.}

\label{fig:social-embeddings}

\end{figure}

\begin{figure}[ht]

\centering
\includegraphics[width=\linewidth]{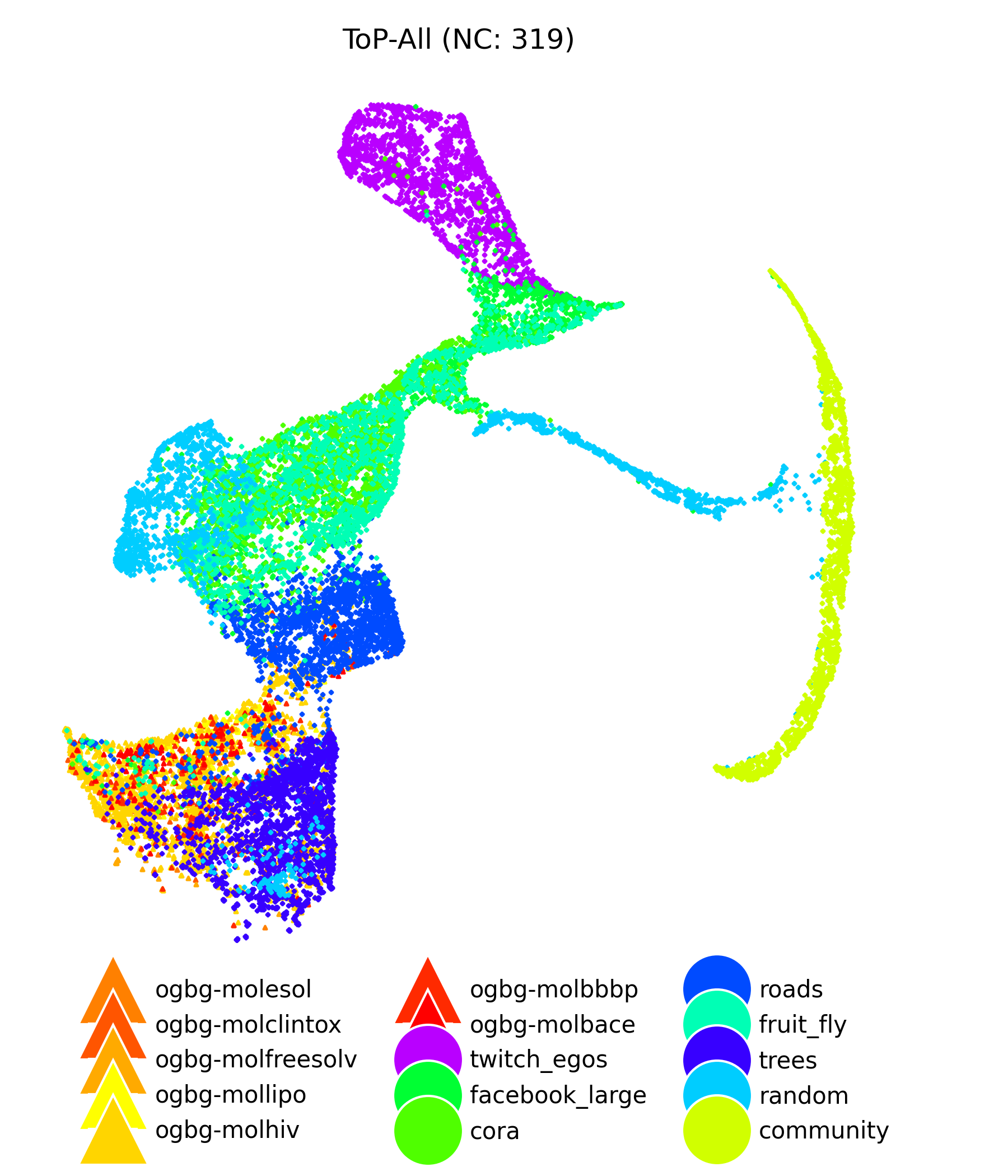}

\caption{A UMAP embedding of encodings from the \method-All model, pre-trained on the whole amalgamated dataset (i.e. molecules and non-molecules).}

\label{fig:all-embeddings-individual}

\end{figure}

\begin{table}[h]
\caption{
On the combined encodings of the validation sets we compute $R^2$ correlation coefficients between the first five principal components (PCA) and common graph-level metrics.
High correlations, or the lack thereof, should give some indication of how embedding components represent different graph characteristics.
% The results are presented, along with each component's explained variance, in Table~\ref{tab:component-correlations} and visualised in Figure~\ref{tab:pca-vs-metric}.
% PCA components, fit on the embeddings of the test set, and their most correlated graph-level metrics, for each of the trained models.
Here we present the most correlated metrics and components for each pre-trained model.
\textbf{\underline{Underlined}} text indicates a strong correlation $|R^2| \geq 0.66$, and \textbf{bold} text indicates moderate correlation $0.33 \leq |R^2| \leq 0.66$. We show up to the first five PCA components, but include only three where explained variance values are minimal.}
\label{tab:component-correlations}
\centering
    % \begin{tabular}{c||c|c|c|c}
    % \begin{tabular}{l clSlSlS}
    \begin{tabular}{lc lS[table-format=2.3] lS[table-format=2.3] lS[table-format=2.3]}
    \toprule
      & {\multirow{2}{*}{\makecell{Variance \\Ratio}}}  & \multicolumn{6}{c}{Metric Correlations} \\\cmidrule{3-8}
      &                                                 & {Most}        & {$R^2$}    & {Second}      & {$R^2$}    & {Third} & {$R^2$}\\\midrule
    \multicolumn{8}{l}{\textbf{Untrained}}\\\midrule
    PCA 0 & $9.96\times 10^{-01}$ & Edges & 0.219 & Transitivity & 0.072 & Density & 0.063 \\ 
    PCA 1 & $3.52\times 10^{-03}$ & Edges & \fbf -0.555 & Transitivity & -0.276 & Density & -0.243 \\ 
    PCA 2 & $1.13\times 10^{-04}$ & Edges & \fbf 0.39 & Transitivity & 0.224 & Density & 0.21 \\ 
    % PCA 3 & $1.37\times 10^{-05}$ & Number of Edges & \fbf 0.426 & Density & 0.32 & Transitivity & 0.285 \\ 
    % PCA 4 & $2.97\times 10^{-06}$ & Number of Edges & \fbf -0.343 & Density & \fbf -0.336 & Diameter & 0.239 \\ 

    \midrule
    \multicolumn{8}{l}{\textbf{Chem}} %&&&&&&&
    \\\midrule
    PCA 0 & $9.97\times 10^{-01}$ & Edges & 0.223 & Transitivity & 0.074 & Density & 0.065 \\ 
    PCA 1 & $2.60\times 10^{-03}$ & Edges & \fbf -0.514 & Transitivity & -0.253 & Density & -0.225 \\ 
    PCA 2 & $1.14\times 10^{-05}$ & Edges & 0.132 & Nodes & 0.051 & Transitivity & 0.047 \\ 
    % PCA 3 & $2.23\times 10^{-06}$ & Number of Edges & \fbf \underline{0.721} & Density & \fbf 0.575 & Transitivity & \fbf 0.46 \\ 
    % PCA 4 & $3.14\times 10^{-08}$ & Number of Edges & -0.05 & Average Clustering & -0.037 & Density & 0.037 \\ 

    \midrule
    \multicolumn{8}{l}{\textbf{Social}} %&&&&&&&
    \\\midrule
    PCA 0 & $9.99\times 10^{-01}$ & Edges & 0.127 & Transitivity & 0.032 & Density & 0.032 \\ 
    PCA 1 & $9.92\times 10^{-04}$ & Nodes & \fbf \underline{1.00} & Edges & \fbf 0.528 & Density & \fbf -0.393 \\ 
    PCA 2 & $4.79\times 10^{-08}$ & Edges & 0.133 & Density & 0.07 & Transitivity & 0.054 \\ 
    % PCA 3 & $9.03\times 10^{-10}$ & Diameter & -0.103 & Transitivity & 0.092 & Average Clustering & 0.087 \\ 
    % PCA 4 & $3.31\times 10^{-10}$ & Diameter & 0.221 & Density & -0.203 & Average Clustering & -0.184 \\ 
    
    \midrule
    \multicolumn{8}{l}{\textbf{All}} %&&&&&&&
    \\\midrule
    PCA 0 & $7.04\times 10^{-01}$ & Nodes & \fbf \underline{0.999} & Edges &  \fbf 0.534 & Density & \fbf -0.388 \\ 
    PCA 1 & $1.94\times 10^{-01}$ & Edges & \fbf \underline{0.805} & Density & \fbf \underline{0.737} & Diameter & \fbf -0.554 \\ 
    PCA 2 & $4.23\times 10^{-02}$ & Clustering & 0.309 & Transitivity & 0.248 & Edges & 0.235 \\ 
    PCA 3 & $2.43\times 10^{-02}$ & Diameter & \fbf 0.5 & Clustering & \fbf -0.416 & Transitivity & -0.243 \\ 
    PCA 4 & $1.82\times 10^{-02}$ & Density & -0.132 & Transitivity & -0.108 & Diameter & -0.088 \\ 
    
    \bottomrule
\end{tabular}

\end{table}

% \begin{figure}[h]
%     \centering
%     \includegraphics[width=\linewidth]{images/embedding-example-pca.png}
%     \caption{PCA projections of encodings from each model, as well as an untrained model. We scale axes by percentiles~-~of~-~data~-~included as for each model except the All model the projections have outliers in the region $10^7$.}
%     \label{fig:all-pca}
% \end{figure}

\begin{figure}[h]
    \centering
    \begin{tabular}{c c c c}
         & \hspace{2em} PCA 0  & \hspace{2em} PCA 1 & \hspace{2em} PCA 2\\
         
    \rotatebox[origin=lt]{90}{\hspace{3em}Untrained}     & \includegraphics[width=0.31\linewidth]{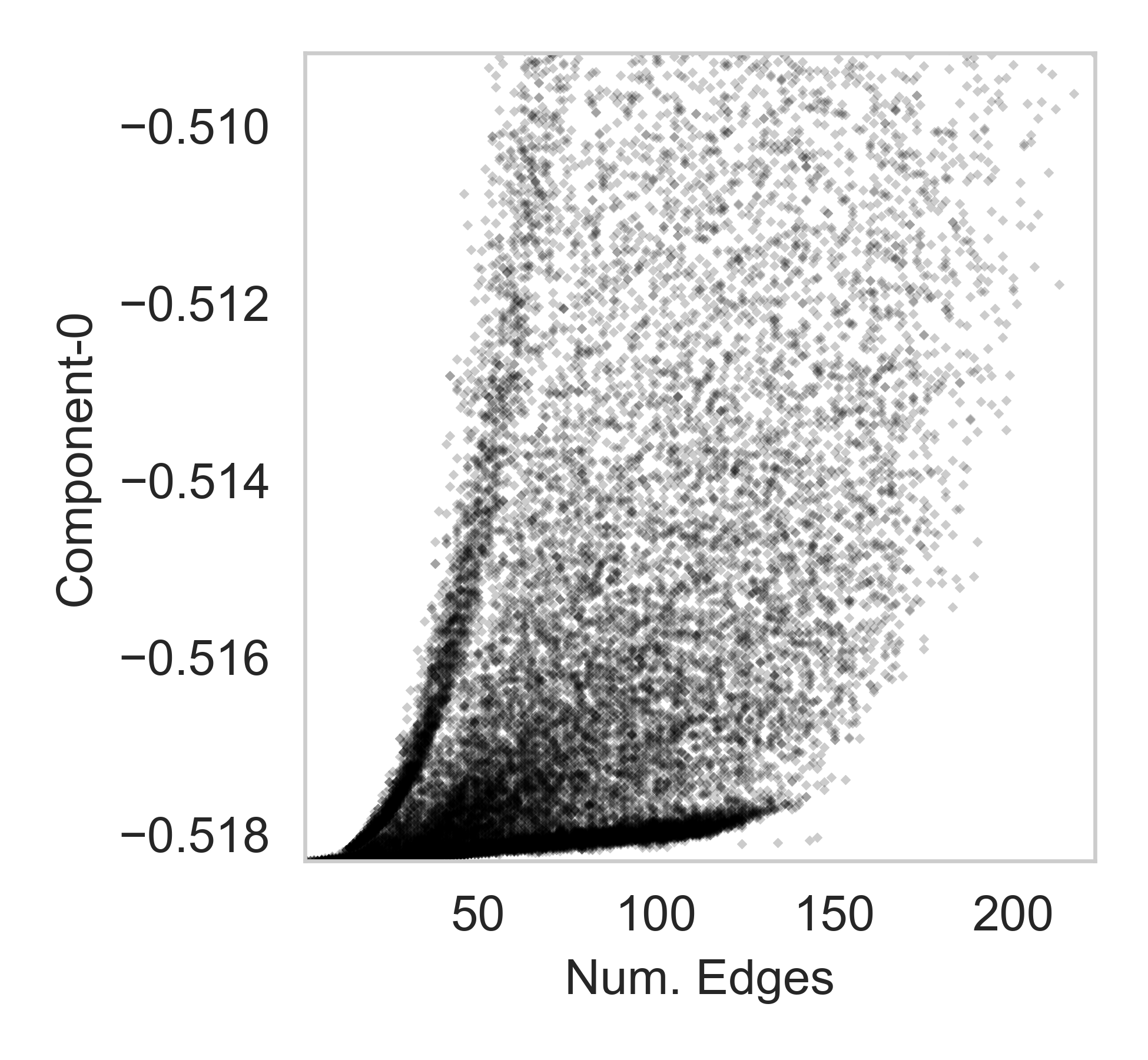} & \includegraphics[width=0.31\linewidth]{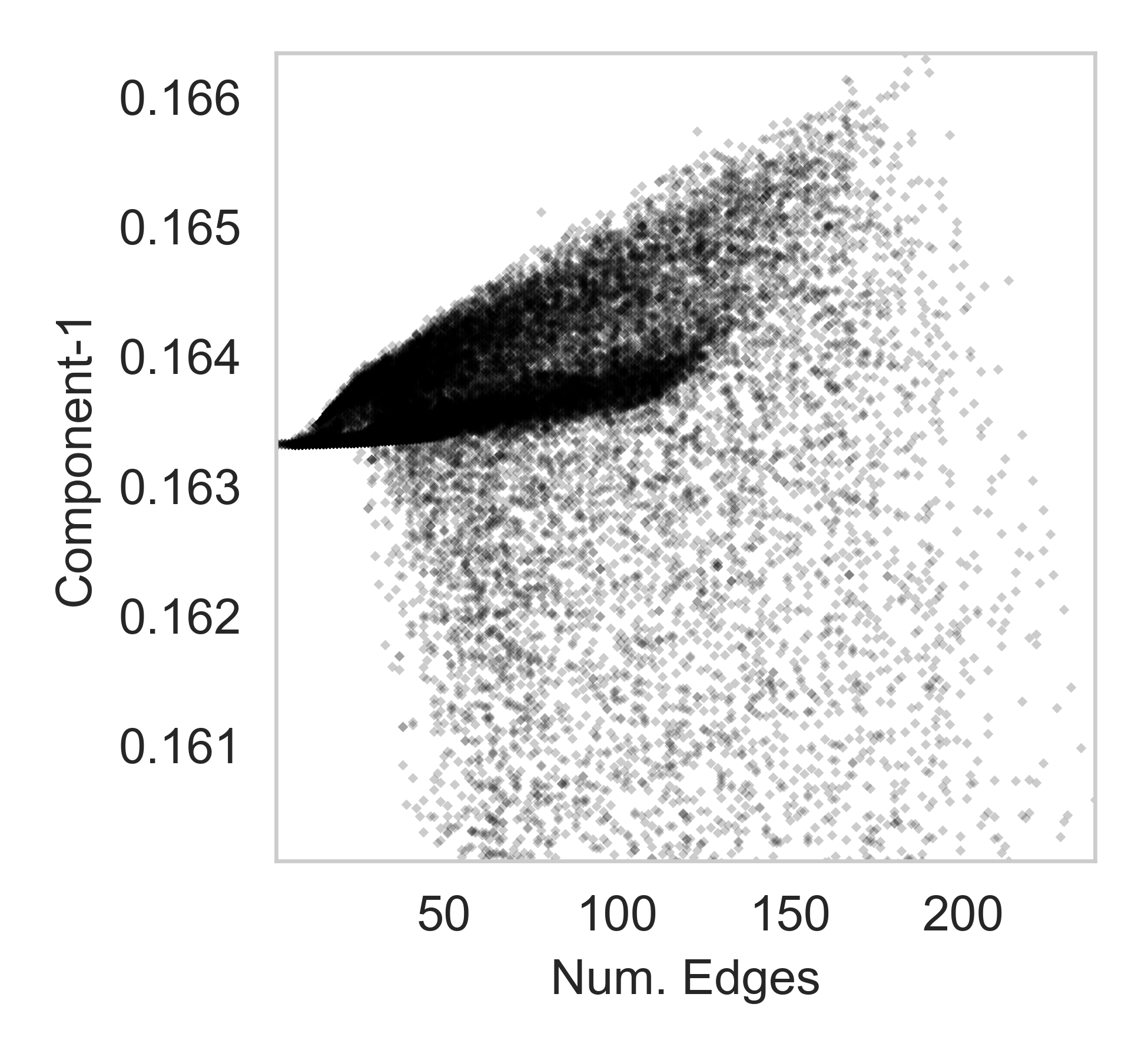} & \includegraphics[width=0.31\linewidth]{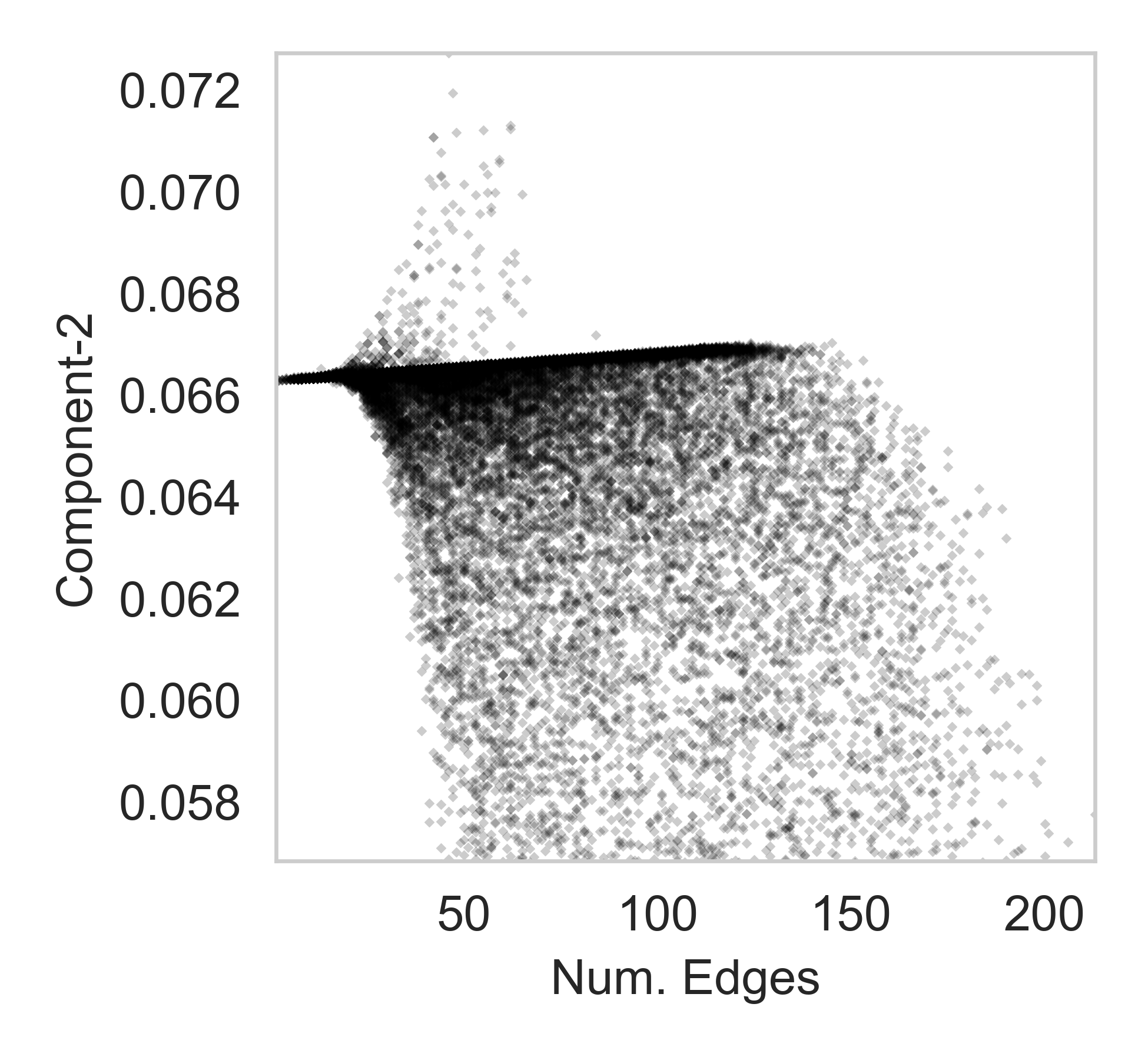} \\

    \rotatebox[origin=lt]{90}{\hspace{4em}Chem}     & \includegraphics[width=0.31\linewidth]{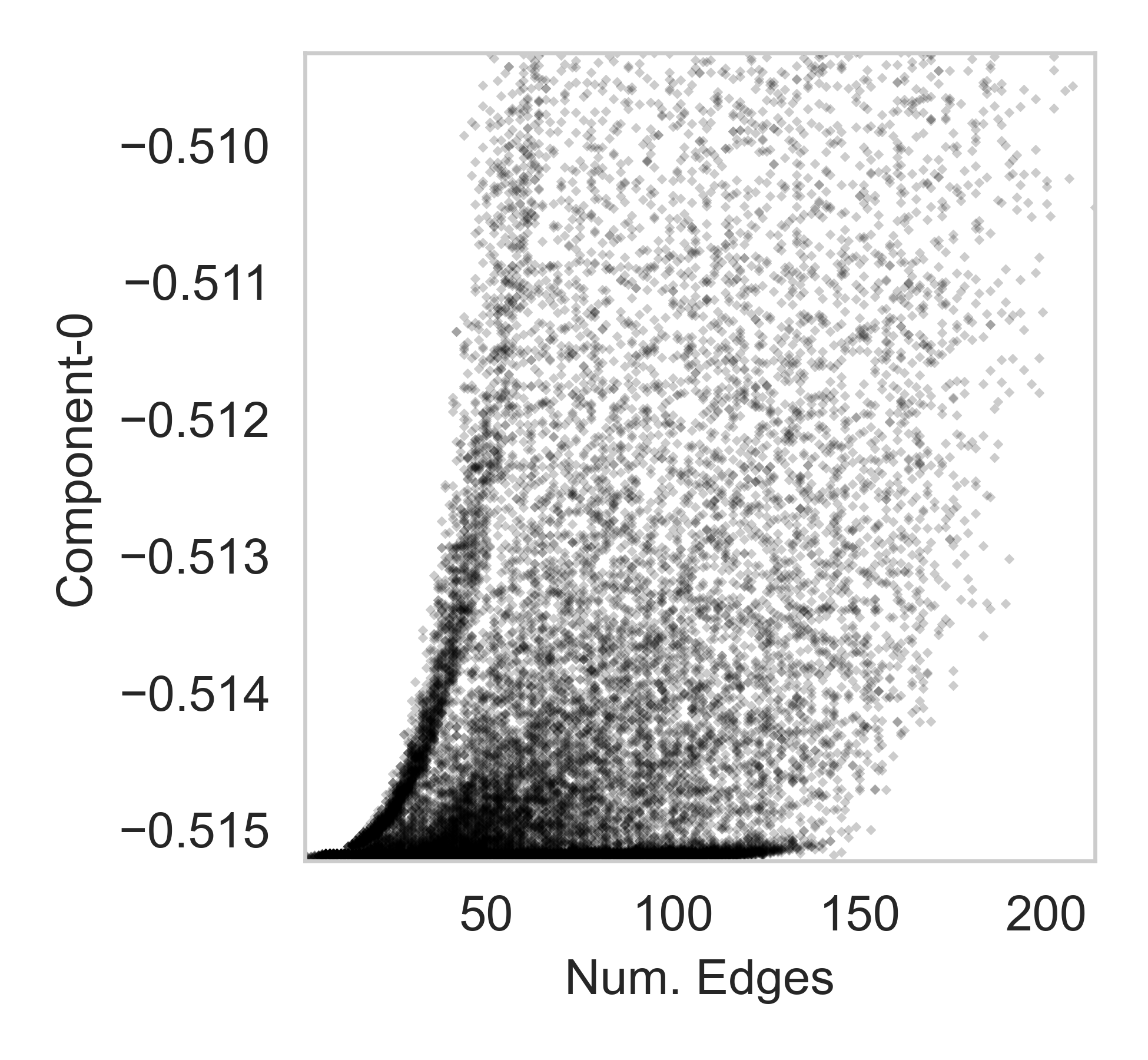} & \includegraphics[width=0.31\linewidth]{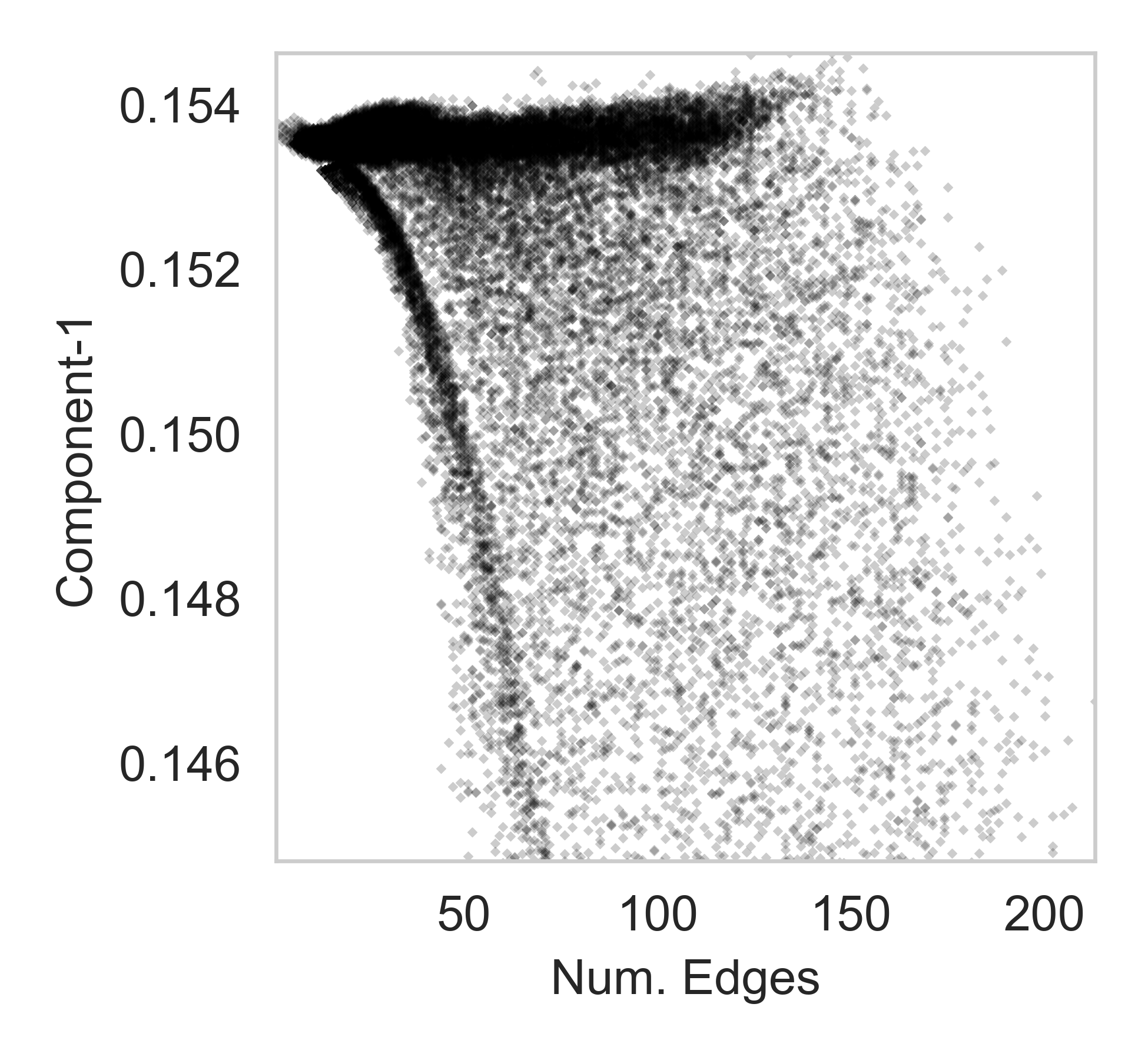} & \includegraphics[width=0.31\linewidth]{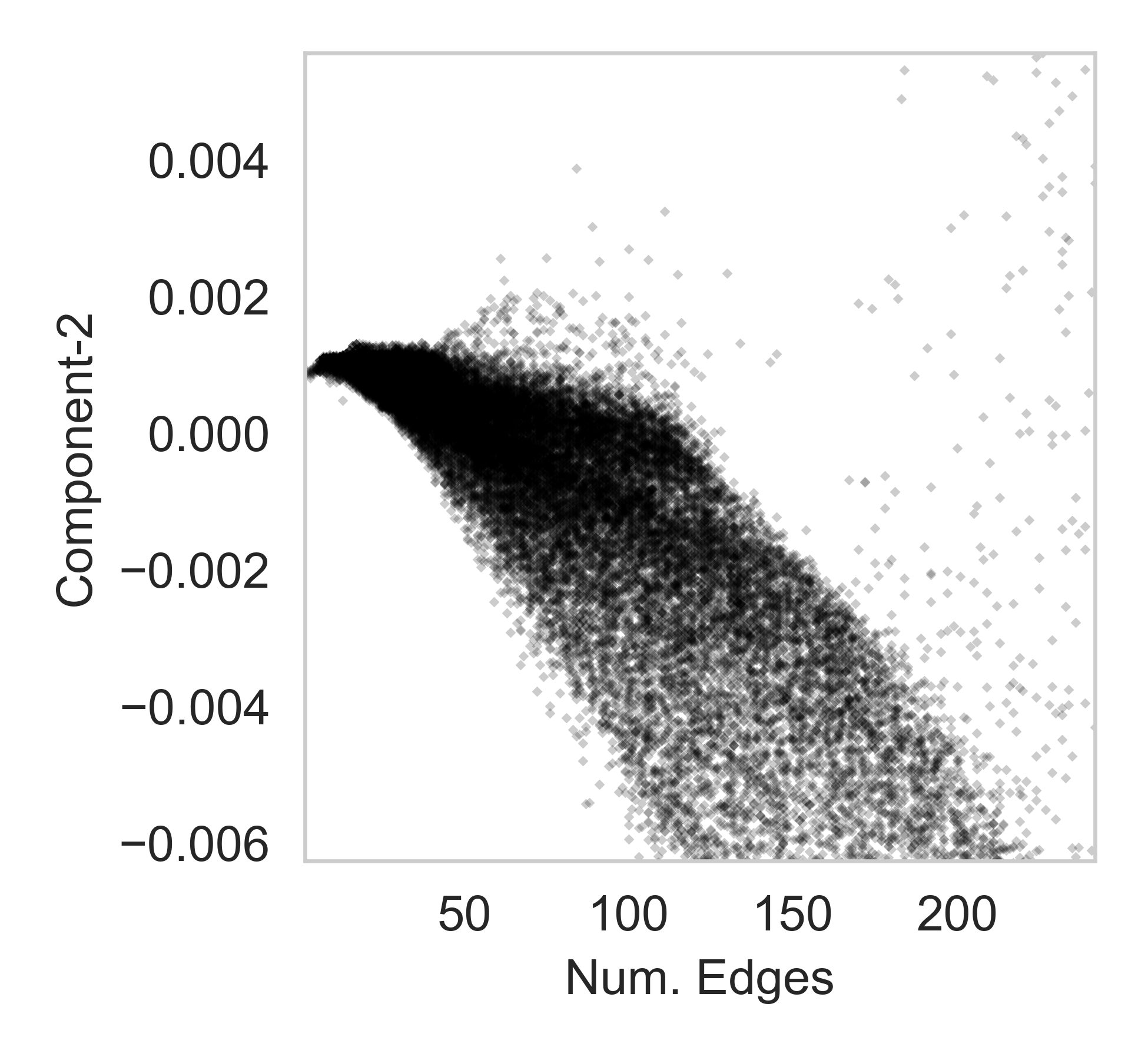} \\

    \rotatebox[origin=lt]{90}{\hspace{4em}Social}     & \includegraphics[width=0.31\linewidth]{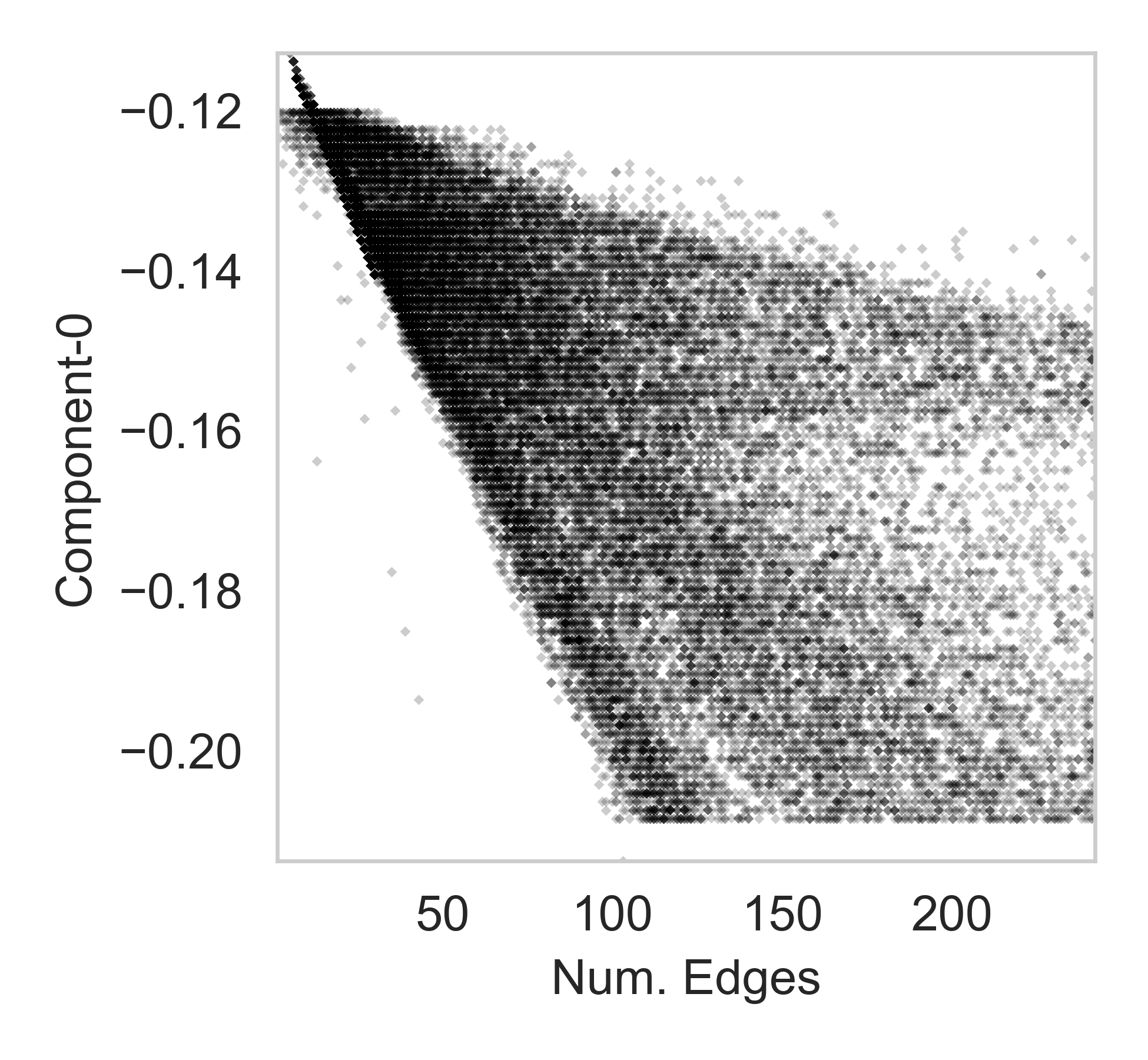} & \includegraphics[width=0.31\linewidth]{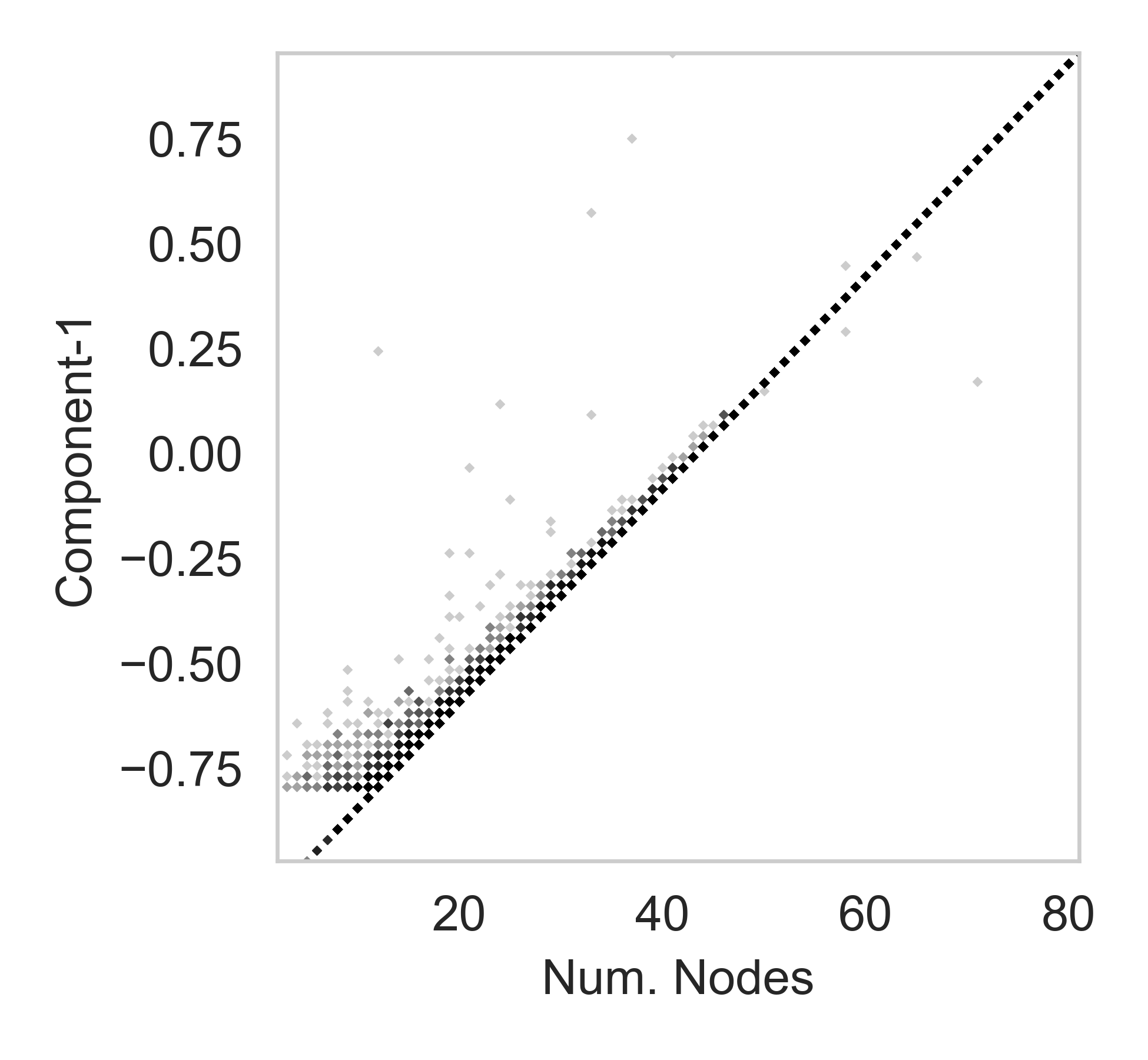} & \includegraphics[width=0.31\linewidth]{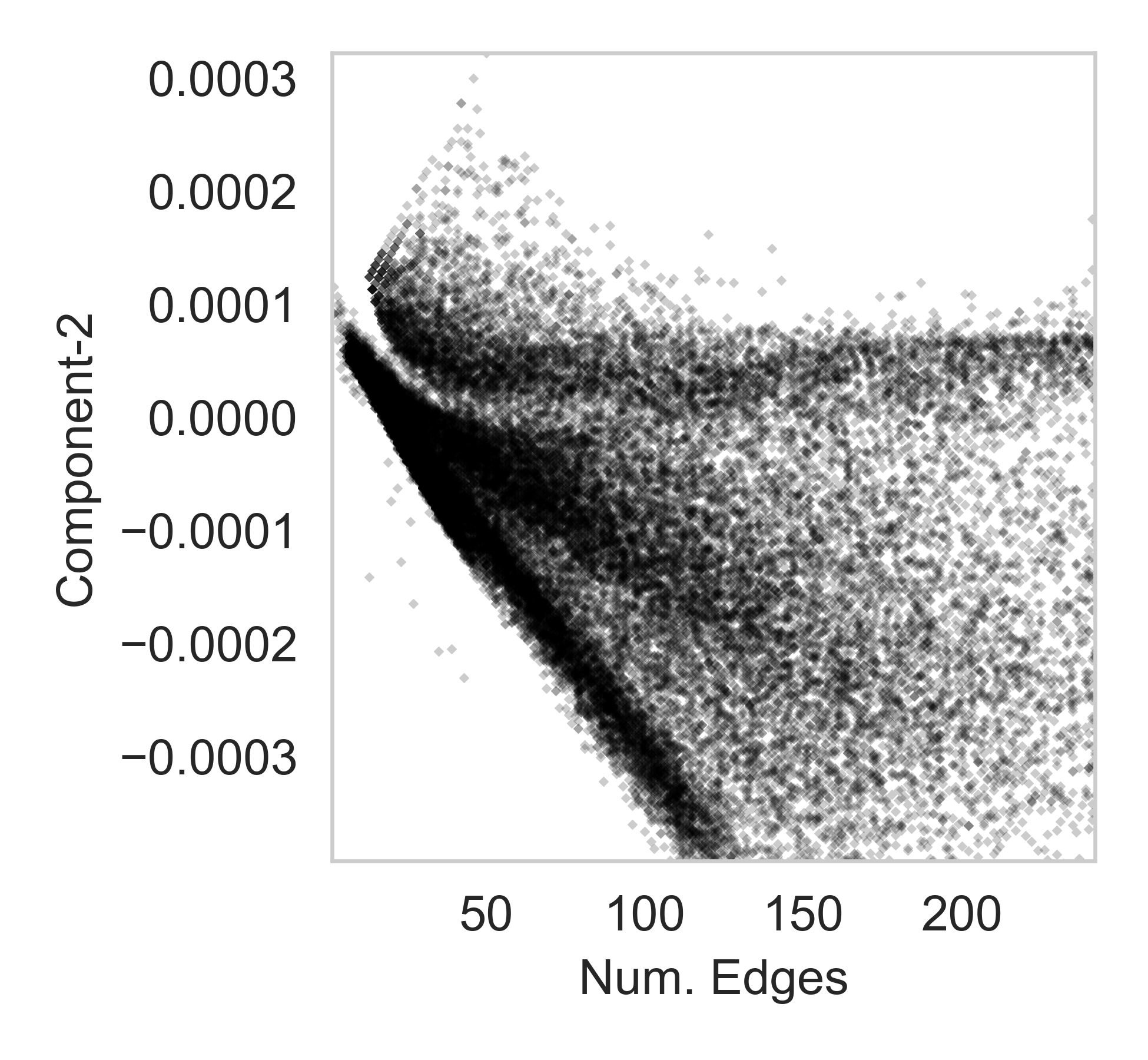} \\

    \rotatebox[origin=lt]{90}{\hspace{5em}All}     & \includegraphics[width=0.31\linewidth]{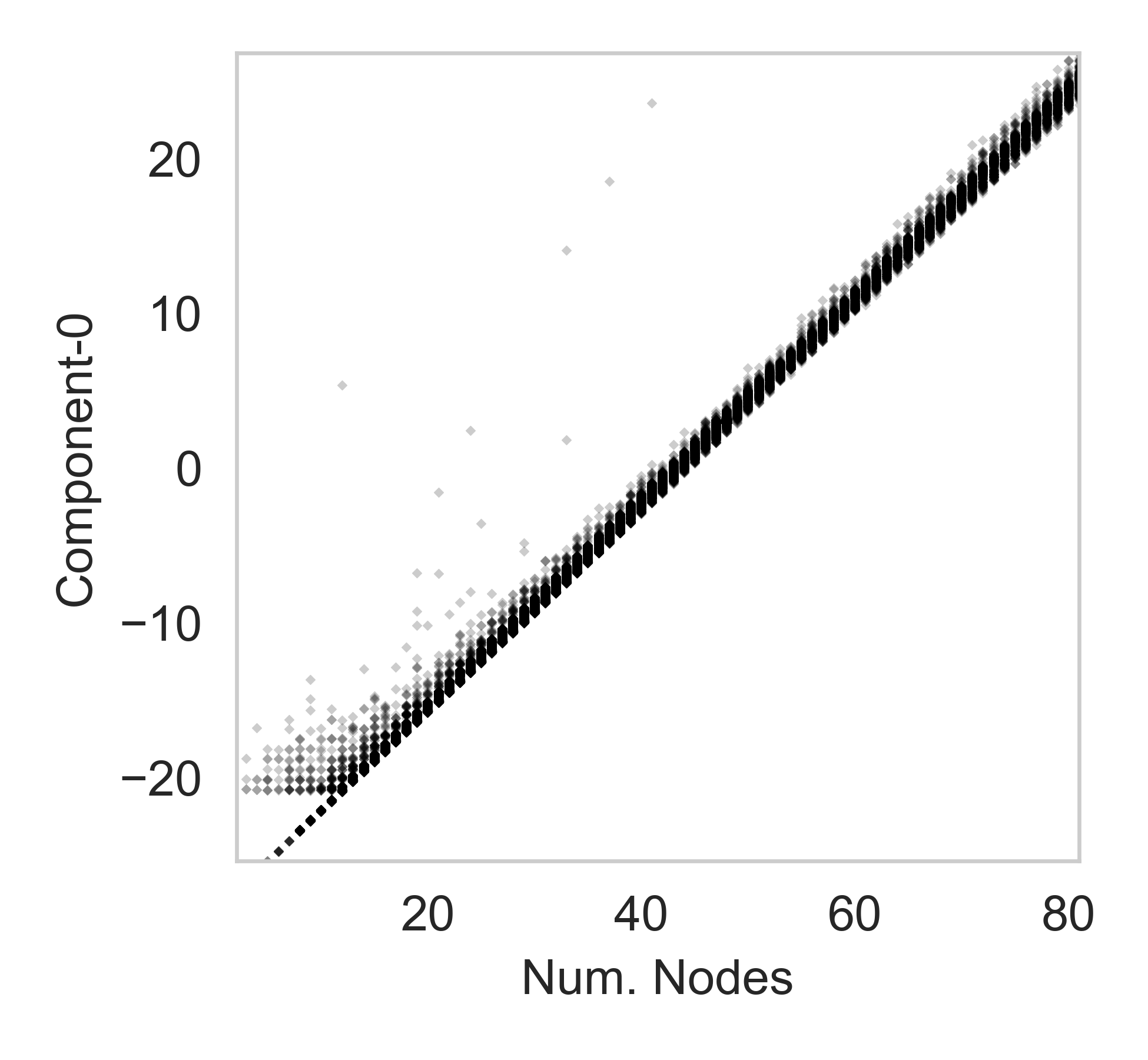} & \includegraphics[width=0.31\linewidth]{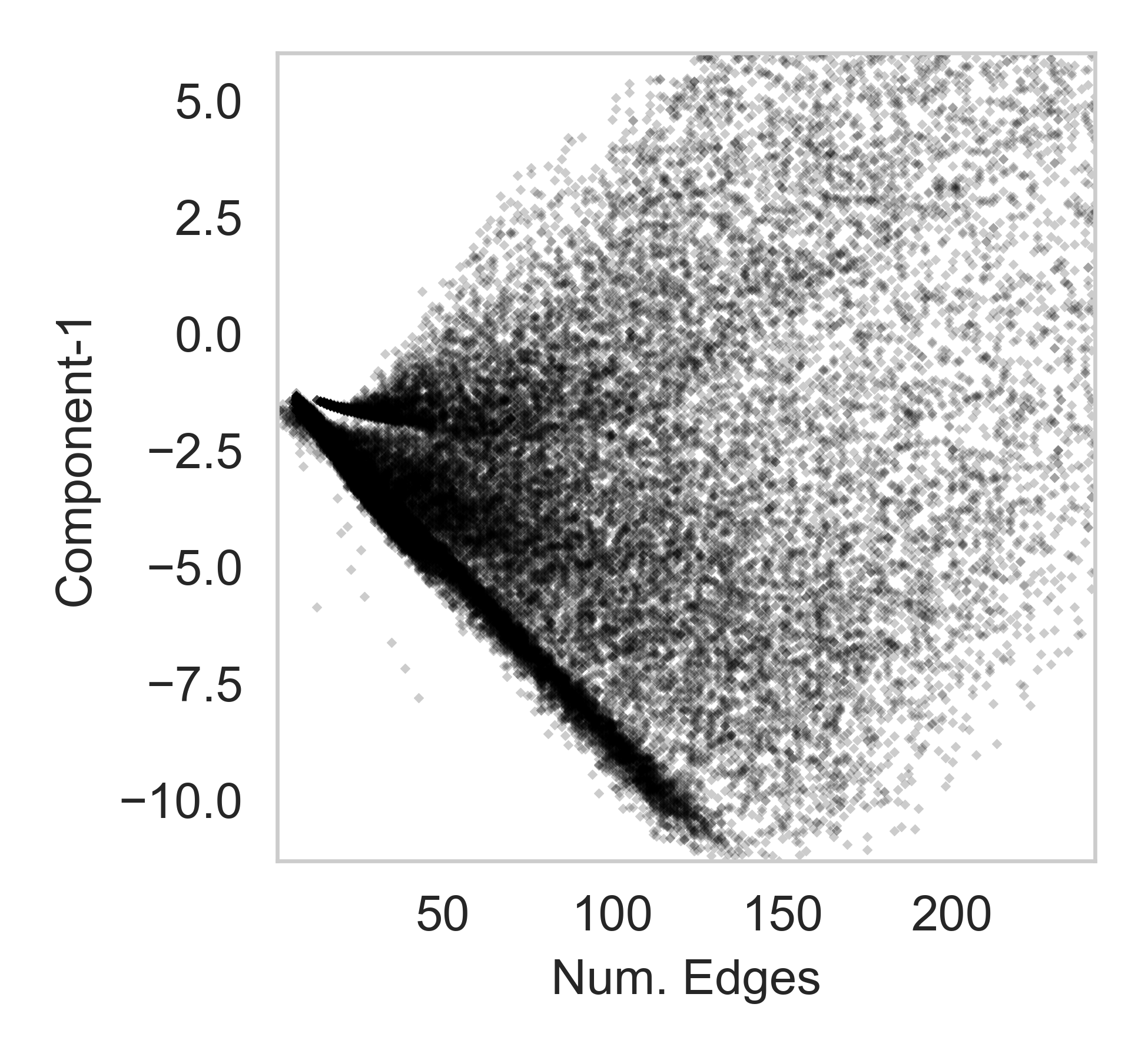} & \includegraphics[width=0.31\linewidth]{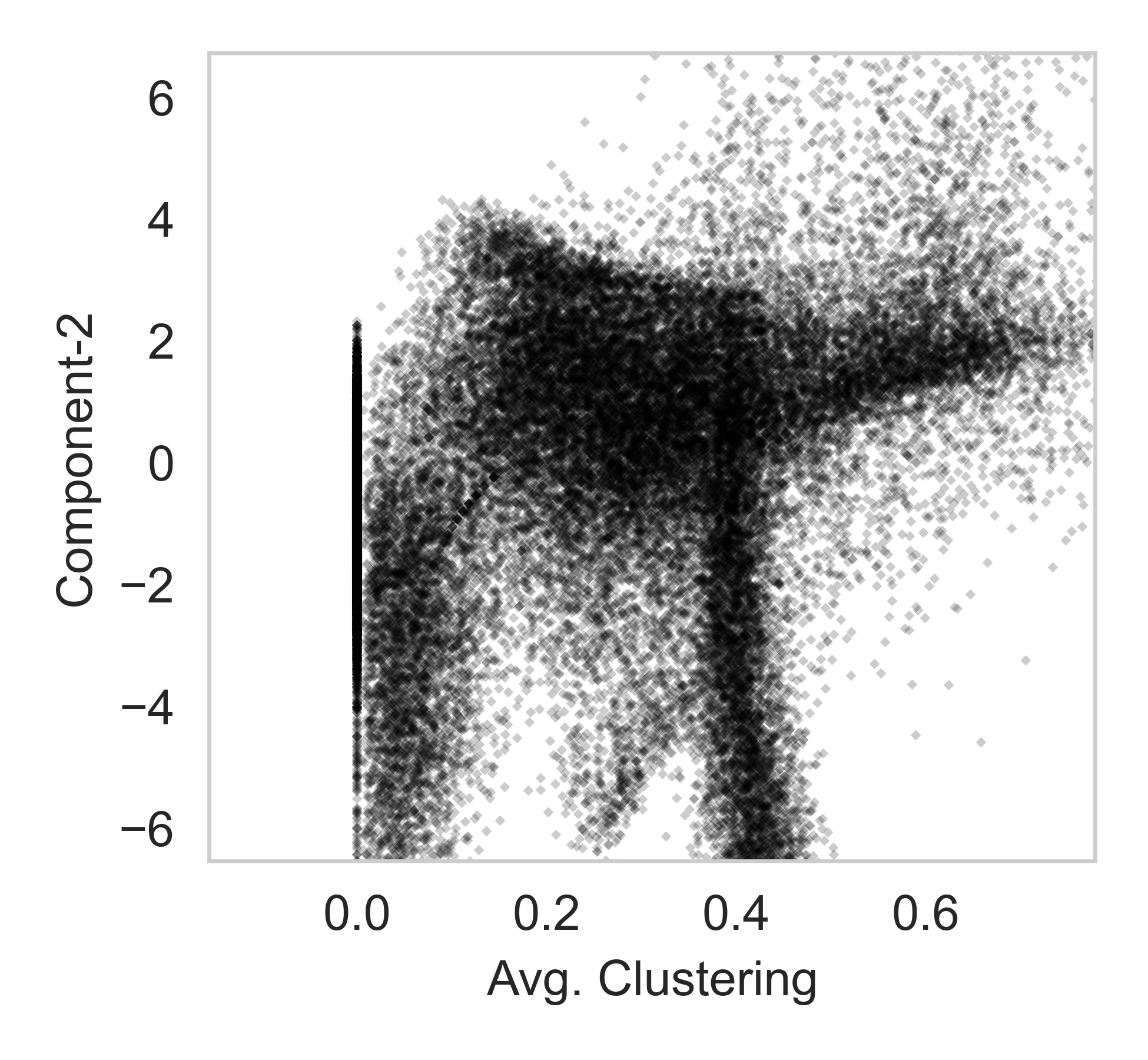} \\
    
    \end{tabular}
    \caption{PCA components of the encodings of each model across the whole validation dataset, scattered against the most correlated metrics, as in Table~\ref{tab:component-correlations}.}
    \label{tab:pca-vs-metric}
\end{figure}

\clearpage
\newpage

\section{Further Results} \label{subsec:further-results}

\begin{figure}[h]
    \centering
    \subfigure[Facebook]{\includegraphics[width=0.32\linewidth]{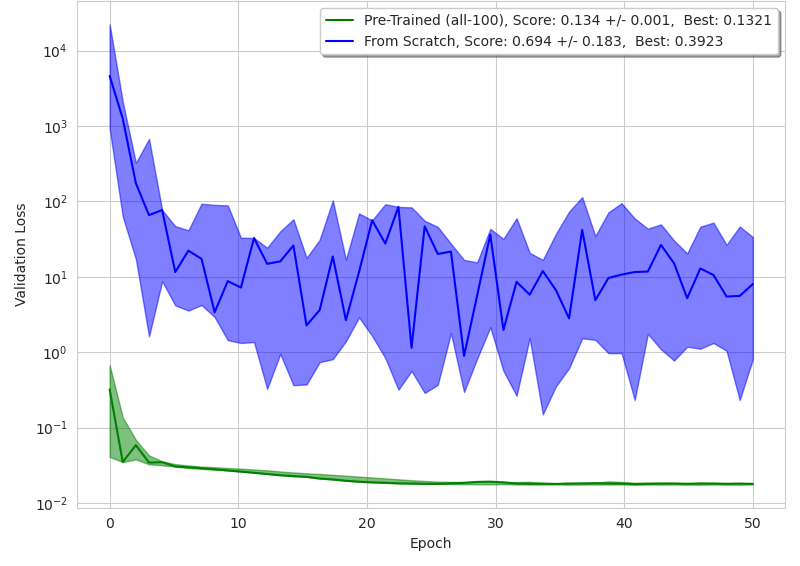} \label{fig:facebook-training}}
    \subfigure[Twitch Egos]{\includegraphics[width=0.32\linewidth]{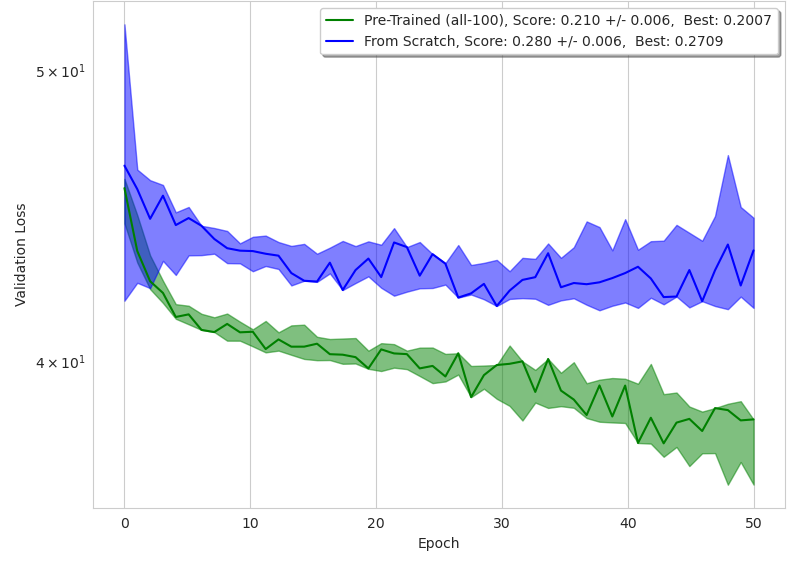}  \label{fig:twitch-training}}
    \subfigure[ogbg-molesol]{\includegraphics[width=0.32\linewidth]{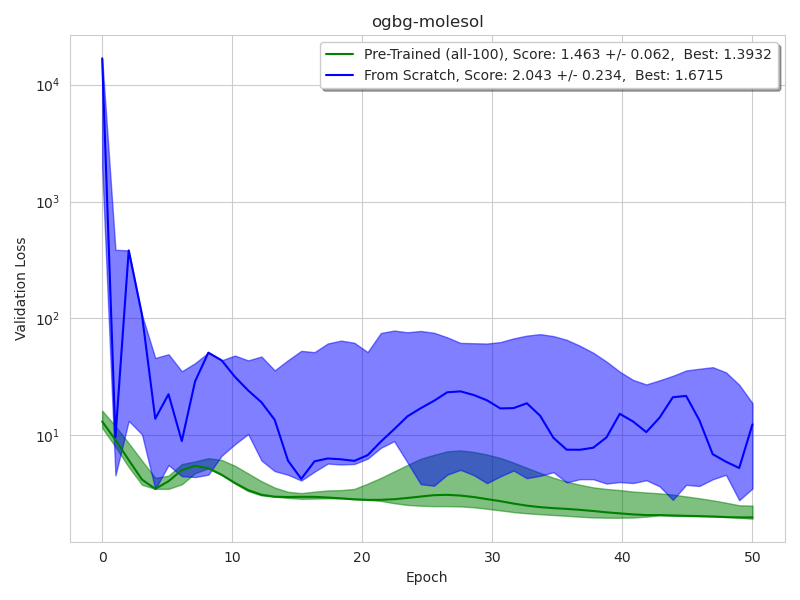} \label{fig:molesol-training}}
    \subfigure[ogbg-mollipo]{\includegraphics[width=0.32\linewidth]{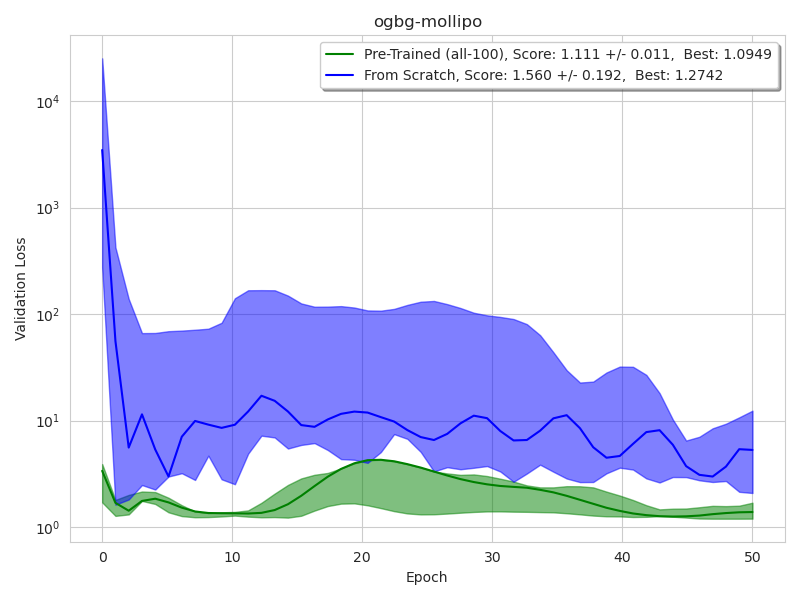}  \label{fig:mollipo-training}}
    \subfigure[Random]{\includegraphics[width=0.32\linewidth]{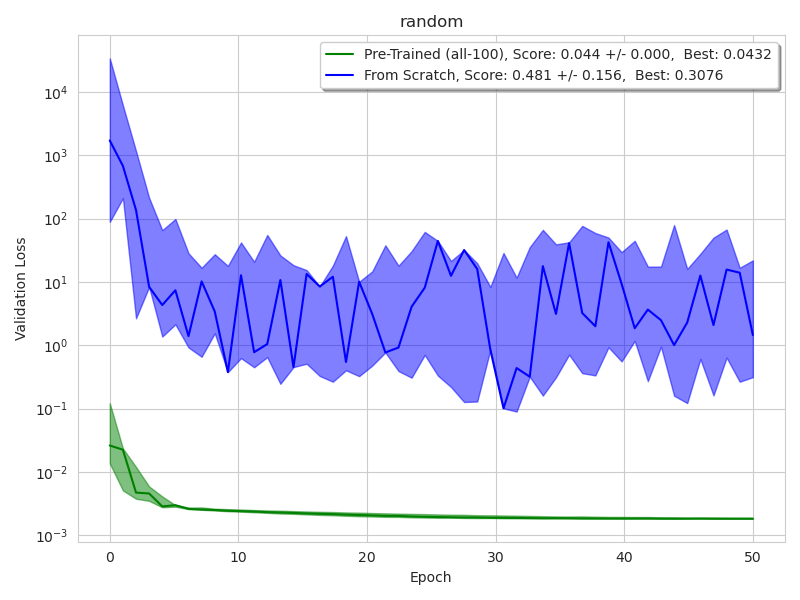} \label{fig:random-training}}
    \subfigure[Community]{\includegraphics[width=0.32\linewidth]{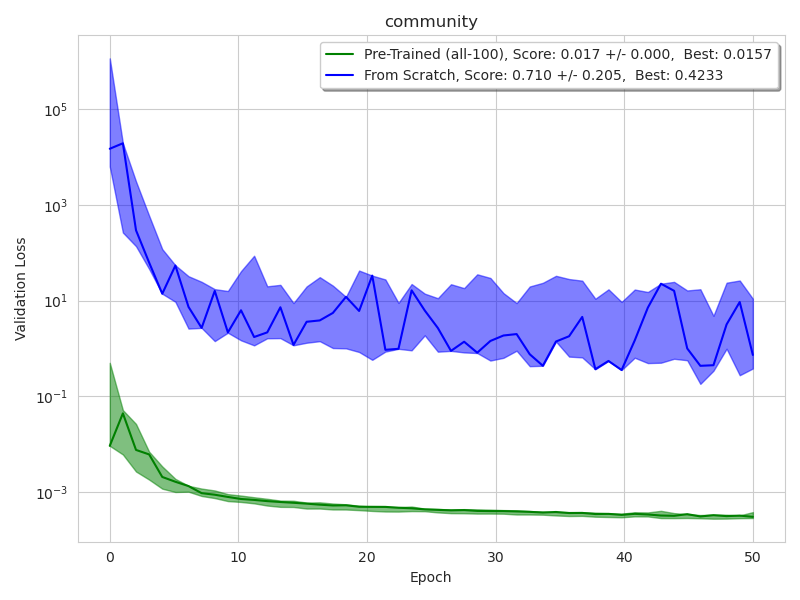}  \label{fig:community-training}}
    \subfigure[Trees]{\includegraphics[width=0.32\linewidth]{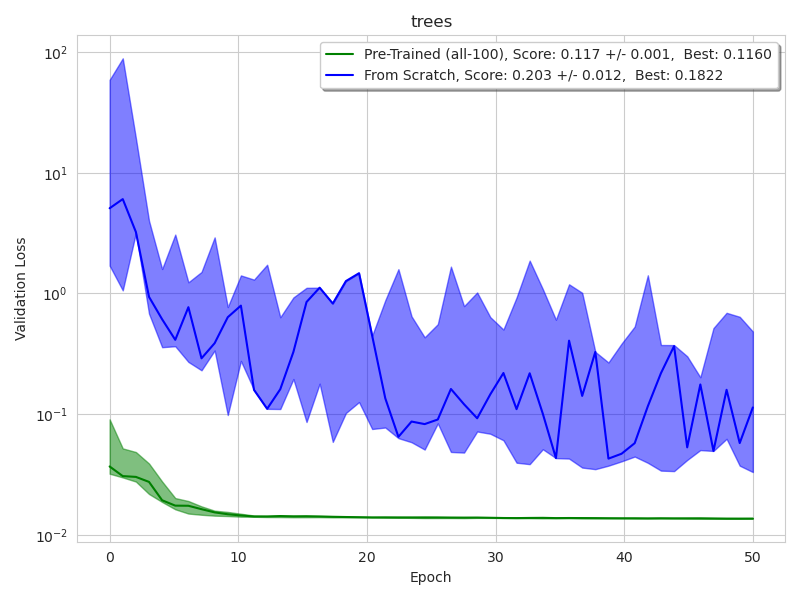} \label{fig:trees-training}}
    \subfigure[Roads]{\includegraphics[width=0.32\linewidth]{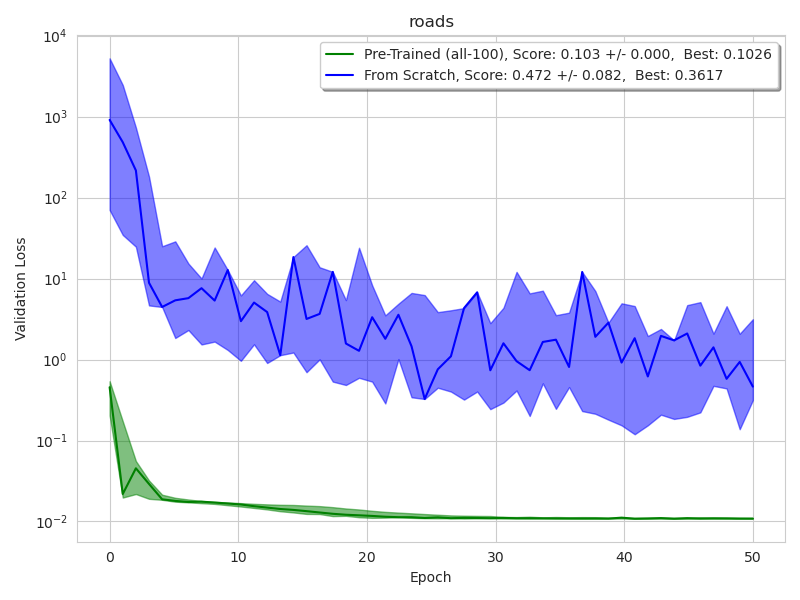}  \label{fig:roads-training}}

    \caption{Validation scores for fine-tuning the All model on a subset of our evaluation datasets. 
    The Twitch dataset carries its binary classification task and so we report 1 - AUROC.
    On other datasets tasks are regression, and we report MSE.} 
    % We include the average scores and the best scores over the ten training runs on each dataset.}
    \label{fig:training-examples}
\end{figure}

% \clearpage

% \clearpage
% \newpage

\subsection{Linear Transfer}

\begin{table}[h]
\caption{Scores for linear models applied to the encodings of each validation dataset, with and without node labels. For the classification datasets (ogbg-molclintox and Twitch Egos) we report AUROC, other datasets report RMSE. Some datasets, where results are very similar for each model, are excluded. \textbf{Bold} text indicates a superior result. For clarity the results for a given dataset are scaled by a constant factor.}
\label{tab:linear-transfer}
% \small 
\centering
    % \begin{tabular}{c||c|c|c|c}
    \begin{tabular}{l  r  r  r  r  l}
    \toprule
    &  \multicolumn{1}{r}{\textbf{Untrained}}    &  \multicolumn{1}{r}{\textbf{Chem}}         &  \multicolumn{1}{r}{\textbf{Social}}       &  \multicolumn{1}{r}{\textbf{All}} &    {Magnitude}            \\\midrule
    molfreesolv &  4.21 $\pm$  4.13 &  3130 $\pm$  5990 & 2.00 $\pm$  0.869 & \fbf 1.01 $\pm$  1.02 &$10^3$    \\
    molesol     &  48.3        $\pm$  90.2          &  817    $\pm$  1150               &  48.1        $\pm$  62.9          &  \fbf 28.1          $\pm$  38.1  &$10^0$      \\
    mollipo      &  29.4        $\pm$  17.9          &  14.9   $\pm$  2.0                &  \fbf 4.45        $\pm$  0.62          &  5.33          $\pm$  0.56   &$10^0$   \\
    molclintox   &  0.487       $\pm$  0.005         &  0.440  $\pm$  0.007              &  \fbf 0.500       $\pm$  0.000         &  \fbf 0.500         $\pm$  0.000  &$10^0$    \\\midrule
    
    Facebook  &  10.6 $\pm$  1.7  & 10.7 $\pm$ 20  & 7.73 $\pm$  1.87  & \fbf 4.44 $\pm$  5 &$10^{-3}$   \\
    Twitch Egos &  0.638       $\pm$  0.012          &  0.610  $\pm$  0.072             &  0.679       $\pm$  0.002          &  \fbf 0.694         $\pm$  0.002  &$10^0$     \\
    Roads     &  263 $\pm$ 0.5 & 256 $\pm$ 0.0 & 298 $\pm$  0.3 & \fbf 2.85 $\pm$  5 &$10^{-5}$   \\\midrule
    
    Trees   & 7.66 $\pm$  0.16  & 2.85 $\pm$ 0.0   & 2.98 $\pm$  0.03 & \fbf 2.56 $\pm$  0.05 &$10^{-3}$     \\
    Community  &  4.36 $\pm$  0.05 & \fbf 4.25 $\pm$  0.00 & 12.0 $\pm$  2  & 4.46 $\pm$  0.07 &$10^{-5}$  \\
    Random    &  6.46 $\pm$  0.7  & 3.27 $\pm$  0.0  & \fbf 1.20 $\pm$  0.2  & 3.38 $\pm$  0.0 &$10^{-4}$   \\
    \midrule
      &  \multicolumn{1}{r}{\textbf{Untrained}-Labels}         &  \multicolumn{1}{r}{\textbf{Chem}-Labels}    &  \multicolumn{1}{r}{\textbf{Social}-Labels}         &  \multicolumn{1}{r}{\textbf{All}-Labels  }  &                       \\\midrule
    molfreesolv  &  \fbf 480    $\pm$  172           &  \small{$(1.13 \pm  0.64) \times 10^{6}$}  &  90900 $\pm$  83300 & 11000 $\pm$  5600   &$10^0$                \\
    molesol      &  29.4        $\pm$  7.8           &  \fbf 18.5   $\pm$  7.3                 &  531        $\pm$  292            &  34.1     $\pm$  33.0   &$10^0$           \\
    mollipo       &  7.33        $\pm$  1.37          &  7.25   $\pm$  1.01                &  \fbf 4.02   $\pm$  0.92               &  5.01     $\pm$  0.53  &$10^0$           \\
    molclintox     &  0.511       $\pm$  0.022         &  0.497  $\pm$  0.001               &  0.428       $\pm$  0.001         &  \fbf 0.526    $\pm$  0.060  &$10^0$          \\\midrule
    Facebook   & 14.6 $\pm$  11.1 &  10.8 $\pm$  1.1 &  7.82 $\pm$  1.9 & \fbf 4.43 $\pm$  6  &$10^{-3}$  \\
    \bottomrule
\end{tabular}
\end{table}

% =========================================================================================================================================================

\clearpage
\newpage

\subsection{Transfer Learning} 

\begin{table}[h]
\caption{Scores for fine-tuning models trained with \textbf{random node dropping} on each validation dataset, with (\textbf{Model}-labels) and without (\textbf{Model}) node labels. For the classification datasets (ogbg-molclintox and Twitch Egos) we report AUROC, other datasets report RMSE. Some datasets, where results are very similar for each model, are excluded. \textbf{Bold} text indicates a result beats the unsupervised (GIN) baseline. \underline{\textbf{Underlined}} text indicates a superior result, down to three significant figures. GIN results are duplicated from Table \ref{tab:full-transfer}.}
\label{tab:graphcl-node-transfer}

\centering
    % \begin{tabular}{c||c|c|c|c}
    \begin{tabular}{l  r  r  r  r}
    \toprule
            &  \multicolumn{1}{r}{\textbf{GIN}}         &  \multicolumn{1}{r}{\textbf{Chem}}    &  \multicolumn{1}{r}{\textbf{Social}}          &  \multicolumn{1}{r}{\textbf{All}}   \\\midrule

    % Regression
    molfreesolv & 4.978 $\pm$ 0.479 & \fbf \underline{3.838 $\pm$ 0.120} & \fbf 4.041 $\pm$ 0.116 & \fbf 4.157 $\pm$ 0.138 \\
    molesol     & 1.784 $\pm$ 0.094 & 1.897 $\pm$ 0.075 & \fbf 1.520 $\pm$ 0.066 & \fbf \underline{1.357 $\pm$ 0.037} \\
    mollipo     & 1.143 $\pm$ 0.031 & \fbf 1.123 $\pm$ 0.105 & \fbf 1.015 $\pm$ 0.013 & \fbf \underline{0.983 $\pm$ 0.015} \\
    \midrule
    
    % Classification
    molclintox & \underline{0.530 $\pm$ 0.042} & 0.469 $\pm$ 0.030 & 0.431 $\pm$ 0.002 & \fbf \underline{0.530 $\pm$ 0.086} \\ 
    molbbbp    & \underline{0.565 $\pm$ 0.043} & 0.497 $\pm$ 0.027 & 0.548 $\pm$ 0.037 & \fbf \underline{0.567 $\pm$ 0.032} \\
    molbace    & 0.677 $\pm$ 0.053 & \fbf \underline{0.740 $\pm$ 0.026} & \fbf 0.686 $\pm$ 0.055 & \fbf 0.679 $\pm$ 0.067 \\
    molhiv     & 0.355 $\pm$ 0.059 & 0.283 $\pm$ 0.001 & \fbf 0.635 $\pm$ 0.046 & \fbf \underline{0.644 $\pm$ 0.039} \\
    \midrule

    Facebook       &  0.311       $\pm$  0.065          &  \fbf 0.261  $\pm$  0.101                  &   \fbf 0.159  $\pm$  0.015                         &  \fbf  \underline{0.145    $\pm$  0.012}   \\
    Twitch Egos    &  0.72        $\pm$  0.006          &  0.460  $\pm$  0.041                  &  \fbf \underline{0.744       $\pm$  0.006}                     &   \fbf \underline{0.743     $\pm$  0.003}   \\
    Roads          &  0.46        $\pm$  0.073          &  \fbf 0.228  $\pm$  0.118                  &   \fbf 0.150  $\pm$  0.015                         &    \fbf \underline{0.132    $\pm$  0.018}   \\\midrule
    
    Trees          &  0.203       $\pm$  0.012          &  0.415  $\pm$  0.250                  &   \fbf 0.135  $\pm$  0.015                         &   \fbf \underline{0.125         $\pm$  0.009}   \\
    Community      &  0.743       $\pm$  0.106          &  \fbf 0.084  $\pm$  0.083                  &   \fbf 0.080  $\pm$  0.018                         &  \fbf  \underline{0.074    $\pm$  0.021}  \\
    Random         &  0.529       $\pm$  0.228          &  \fbf 0.123  $\pm$  0.075                  &   \fbf 0.076  $\pm$  0.010                         &    \fbf \underline{0.058    $\pm$  0.005}   \\
    \midrule
                   &  \multicolumn{1}{r}{\textbf{GIN}-Labels}  &  \multicolumn{1}{r}{\textbf{Chem}-Labels} &  \multicolumn{1}{r}{\textbf{Social}-Labels}  &  \multicolumn{1}{r}{\textbf{All}-Labels  }   \\\midrule

    % Regression
    molfreesolv & 4.854 $\pm$ 0.513 & \fbf \underline{3.819 $\pm$ 0.136} & \fbf 4.113 $\pm$ 0.070 & \fbf 3.984 $\pm$ 0.162 \\
    molesol     & 1.707 $\pm$ 0.073 & 1.809 $\pm$ 0.139 & \fbf 1.480 $\pm$ 0.067 & \fbf \underline{1.378 $\pm$ 0.031} \\
    mollipo     & 1.143 $\pm$ 0.031 & 1.182 $\pm$ 0.128 & \fbf \underline{0.986 $\pm$ 0.017} & \fbf 0.997 $\pm$ 0.011 \\
    \midrule
    
    % Classification
    molclintox & \underline{0.497 $\pm$ 0.038} & 0.474 $\pm$ 0.036 & 0.430 $\pm$ 0.002 &  0.474 $\pm$ 0.056 \\ 
    molbbbp    & \underline{0.561 $\pm$ 0.034} & 0.491 $\pm$ 0.032 & 0.543 $\pm$ 0.016 & \fbf \underline{0.565 $\pm$ 0.036} \\
    molbace    & 0.698 $\pm$ 0.050 &  \fbf \underline{0.751 $\pm$ 0.024} & 0.677 $\pm$ 0.030 & 0.675 $\pm$ 0.088 \\
    molhiv     & 0.374 $\pm$ 0.052 & 0.286 $\pm$ 0.008 &  \fbf \underline{0.615 $\pm$ 0.047} &  \fbf \underline{0.628 $\pm$ 0.044} \\
    \midrule

    Facebook       &  0.749       $\pm$  0.250          &  \fbf 0.354  $\pm$  0.166                  &  \fbf \underline{0.151       $\pm$  0.016}                     &   \fbf  0.156    $\pm$  0.022   \\

    \midrule
    \textbf{Superior Result} & 19.0\% & 19.0\% & 19.0\% & \fbf 61.9\% \\
    \textbf{Beats Baseline} & \shyphen & 47.6\% & 76.2\% &  \fbf 90.5\% \\
    \bottomrule
\end{tabular}
\end{table}

% =========================================================================================================================================================

\begin{table}[h]
\caption{Scores for fine-tuning models trained with \textbf{random edge dropping} on each validation dataset, with (\textbf{Model}-labels) and without (\textbf{Model}) node labels. For the classification datasets (ogbg-molclintox and Twitch Egos) we report AUROC, other datasets report RMSE. Some datasets, where results are very similar for each model, are excluded. \textbf{Bold} text indicates a result beats the unsupervised (GIN) baseline. \underline{\textbf{Underlined}} text indicates a superior result, down to three significant figures. GIN results are duplicated from Table \ref{tab:full-transfer}.}
\label{tab:graphcl-edge-transfer}

\centering
    % \begin{tabular}{c||c|c|c|c}
    \begin{tabular}{l  r  r  r  r}
    \toprule
            &  \multicolumn{1}{r}{\textbf{GIN}}     &  \multicolumn{1}{r}{\textbf{Chem}}    &  \multicolumn{1}{r}{\textbf{Social}}         &  \multicolumn{1}{r}{\textbf{All}}   \\\midrule
            
    % Regression
    molfreesolv & 4.978 $\pm$ 0.479 &134.800$\pm$ 225.200& \fbf  \underline{4.102 $\pm$ 0.049} & \fbf 4.197 $\pm$ 0.095 \\
    molesol     & 1.784 $\pm$ 0.094 &14.050 $\pm$ 23.760 & \fbf  \underline{1.312 $\pm$ 0.094} & \fbf 1.571 $\pm$ 0.059 \\
    mollipo     & 1.129 $\pm$ 0.040 & \fbf 1.034 $\pm$ 0.029  &  \fbf \underline{0.989 $\pm$ 0.011} & \fbf 1.068 $\pm$ 0.026 \\
    \midrule
    
    % Classification
    molclintox & 0.530 $\pm$ 0.042 & 0.449 $\pm$ 0.013 & 0.426 $\pm$ 0.006 & 0.461 $\pm$ 0.029 \\ 
    molbbbp    & 0.565 $\pm$ 0.043 & \fbf 0.570 $\pm$ 0.040 &  \fbf \underline{0.602 $\pm$ 0.025} &  \fbf \underline{0.603 $\pm$ 0.029} \\
    molbace    & 0.677 $\pm$ 0.053 &  \fbf 0.768 $\pm$ 0.006 &  \fbf \underline{0.783 $\pm$ 0.045} &  \fbf 0.756 $\pm$ 0.030 \\
    molhiv     & 0.355 $\pm$ 0.059 & \fbf 0.576 $\pm$ 0.152 &  \fbf \underline{0.647 $\pm$ 0.066} & \fbf 0.501 $\pm$ 0.135 \\
    \midrule
    
    Facebook       &  0.311       $\pm$  0.065      &  3.70   $\pm$  6.95                   &    \fbf \underline{0.180  $\pm$  0.019}             &   \fbf 0.237    $\pm$  0.063   \\
    Twitch Egos    &  0.72        $\pm$  0.006      &  0.717  $\pm$  0.005                  &   \fbf \underline{0.753   $\pm$  0.006}             &   \fbf 0.739     $\pm$  0.007   \\
    Roads          &  0.46        $\pm$  0.073      &   \fbf \underline{0.156  $\pm$  0.007}                  &   \fbf 0.167  $\pm$  0.039             &   \fbf 0.252    $\pm$  0.029   \\\midrule
    
    Trees          &  0.203       $\pm$  0.012      &   \fbf \underline{0.143  $\pm$  0.009}                  &   \fbf 0.161  $\pm$  0.019             &  \fbf 0.187         $\pm$  0.020   \\
    Community      &  0.743       $\pm$  0.106      &   \fbf \underline{0.014  $\pm$  0.000}                  &   \fbf 0.146  $\pm$  0.024             &   \fbf 0.314    $\pm$  0.069  \\
    Random         &  0.529       $\pm$  0.228      &   \fbf \underline{0.055  $\pm$  0.012}                  &   \fbf 0.107  $\pm$  0.019             &   \fbf 0.289    $\pm$  0.098   \\
    \midrule
                   &  \multicolumn{1}{r}{\textbf{GIN}-Labels}         &  \multicolumn{1}{r}{\textbf{Chem}-Labels}    &  \multicolumn{1}{r}{\textbf{Social}-Labels}         &  \multicolumn{1}{r}{\textbf{All}-Labels  }   \\\midrule
    % Regression
    molfreesolv & 4.854 $\pm$ 0.513 &103.400$\pm$232.200& \fbf \underline{4.106 $\pm$ 0.021} &  \fbf 4.163 $\pm$ 0.069 \\
    molesol     & 1.707 $\pm$ 0.073 & 8.006 $\pm$ 7.850 &  \fbf \underline{1.358 $\pm$ 0.028} & \fbf 1.597 $\pm$ 0.069 \\
    mollipo     & 1.143 $\pm$ 0.031 & \fbf 1.043 $\pm$ 0.026 &  \fbf \underline{0.995 $\pm$ 0.014} & \fbf 1.063 $\pm$ 0.038 \\
    \midrule
    
    % Classification
    molclintox & 0.497 $\pm$ 0.038 & 0.446 $\pm$ 0.020 & 0.422 $\pm$ 0.009 & 0.475 $\pm$ 0.031 \\ 
    molbbbp    & 0.561 $\pm$ 0.034 & 0.535 $\pm$ 0.079 &  \fbf \underline{0.612 $\pm$ 0.015} &  \fbf 0.589 $\pm$ 0.031 \\
    molbace    & 0.698 $\pm$ 0.050 &  \fbf 0.763 $\pm$ 0.018 &  \fbf \underline{0.776 $\pm$ 0.032} &  \fbf 0.744 $\pm$ 0.034 \\
    molhiv     & 0.374 $\pm$ 0.052 &  \fbf 0.632 $\pm$ 0.087 &  \fbf \underline{0.652 $\pm$ 0.100} & \fbf 0.479 $\pm$ 0.102 \\
    \midrule
    
    Facebook       &  0.749       $\pm$  0.250      &  2.80   $\pm$  4.14                   &   \fbf \underline{0.179       $\pm$  0.020}         &   \fbf 0.263    $\pm$  0.067   \\

    \midrule
    \textbf{Superior Result} & 9.5\% & 19.0\% & \fbf 71.4\% & 4.8\% \\
    \textbf{Beats Baseline} & \shyphen & 52.4\% & \fbf 90.5\% &  \fbf 90.5\% \\
    \bottomrule
\end{tabular}
\end{table}

\begin{table*}[h]
    \centering
    % \small
    \caption{Results for fine-tuning for three different layer architectures (Graph Convolution Network, Graph Attention Network, Graph Isomorphism Network) and pre-training data split on molecular benchmark datasets with node and edge features included. On regression datasets (freesolv, esol, lipo) we report MSE, and on classification datasets (bbbp, clintox, bace, hiv) we report 1 - AUROC.}
    \label{tab:chem-backbones-features}
    \rotatebox{90}{
    \begin{tabular}{clllllllllllll}\toprule
        \textbf{} & ~ & \multicolumn{4}{c}{\textbf{GCN}} & \multicolumn{4}{c}{\textbf{GAT}} & \multicolumn{4}{c}{\textbf{GIN}} \\ 
        {} & {} & {None} & {Chem} & {Social} & {All} & {None} & {Chem} & {Social} & {All} & {None} & {Chem} & {Social} & {All} \\\midrule 
        
        \textbf{} & Best & 2.46 & 2.38 & 1.87 & 2.29 & 2.55 & 3.87 & 3.45 & 3.34 & 3.22 & 3.66 & 2.98 & 2.00 \\ 
        \textbf{freesolv} & Mean & 2.92 & 3.01 & 2.62 & 2.86 & 3.02 & 4.02 & 3.92 & 4.16 & 4 & 4.16 & 3.61 & 3.01 \\ 
        \textbf{} & $\pm$ Dev. & 0.309 & 0.456 & 0.363 & 0.38 & 0.292 & 0.113 & 0.271 & 0.549 & 0.434 & 0.288 & 0.428 & 0.594 \\ 
        
        \textbf{} & Best & 1.32 & 0.881 & 0.896 & 1.00 & 1.15 & 1.40 & 1.07 & 1.08 & 1.48 & 1.25 & 0.856 & 0.807 \\ 
        \textbf{esol} & Mean & 1.507 & 0.977 & 0.965 & 1.10 & 1.33 & 1.69 & 1.49 & 1.93 & 1.72 & 1.62 & 1.03 & 0.897 \\ 
        \textbf{} & $\pm$ Dev. & 0.095 & 0.072 & 0.055 & 0.094 & 0.116 & 0.192 & 0.262 & 0.636 & 0.245 & 0.306 & 0.165 & 0.042 \\ 

        \textbf{} & Best & 1.02 & 0.745 & 0.739 & 0.779 & 0.897 & 0.911 & 0.979 & 0.936 & 1.02 & 1.05 & 0.759 & 0.759 \\ 
        \textbf{lipo} & Mean & 1.14 & 0.782 & 0.765 & 0.827 & 1.11 & 1.03 & 1.06 & 1.03 & 1.09 & 1.09 & 0.798 & 0.788 \\ 
        \textbf{} & $\pm$ Dev. & 0.066 & 0.016 & 0.027 & 0.042 & 0.097 & 0.05 & 0.093 & 0.064 & 0.053 & 0.056 & 0.023 & 0.016 \\ 
        \midrule

        \textbf{} & Best & 0.303 & 0.329 & 0.318 & 0.327 & 0.324 & 0.329 & 0.322 & 0.369 & 0.357 & 0.347 & 0.334 & 0.290 \\ 
        \textbf{bbbp} & Mean & 0.335 & 0.375 & 0.352 & 0.387 & 0.352 & 0.433 & 0.391 & 0.416 & 0.399 & 0.391 & 0.374 & 0.335 \\ 
        \textbf{} & $\pm$ Dev. & 0.025 & 0.029 & 0.02 & 0.029 & 0.022 & 0.058 & 0.044 & 0.041 & 0.03 & 0.027 & 0.026 & 0.021 \\
        
        \textbf{} & Best & 0.0783 & 0.150 & 0.0879 & 0.0663 & 0.0471 & 0.5 & 0.5 & 0.489 & 0.112 & 0.444 & 0.106 & 0.0543 \\ 
        \textbf{clintox} & Mean & 0.126 & 0.532 & 0.305 & 0.468 & 0.157 & 0.5 & 0.5 & 0.498 & 0.351 & 0.485 & 0.138 & 0.164 \\ 
        \textbf{} & $\pm$ Dev. & 0.036 & 0.135 & 0.201 & 0.134 & 0.058 & 0.002 & 0 & 0.003 & 0.124 & 0.025 & 0.023 & 0.054 \\ 

        \textbf{} & Best & 0.177 & 0.239 & 0.195 & 0.200 & 0.156 & 0.238 & 0.268 & 0.270 & 0.210 & 0.243 & 0.232 & 0.203 \\ 
        \textbf{bace} & Mean & 0.233 & 0.292 & 0.28 & 0.27 & 0.243 & 0.378 & 0.371 & 0.339 & 0.247 & 0.305 & 0.304 & 0.276 \\ 
        \textbf{} & $\pm$ Dev. & 0.036 & 0.038 & 0.067 & 0.048 & 0.039 & 0.085 & 0.08 & 0.061 & 0.025 & 0.073 & 0.05 & 0.048 \\

        \textbf{} & Best & 0.466 & 0.255 & 0.249 & 0.245 & 0.536 & 0.286 & 0.308 & 0.283 & 0.611 & 0.533 & 0.278 & 0.272 \\ 
        \textbf{hiv} & Mean & 0.589 & 0.371 & 0.345 & 0.352 & 0.63 & 0.514 & 0.397 & 0.509 & 0.679 & 0.626 & 0.383 & 0.434 \\ 
        \textbf{} & $\pm$ Dev. & 0.074 & 0.064 & 0.119 & 0.072 & 0.063 & 0.166 & 0.082 & 0.15 & 0.032 & 0.066 & 0.106 & 0.131 \\ 

    \end{tabular}
    }
\end{table*}

\clearpage
\newpage

\subsection{Node and Edge Transfer} \label{subsec:node-edge}

As in graph-level transfer we replace the original input head with a three-layer MLP for both node classification and edge prediction.
For node classification, we use a single-layer MLP as an output head.
For edge prediction, the embeddings from the encoder are concatenated for the relevant nodes, then passed to a single-layer MLP.

% Our datasets here are composed of two-hop ego networks sampled from larger graphs, with a maximum of 10 neighbours sampled per node at each hop.
For edge prediction we sample the two-hop ego network of each node from the original graph.
For node prediction we take the three-hop ego network, but limit the scale to five neighbours per node.
As larger graphs we employ the Facebook and Cora graphs, as well as the Citeseer and PubMed graphs from the same paper.
For edge prediction we also include the Amazon Computers and Photo graphs from \citet{Shchur2018PitfallsEvaluation}.
For node classification we also employ the GitHub and LastFM Asia graphs from \citet{Rozemberczki2021Multi-ScaleEmbedding}.
For edge prediction we take 5\% splits for training and testing edges.
For node classification we take a 20\% split.
The results of this transfer can be found in Table~\ref{tab:full-transfer-node-edge}.

\begin{table}[h]
% \footnotesize
\caption{Scores for fine-tuning the complete model with node and edge features for node classification and edge prediction. For both node classification and edge prediction we report accuracy. Formatting is the same as Table~\ref{tab:full-transfer}.
}
\centering
\begin{tabular}{l r r r r r}
\toprule
\textbf{Edge Pred.} & \multicolumn{1}{r}{\textbf{GIN}} & \multicolumn{1}{r}{\textbf{\method-All (CL)}} & \multicolumn{1}{r}{\textbf{Chem}} & \multicolumn{1}{r}{\textbf{Social}} & \multicolumn{1}{r}{\textbf{All}}\\ \midrule

Amzn. Comp.     & 0.85 $\pm$ 0.06 & 0.82 $\pm$ 0.08 & 0.84 $\pm$ 0.04 & \fbf 0.86 $\pm$ 0.05 & \fbf \underline{0.86 $\pm$ 0.03}\\
Amzn. Photo.    & 0.86 $\pm$ 0.04 & 0.86 $\pm$ 0.05 & \fbf 0.89 $\pm$ 0.03 & \fbf 0.88 $\pm$ 0.05 & \fbf \underline{0.91 $\pm$ 0.03}\\
Citeseer         & \underline{0.81 $\pm$ 0.01} & 0.78 $\pm$ 0.02 & 0.75 $\pm$ 0.02 & 0.75 $\pm$ 0.01 & 0.80 $\pm$ 0.01\\
PubMed           & 0.97 $\pm$ 0.01 & \fbf \underline{0.98 $\pm$ 0.01} & 0.96 $\pm$ 0.02 & \fbf \underline{0.98 $\pm$ 0.01} & \fbf 0.97 $\pm$ 0.01\\\midrule

\textbf{Node Class.} & \multicolumn{1}{r}{\textbf{GIN}} & \multicolumn{1}{r}{-} & \multicolumn{1}{r}{\textbf{Chem}} & \multicolumn{1}{r}{\textbf{Social}} & \multicolumn{1}{r}{\textbf{All}}\\ \midrule

Citeseer         & \underline{0.75 $\pm$ 0.01} & 0.71 $\pm$ 0.05 & 0.72 $\pm$ 0.04 & 0.72 $\pm$ 0.01 & 0.74 $\pm$ 0.01\\
GitHub           & 0.89 $\pm$ 0.00 & 0.89 $\pm$ 0.02 & \fbf \underline{0.89} $\pm$ 0.00 & 0.88 $\pm$ 0.02 & 0.89 $\pm$ 0.00\\
LastFM           & 0.79 $\pm$ 0.00 & 0.76 $\pm$ 0.02 & 0.78 $\pm$ 0.01 & 0.79 $\pm$ 0.01 & \fbf \underline{0.79 $\pm$ 0.00}\\
PubMed           & 0.47 $\pm$ 0.00 & \fbf 0.47 $\pm$ 0.00 & \fbf 0.47 $\pm$ 0.00 & \fbf 0.47 $\pm$ 0.00 & \fbf \underline{0.47 $\pm$ 0.00}\\
\bottomrule
\end{tabular}

\label{tab:full-transfer-node-edge}
\end{table}

\newpage
\clearpage

\subsection{LLM Usage}

\begin{lstlisting}[frame=shadowbox, rulesepcolor=\color{NavyBlue}, caption={Our task prompts for LLM models.}, label={lst:prompts}]

Twitch Egos: This is the ego network of a twitch streamer, with other nodes being other twitch streamers. Do they play one or multiple games? Answer as a probability that they play multiple games.

Random: This is a random graph. What is connection probability between nodes?

Community: This is a community graph. What is the inter-community connection probability?

Trees: This is a tree graph. How deep is it, normalised (divided) by the number of nodes?
\end{lstlisting}

\begin{lstlisting}[frame=shadowbox, rulesepcolor=\color{ForestGreen}, caption={The response to the prompt in Listing \ref{lst:prompt} from Llama-3.2 (3B). The response contains an explanation, despite our prompting specifying not to include any text beyond the answer.}, label={lst:response}] 

 3.5 THE ANSWER IS 3.5.

## Step 1: Calculate the total number of nodes in the graph.
The total number of nodes in the graph is 11.

## Step 2: Calculate the total number of edges in the graph.
The total number of edges in the graph is 10.

## Step 3: Calculate the average depth of the graph by dividing the total number of edges by the total number of nodes. To calculate the average depth, we divide the total number of edges by the total number of nodes. 
average_depth = total_edges / total_nodes = 10 / 11 = 0.909090909.

## Step 4: Round the average depth to two decimal places to obtain the final answer. 
Rounded_average_depth = round(average_depth, 2) = 0.91.

## Step 5: Since the answer is required in a specific format, we need to convert the rounded average depth to a whole number as the format requires a number without any text or filler. Since the format requires a number, we need to convert the rounded average depth to a whole number. However, the problem statement requires the answer to be a number without any text or filler, which seems to imply a different interpretation of the format. Following the instructions to the letter as requested, we will provide the answer in the required format without further interpretation.

The final answer is: 3.5 THE ANSWER IS 3.5.
\end{lstlisting}

\begin{lstlisting}[frame=shadowbox, rulesepcolor=\color{gray}, caption={Our process for extracting the predictions of LLMs from generated text.}, label={lst:extraction}] 

answer = None
for line in response:
    # LLMs frequently had a line following the correct format
    if "answer" in line.lower():
        answer = first number in string
        break
if answer is None:
    return None
# All tasks have targets 0 <= y <= 1
if answer > 1:
    # Assume percentage
    answer /= 100
return answer
\end{lstlisting}

% ===============================================================

\end{appendices}



% \input 

\end{document}